\documentclass[shortAfour,sageh,times]{sagej}
\usepackage{moreverb,url}
\usepackage[colorlinks,bookmarksopen,bookmarksnumbered,citecolor=red,urlcolor=red]{hyperref}

\usepackage[printonlyused,withpage]{acronym}
\usepackage{amscd}
\usepackage{tikz}

\hypersetup{
    pdftitle={},
    pdfauthor={},
    pdfkeywords={},
    pdfsubject={},
    pdfstartview=FitH,
    bookmarks=true,
    bookmarksopen=true,
    breaklinks=true,
    colorlinks=true,
    linkcolor=black,
    anchorcolor=black,
    citecolor=black,
    filecolor=black,
    menucolor=black,
    pagecolor=black,
    urlcolor=black
}

\newcommand{\bbm}{\begin{bmatrix}}
\newcommand{\ebm}{\end{bmatrix}}
\newcommand{\mbf}{\mathbf}
\newcommand{\mbs}[1]{{\boldsymbol{#1}}}

\newcommand{\beq}{\begin{equation}}
\newcommand{\eeq}{\end{equation}}
\newcommand{\bdis}{\begin{displaymath}}
\newcommand{\edis}{\end{displaymath}}
\newcommand{\beqn}[1]{\begin{subequations}\label{eq:#1}\begin{eqnarray}}
\newcommand{\eeqn}{\end{eqnarray}\end{subequations}}

\newcommand{\pri}[1]{\check{#1}}
\renewcommand{\vec}{\mbox{vec}}

\newcommand{\tr}{\mbox{tr}}

\acrodef{BA}{Bundle Adjustment}
\acrodef{DNN}{Deep Neural Network}
\acrodef{EM}{Expectation Maximization}
\acrodef{GEM}{Generalized Expectation Maximization}
\acrodef{HMM}{Hidden Markov Model}
\acrodef{LDS}{Linear Dynamical System}
\acrodef{LQG}{Linear Quadratic Gaussian}
\acrodef{LQR}{Linear Quadratic Regulator}
\acrodef{LTI}{Linear Time-Invariant}
\acrodef{RTS}{Rauch-Tung-Striebel}
\acrodef{SGD}{Stochastic Gradient Descent}
\acrodef{SLAM}{Simultaneous Localization and Mapping}
\acrodef{GVI}{Gaussian Variational Inference}
\acrodef{ESGVI}{Exactly Sparse Gaussian Variational Inference}
\acrodef{MAP}{Maximum A Posteriori}
\acrodef{ML}{Maximum Likelihood}
\acrodef{KL}{Kullback-Leibler}
\acrodef{PDF}{Probability Density Function}
\acrodef{NEES}{Normalized Estimation Squared Error}
\acrodef{KF}{Kalman Filter}
\acrodef{ISPKF}{Iterated Sigmapoint Kalman Filter}
\acrodef{ESGVI-GN}{ESGVI Gauss-Newton}
\acrodef{ELBO}{Evidence Lower Bound}
\acrodef{NGD}{Natural Gradient Descent}
\acrodef{FIM}{Fisher Information Matrix}
\acrodef{RANSAC}{Random Sample And Consensus}

\newcommand{\change}[1]{\color{black}#1\color{black}}

\graphicspath{{figs/}}

\begin{document}

\runninghead{Barfoot et al.}

\title{Exactly Sparse Gaussian Variational Inference with Application to Derivative-Free Batch Nonlinear State Estimation}

\author{Timothy D. Barfoot\affilnum{1} and James R. Forbes\affilnum{2} and David J. Yoon\affilnum{1}}
\affiliation{\affilnum{1}Institute for Aerospace Studies, University of Toronto\\
\affilnum{2}Department of Mechanical Engineering, McGill University}
\corrauth{Timothy D. Barfoot,
Institute for Aerospace Studies,
University of Toronto,
Toronto, Ontario,
M3H~5T6, Canada.}
\email{tim.barfoot@utoronto.ca}

\begin{abstract}
We present a \ac{GVI} technique that can be applied to large-scale nonlinear batch state estimation problems.  The main contribution is to show how to fit \change{both the mean and (inverse) covariance of a Gaussian } to the posterior efficiently, by exploiting factorization of the joint likelihood of the state and data, as is common in practical problems.  \change{This is different than \ac{MAP} estimation, which seeks the point estimate for the state that maximizes the posterior (i.e., the mode). } The proposed \ac{ESGVI} technique stores the inverse covariance matrix, which is typically very sparse (e.g., block-tridiagonal for classic state estimation).  We show that the only blocks of the (dense) covariance matrix that are required during the calculations correspond to the non-zero blocks of the inverse covariance matrix, and further show how to  calculate these blocks efficiently in the general \ac{GVI} problem.  \ac{ESGVI} operates iteratively, and while we can use analytical derivatives at each iteration, Gaussian cubature can be substituted, thereby producing an efficient derivative-free batch formulation.  \ac{ESGVI} simplifies to precisely the \ac{RTS} smoother in the batch linear estimation case, but goes beyond the `extended' \ac{RTS} smoother in the nonlinear case since it finds the best-fit Gaussian \change{(mean and covariance), not the \ac{MAP} point estimate. }  We demonstrate the technique on controlled simulation problems and a batch nonlinear \ac{SLAM} problem with an experimental dataset.
\end{abstract}


\maketitle

\section{Introduction}

Gauss pioneered the method of least squares out of necessity to predict the position of the dwarf planet Ceres after passing behind the Sun.  In his initial treatment of the subject \citep{Gauss1809}, he presented what we would consider a `likelihood' function, which was expressed as an exponential function of quadratic terms,
\begin{equation}
L(\mbf{x}) = \exp\left(-\frac{1}{2}(\mbf{x}-\mbf{z})^T\mbf{W}^{-1} (\mbf{x}-\mbf{z})\right),
\end{equation}
where $\mbf{x}$ is the state to be estimated, $\mbf{z}$ are measurements, and $\mbf{W}^{-1}$ is a weighting matrix. Gauss recognized that $L(\mbf{x})$ is maximized when $(\mbf{x}-\mbf{z})^T\mbf{W}^{-1} (\mbf{x}-\mbf{z})$ is minimized, leading to the weighted least-squares solution. He later proved that the least-squares estimate is optimal without any assumptions regarding the distribution errors \citep{Gauss1821,Gauss1823}, and his more general result was rediscovered by \citet{Markoff1912}, leading to the more commonly known Gauss-Markov theorem \citep{Bjorck1996}. 

If we adopt a Bayesian perspective \citep{bayes}, our goal is to compute the full posterior, $p(\mbf{x}|\mbf{z})$, by refining a prior, $p(\mbf{x})$, not just a point estimate, based on some measurements, $\mbf{z}$:
\begin{equation}
p(\mbf{x} | \mbf{z}) = \frac{p(\mbf{z} | \mbf{x}) p(\mbf{x})}{p(\mbf{z})} = \frac{p(\mbf{x},\mbf{z})}{p(\mbf{z})}.
\end{equation}
The full posterior is not a Gaussian \ac{PDF} for nonlinear measurement models, $p(\mbf{z}|\mbf{x})$.  We are therefore often satisfied with finding the maximum of the Bayesian posterior, which is called the \acf{MAP} approach.  The connection to least squares (for Gaussian noise) is seen by taking the negative logarithm of the likelihood function (and dropping constant terms), resulting in a nonlinear quadratic loss function that is minimized:
\begin{eqnarray}\nonumber
V(\mbf{x}) & = & -\ln p(\mbf{x},\mbf{z}) \; = \; \frac{1}{2}\,\mbf{e}\left(\mbf{x},\mbf{z}\right)^T \mbf{W}^{-1}\, \mbf{e}\left(\mbf{x},\mbf{z}\right),\\ 
\mbf{x}^\star & = & \arg \min_{\mbf{x}} V(\mbf{x}),
\end{eqnarray}
where $\mbf{e}(\cdot,\cdot)$ is the error and is a nonlinear function of the state, $\mbf{x}$, and measurements, $\mbf{z}$.  \change{The result is a {\em point solution}, the most likely $\mbf{x}$ given $\mbf{z}$. } A Bayesian prior can easily be included in the loss function and thus we refer to this problem as \ac{MAP} rather than \ac{ML} (no prior).  Although there are various methods for minimizing the above loss function, perhaps the most well known dates back, again, to Gauss, who described how nonlinear least-squares problems can be linearized and refined in an iterative process \citep{Abdulle2002}, a method that is now known as Gauss-Newton (GN), or the method of differential corrections \citep{Ortega1970}.  To this day, \ac{MAP} is the dominant approach employed for batch nonlinear estimation problems.

Rather than finding the maximum of the Bayesian posterior, our approach in this paper
\change{will be to find the best Gaussian approximation, in terms of the mean and (inverse) covariance, }
to the full posterior that is `closest' in terms of the \ac{KL} divergence between the two \citep{kullback51}.  This approach is referred to as {\em variational inference} or {\em variational Bayes} \citep{bishop06}.  As we will restrict ourselves to Gaussian approximations of the posterior, we will refer to this as {\em Gaussian variational inference} (\ac{GVI}).  While \ac{GVI} is not new, it is not commonly used in batch estimation problems, where the state size, $N$, can be very large.  Our main contribution in this paper, is to show how to make \ac{GVI} tractable for large-scale estimation problems.  Specifically, we will show how to exploit a joint likelihood for the state and measurements that \change{can be factored}, 
\begin{equation}
p(\mbf{x},\mbf{z}) = \prod_{k=1}^K p(\mbf{x}_k,\mbf{z}_k),
\end{equation}
where $\mbf{x}_k$ is a subset of the variables in $\mbf{x}$.   This type of factorization is very common in real-world robotics problems, for example, since each measurement typically only involves a small subset of the state variables and this is already exploited in the \ac{MAP} approach \citep{brown58,thrun04,walter07} for efficient solutions.  We extend this exploit to the \ac{GVI} approach by identifying that the inverse covariance matrix is {\em exactly sparse} when the likelihood factors, and most importantly, that we never actually need to compute the entire covariance matrix, which is typically dense and of size $N \times N$.  As a by-product of our approach, we also show how to use cubature points (e.g., sigmapoints) for some of the required calculations, resulting in an efficient derivative-free implementation for large-scale batch estimation.

The paper is organized as follows.  Section~\ref{sec:relatedwork} reviews some related work.  Section~\ref{sec:gaussianvariationalinference} sets up our \ac{GVI} approach in terms of the \ac{KL} functional that we seek to minimize.  It then derives a Newton-style iterative optimizer to calculate the \change{mean and (inverse) covariance } of the Gaussian approximation.  Section~\ref{sec:exactsparsity} shows how we can exploit a factored likelihood not only by showing the inverse covariance is exactly sparse (as it is in the \ac{MAP} formulation) but also showing that we only ever require the blocks of the covariance matrix corresponding to the non-zero blocks of the inverse covariance.  It also summarizes an existing method for calculating these required blocks of the covariance and shows how we can make use of sample-based methods to avoid the need to calculate derivatives of our models.  Section~\ref{sec:extensions} presents an alternate formulation of the variational approach that is more approximate but also more efficient and also shows how we can fold parameter estimation into the framework while still exploiting sparsity.  Section~\ref{sec:evaluation} provides some toy problems and a real-data robotics demonstration of the method.  Finally, Section~\ref{sec:conclusion} provides our conclusion and suggestions for future work.

\section{Related Work}

\label{sec:relatedwork}

Gaussian estimation has been a key tool employed in fields such as robotics, computer vision, aerospace, and more.  The famous \ac{KF} \citep{kalman60a}, for example, provides a recursive formula to propagate a Gaussian state estimate.  While the \ac{KF} only goes forward in time, the \acf{RTS} smoother \citep{rauch65} carries out forward and backward passes to efficiently estimate the state and can be shown to be carrying out full Bayesian inference for linear models \citep{barfoot17}.  \citet{sarkka13} provides a wonderful presentation of recursive Bayesian inference methods, for both linear and nonlinear models.  In computer vision and robotics, the important \ac{BA} \citep{brown58} / \acf{SLAM} \citep{durrantwhyte06a} problem is often cast as a batch Gaussian estimation problem \citep{triggs00,lu97,thrun05}, with more advanced solution methods required than simple forward/backward passes \citep{kaess08,kaess11}.

While the recursive methods are fundamentally important, here we concern ourselves with problems that require batch Gaussian inference.  In robotics, some canonical problems are batch trajectory estimation, pose-graph relaxation \citep{bourmaud16}, and \ac{BA}/\ac{SLAM}.  However, we can also pose control/planning \citep{dong16,mukadam18}, calibration \citep{pradeep14}, and three-dimensional modelling problems \citep{li11} as Gaussian inference, such that the number of commonplace applications is quite large.  Despite the widespread need for this tool, almost without exception we rely on \ac{MAP} estimation to `fit' a Gaussian, which is to say we find the most likely state in the Bayesian posterior and call this the `mean', then fit a Gaussian centered at the most likely state, which is referred to as the Laplace approximation \citep[p. 315]{bishop06}.  For linear models, this in fact does produce the exact Gaussian posterior.  For nonlinear models, however, the posterior is not Gaussian and then the Laplace approximation is a convenient approach that can be computed efficiently for large-scale problems.  The primary goal of this paper is to revisit the batch Gaussian inference problem in search of improvements over this popular method.

Within recursive estimation, attempts have been made to go beyond \ac{MAP}, in order to perform better on nonlinear problems.  The Bayes filter \citep{jazwinsky70} is a general method that can be approximated in many different ways including through the use of Monte Carlo integration \citep{thrun06} or the use of cubature rules (e.g., sigmapoints) \citep{julier96, sarkka13}.  These sample-based extensions also bring the convenience of not requiring analytical derivatives of nonlinear models to be calculated.  Thus, a secondary goal of the paper is to find principled ways of incorporating sample-based techniques within batch Gaussian inference.  

As we will see, the starting point for our paper will be a variational Bayes setup \citep{bishop06}.  We aim to find the Gaussian approximation that is closest to the full Bayesian posterior in terms of the \ac{KL} divergence between the two \citep{kullback51}.  This is a paradigm shift from the \ac{MAP} approach where the only parameter to be optimized is the `mean', while the Laplace-style covariance is computed post hoc.  In \ac{GVI}, we seek to find the best mean and covariance from the outset.  The challenge is how to do this efficiently for problems with a large state size; if the mean is size $N$, then the covariance will be $N \times N$, which for real-world problems could be prohibitively expensive.  However, as we will show, we can carry out full \ac{GVI} by exploiting the same problem structures we usually do in the \ac{MAP} approach.  This will come at the expense of some increased computational cost, but the computational complexity as a function of $N$ does not increase.  \change{\citet{ranganathan07} and recently \citet{davison19} discuss the use of loopy belief propagation to carry out large-scale Gaussian inference for robotics problems; our motivation is somewhat different in that we seek to improve on \ac{MAP} whereas these works investigate parallelization of the computations and further approximate the Gaussian variational estimate.} 

While this result is new in robotics, \citet{opper09} discuss a similar \ac{GVI} approach in machine learning.  They begin with the same \ac{KL} divergence and show how to calculate the derivatives of this functional with respect to the Gaussian parameters.  They go on to apply the method to Gaussian process regression problems \citep{rasmussen06}, of which batch trajectory estimation can be viewed as a special case \citep{barfoot14,anderson15}.  Our paper extends this work in several significant ways including (i) generalizing to any \ac{GVI} problem where the likelihood \change{can be factored}, (ii) devising a Newton-style iterative solver for both mean and inverse covariance, (iii) explicitly showing how to exploit problem-specific structure in the case of a factored likelihood to make the technique efficient, (iv) applying Gaussian cubature to avoid the need to calculate derivatives, and (v) demonstrating the approach on problems of interest in robotics.

\citet{kokkala14,kokkala16}, \citet{alaluhtala15}, \citet{garcia15}, \citet{gavsperin11}, and \citet{schon11} discuss a very similar approach to our \ac{GVI} scheme in the context of nonlinear smoothers and filters; some of these works also carry out parameter estimation of the motion and observation models, which we also discuss as it fits neatly into the variational approach \citep{neal98,ghahramani99}.   These works start from the same \ac{KL} divergence, show how to exploit factorization of the joint likelihood, and discuss how to apply sigmapoints \citep{kokkala14,kokkala16,gavsperin11} or particles \citep{schon11} to avoid the need to compute derivatives.  \citet{garcia15} is a filtering paper that follows a similar philosophy to the current paper by statistically linearizing about an iteratively improved posterior.  Our paper extends these works by (i) generalizing to any large-scale batch \ac{GVI} problems where the likelihood \change{can be factored } (not restricted to smoothers with block-tridiagonal inverse covariance), (ii) devising a Newton-style iterative solver for both mean and inverse covariance, (iii) explicitly showing how to exploit problem-specific structure in the case of a factored likelihood to make the technique efficient, and (iv) demonstrating the approach on problems of interest in robotics.

There have been a few additional approaches to applying sampled-based techniques to batch estimation; however, they are quite different from ours. \citet{park09} and \citet{roh07} present a batch estimator that uses the sigmapoint Kalman filter framework as the optimization method.  For each major iteration, they compute sigmapoints for the estimated mean and propagate them through the motion model over all time steps. Then, the measurements from all of these propagated sigmapoints are stacked in a large column vector and the standard sigmapoint measurement update is applied. Although this method has been reported to work well \citep{park09,roh07}, it is expensive because it requires constructing the full covariance matrix to obtain the Kalman gain in the measurement update step.  In this paper, we work with the inverse covariance matrix and show how to avoid ever constructing the full covariance matrix, which opens to the door to use on large-scale estimation problems.

\section{Gaussian Variational Inference}

\label{sec:gaussianvariationalinference}

\change{
This section poses the problem we are going to solve and proposes a general solution. Exploiting application-specific structure is discussed later, in Section~\ref{sec:exactsparsity}.  We first define the loss functional that we seek to minimize, then derive an optimization scheme in order to minimize it with respect to the parameters of a Gaussian.  As an aside, we show that our optimization scheme is equivalent to so-called \ac{NGD}.  Following this, we work our optimization scheme into a different form in preparation for exploiting application-specific structure and finally show that we can recover the classic \ac{RTS} smoother in the linear case.
}

\subsection{Loss Functional}

As is common in variational inference \citep{bishop06}, we seek to minimize the \ac{KL} divergence \citep{kullback51} between the true Bayesian posterior, $p(\mbf{x} | \mbf{z})$, and an approximation of the posterior, $q(\mbf{x})$, which in our case will be a multivariate Gaussian \ac{PDF},
\begin{gather}
\hspace*{-2.2in} q(\mbf{x}) = \mathcal{N}(\mbs{\mu}, \mbs{\Sigma}) \\ \hspace{0.3in} = \frac{1}{\sqrt{(2\pi)^N |\mbs{\Sigma}|}} \exp\left( -\frac{1}{2} (\mbf{x} - \mbs{\mu})^T \mbs{\Sigma}^{-1} (\mbf{x} - \mbs{\mu}) \right), \nonumber 
\end{gather}
where $|\cdot|$ is the determinant.  For practical robotics and computer vision problems, the dimension of the state, $N$, can become very large and so the main point of our paper is to show how to carry out \ac{GVI} in an efficient manner for large-scale problems\footnote{Note, we choose not to make a mean-field approximation, instead allowing all variables to be correlated}.

As \ac{KL} divergence is not symmetrical, we have a choice of using $\mbox{KL}(p||q)$ or $\mbox{KL}(q||p)$.  \citet[p. 467]{bishop06} provides a good discussion of the differences between these two functionals.  The former expression is given by
\begin{eqnarray}
\mbox{KL}(p||q) & = & - \int^{\mbs{\infty}}_{-\mbs{\infty}} p(\mbf{x}|\mbf{z}) \ln \left( \frac{q(\mbf{x})}{p(\mbf{x} | \mbf{z})} \right) \, d\mbf{x} \nonumber \\ & = & \mathbb{E}_p\left[ \ln p(\mbf{x}|\mbf{z}) - \ln q(\mbf{x})\right],
\end{eqnarray}
while the latter is
\begin{eqnarray}
\mbox{KL}(q||p) & = & - \int^{\mbs{\infty}}_{-\mbs{\infty}} q(\mbf{x}) \ln \left( \frac{p(\mbf{x} | \mbf{z})}{q(\mbf{x})} \right) \, d\mbf{x} \nonumber  \\ & = & \mathbb{E}_q\left[ \ln q(\mbf{x}) - \ln p(\mbf{x}|\mbf{z})\right],
\end{eqnarray}
where $\mbf{x} \in \mathbb{R}^N$ is the latent state that we seek to infer from data, $\mbf{z} \in \mathbb{R}^D$, and $\mathbb{E}[\cdot]$ is the expectation operator.  The key practical difference that leads us to choose $\mbox{KL}(q||p)$ is that the expectation is over our Gaussian estimate, $q(\mbf{x})$, rather than the true posterior, $p(\mbf{x}|\mbf{z})$.  We will show that we can use this fact to devise an efficient iterative scheme for $q(\mbf{x})$ that best approximates the posterior.  Moreover, our choice of $\mbox{KL}(q||p)$ leads naturally to also estimating parameters of the system \citep{neal98}, which we discuss in Section~\ref{sec:paramest}.  \change{\citet{ranganathan07} and \citet{davison19} discuss approximate Gaussian inference for robotics problems using loopy belief propagation, which is based on $\mbox{KL}(p||q)$ \citep[p. 505]{bishop06}; their emphasis is on parallelizing the computations whereas we are focused on improving the estimate over \ac{MAP}.}

We observe that our chosen \ac{KL} divergence can be written as
\begin{multline}
\mbox{KL}(q||p) =  \mathbb{E}_q[ -\ln p(\mbf{x},\mbf{z})]  \\ - \underbrace{\frac{1}{2} \ln \left( (2\pi e)^N |\mbs{\Sigma}| \right)}_{\mbox{entropy}} + \underbrace{\ln p(\mbf{z})}_{\mbox{constant}},
\end{multline}
where we have used the expression for the {\em entropy}, $-\int q(\mbf{x}) \ln q(\mbf{x}) d\mbf{x}$, for a Gaussian.  Noticing that the final term is a constant (i.e., it does not depend on $q(\mbf{x})$) we define the following loss functional that we seek to minimize with respect to $q(\mbf{x})$:
\begin{equation}
\label{eq:functional}
V(q) = \mathbb{E}_q[ \phi(\mbf{x})] + \frac{1}{2} \ln \left( |\mbs{\Sigma}^{-1}| \right),
\end{equation}
with $\phi(\mbf{x}) = - \ln p(\mbf{x},\mbf{z})$.
We deliberately switch from $\mbs{\Sigma}$ (covariance matrix) to $\mbs{\Sigma}^{-1}$ (inverse covariance matrix also known as the {\em information matrix} or {\em precision matrix}) in~\eqref{eq:functional} as the latter enjoys sparsity that the former does not; we will carry this forward and use $\mbs{\mu}$ and $\mbs{\Sigma}^{-1}$ as a complete description of $q(\mbf{x})$.  The first term in $V(q)$ encourages the solution to match the data while the second penalizes it for being too certain; although we did not experiment with this, a relative weighting (i.e., a \change{metaparameter}) between these two terms could be used to tune performance on other metrics of interest.  It is also worth mentioning that $V(q)$ is the negative of the so-called \ac{ELBO}, which we will consequently minimize.

\subsection{Optimization Scheme}

\change{Our next task is to define an optimization scheme to minimize the loss functional with respect to the mean, $\mbs{\mu}$, and inverse covariance, $\mbs{\Sigma}^{-1}$.  Our approach will be similar to a Newton-style optimizer.}

After a bit of calculus, the derivatives of our loss functional, $V(q)$, with respect to our Gaussian parameters, $\mbs{\mu}$ and $\mbs{\Sigma}^{-1}$, are given by \citep{opper09}
\begin{subequations}
\begin{eqnarray}
\frac{\partial V(q)}{\partial \mbs{\mu}^T} & = &  \mbs{\Sigma}^{-1} \mathbb{E}_q[ (\mbf{x} - \mbs{\mu}) \phi(\mbf{x})],  \label{eq:deriv1a}   \\
\frac{\partial^2 V(q)}{\partial \mbs{\mu}^T \partial \mbs{\mu}} & = & \mbs{\Sigma}^{-1} \mathbb{E}_q[ (\mbf{x} - \mbs{\mu}) (\mbf{x} - \mbs{\mu})^T \phi(\mbf{x})] \mbs{\Sigma}^{-1}  \nonumber \\ & & \qquad - \; \mbs{\Sigma}^{-1} \,\mathbb{E}_q[\phi(\mbf{x})], \label{eq:deriv1b}  \\
\frac{\partial V(q)}{\partial \mbs{\Sigma}^{-1}} & = &  -\frac{1}{2}\mathbb{E}_q[ (\mbf{x} - \mbs{\mu}) (\mbf{x} - \mbs{\mu})^T\phi(\mbf{x})]  \nonumber  \\ & & \qquad + \; \frac{1}{2}  \mbs{\Sigma} \,\mathbb{E}_q[\phi(\mbf{x})] + \frac{1}{2} \mbs{\Sigma}, \label{eq:deriv2}
\end{eqnarray}
\end{subequations}
\change{
where, comparing~\eqref{eq:deriv1b} and~\eqref{eq:deriv2}, we notice that
\begin{equation}
\label{eq:derivrelationship}
\frac{\partial^2 V(q)}{\partial \mbs{\mu}^T \partial \mbs{\mu}} = \mbs{\Sigma}^{-1} - 2 \mbf{\Sigma}^{-1}  \frac{\partial V(q)}{\partial \mbs{\Sigma}^{-1}} \mbs{\Sigma}^{-1}.
\end{equation}
This relationship is critical to defining our optimization scheme, which we do next.
}

To find extrema, we could attempt to set the first derivatives to zero, but it is not (in general) possible to isolate for $\mbs{\mu}$ and $\mbs{\Sigma}^{-1}$ in closed form.  Hence, we will define an iterative update scheme.  We begin by writing out a Taylor series expansion of $V(q)$ that is second order in $\delta\mbs{\mu}$ but only first order in $\delta\mbs{\Sigma}^{-1}$ \change{(second order would be difficult to calculate and covariance quantities are already quadratic $\mbf{x}$):}
\begin{multline}
\label{eq:taylorV}
V\left(q^{(i+1)}\right) \approx V\left(q^{(i)}\right) + \left(\left. \frac{\partial V(q)}{\partial \mbs{\mu}^T} \right|_{q^{(i)}} \right)^T \delta\mbs{\mu} \\ \qquad\;\; + \frac{1}{2} \delta\mbs{\mu}^T \left(  \left. \frac{\partial^2 V(q)}{\partial \mbs{\mu}^T \partial \mbs{\mu}} \right|_{q^{(i)}} \right) \delta\mbs{\mu} \\ + \mbox{tr}\left( \left. \frac{\partial V(q)}{\partial \mbs{\Sigma}^{-1}} \right|_{q^{(i)}} \; \delta\mbs{\Sigma}^{-1} \right),
\end{multline}
where $\delta\mbs{\mu} = \mbs{\mu}^{(i+1)} - \mbs{\mu}^{(i)}$ and $\delta\mbs{\Sigma}^{-1} = \left( \mbs{\Sigma}^{-1} \right)^{(i+1)} -  \left( \mbs{\Sigma}^{-1} \right)^{(i)}$ with $i$ the iteration index of our scheme. We now want to choose $\delta\mbs{\mu}$ and $\delta\mbs{\Sigma}^{-1}$ to force $V(q)$ to get smaller.

\change{
For the inverse covariance, $\mbs{\Sigma}^{-1}$, if we set the derivative, $\frac{\partial V(q)}{\partial \mbs{\Sigma}^{-1}}$ , to zero (for an extremum) in~\eqref{eq:derivrelationship} we immediately have
\begin{equation}
\label{eq:iter2}
\mbs{\Sigma}^{-1^{(i+1)}} = \left. \frac{\partial^2 V(q)}{\partial \mbs{\mu}^T \partial \mbs{\mu}} \right|_{q^{(i)}}, 
\end{equation}
where we place an index of $(i+1)$ on the left and $(i)$ on the right in order to define an iterative update.  Inserting~\eqref{eq:derivrelationship} again on the right we see that the change to the inverse covariance by using this update can also be written as
\begin{equation}
\delta\mbs{\Sigma}^{-1} = - 2 \left(\mbs{\Sigma}^{-1}\right)^{(i)} \left. \frac{\partial V(q)}{\partial \mbs{\Sigma}^{-1}} \right|_{q^{(i)}}  \left(\mbs{\Sigma}^{-1}\right)^{(i)}.
\end{equation}
Convergence of this scheme will be discussed below\footnote{\change{It is worth mentioning that~\eqref{eq:deriv2} ignores the fact that $\mbs{\Sigma}^{-1}$ is actually a symmetric matrix.  \citet{magnus19} discuss how to calculate the derivative of a function with respect to a matrix while accounting for its symmetry.  They show that if $\mbf{A}$ is the derivative of a function (with respect to a symmetric matrix, $\mbf{B}$) that ignores symmetry, then $\mbf{A} + \mbf{A}^T - \mbf{A} \circ \mbf{1}$ is the derivative accounting for the symmetry of $\mbf{B}$, where $\circ$ is the Hadamard (element-wise) product and $\mbf{1}$ is the identity matrix.  It is not too difficult to see that $\mbf{A} = \mbf{0}$ if and only if $\mbf{A} + \mbf{A}^T - \mbf{A} \circ \mbf{1} = \mbf{0}$ in the case of a symmetric $\mbf{A}$.  Therefore, if we want to find an extremum by setting the derivative to zero, we can simply set $\mbf{A} = \mbf{0}$ as long as $\mbf{A}$ is symmetric and this will account for the symmetry of $\mbf{B}$ correctly.  \citet{barfoot_arxiv20} investigates this issue more thoroughly.}}.
}



For the mean, $\mbs{\mu}$, we will take inspiration from the \ac{MAP} approach to Gaussian nonlinear batch estimation and employ a Newton-style update \citep{nocedal06}.  Since \change{our loss approximation}~\eqref{eq:taylorV} is locally quadratic in $\delta\mbs{\mu}$, we take the derivative with respect to $\delta\mbs{\mu}$ and set this to zero (to find the minimum).  This results in a linear system of equations for $\delta\mbs{\mu}$:
\begin{equation}
\underbrace{\left(  \left. \frac{\partial^2 V(q)}{\partial \mbs{\mu}^T \partial \mbs{\mu}} \right|_{q^{(i)}} \right)}_{\left(\mbs{\Sigma}^{-1}\right)^{(i+1)}} \delta\mbs{\mu} = -\left(\left. \frac{\partial V(q)}{\partial \mbs{\mu}^T} \right|_{q^{(i)}} \right), \label{eq:iter1}
\end{equation}
where we note the convenient reappearance of $\mbs{\Sigma}^{-1}$ as the left-hand side. 

Inserting our chosen scheme for $\delta\mbs{\mu}$ and $\delta\mbs{\Sigma}^{-1}$ into \change{the loss approximation}~\eqref{eq:taylorV}, we have
\begin{multline}
\label{eq:convguarantee}
V\left(q^{(i+1)}\right) - V\left(q^{(i)}\right) \\ \!\!\!\!\!\!\!\!\!\!\!\!\!\! \approx -\frac{1}{2} \underbrace{\delta\mbs{\mu}^T \, \left(\mbs{\Sigma}^{-1}\right)^{(i+1)} \, \delta\mbs{\mu}}_{\substack{\ge 0 \\ \mbox{with equality iff $\delta\mbs{\mu} = \mbf{0}$}}}  \\ - \frac{1}{2} \underbrace{ \, \mbox{tr}\left( \mbs{\Sigma}^{(i)} \, \delta\mbs{\Sigma}^{-1} \, \mbs{\Sigma}^{(i)} \, \delta\mbs{\Sigma}^{-1} \right) }_{\substack{\ge 0 \\ \mbox{with equality iff  $\delta\mbs{\Sigma}^{-1} = \mbf{0}$} \\ \mbox{(see Appendix~\ref{sec:AppDefiniteness})}}} \; \le \; 0,
\end{multline}
which shows that we will reduce our loss, $V(q)$, so long as $\delta\mbs{\mu}$ and $\delta\mbs{\Sigma}^{-1}$ are not both zero; this is true when the derivatives with respect to $\mbs{\mu}$ and $\mbs{\Sigma}^{-1}$ are not both zero, which occurs only at a local minimum of $V(q)$.  This is a local convergence guarantee only as the expression is based on our Taylor series expansion in~\eqref{eq:taylorV}.

\subsection{Natural Gradient Descent Interpretation}

\change{As an aside, } we can interpret our update for $\delta\mbs{\mu}$ and $\delta\mbs{\Sigma}^{-1}$ as carrying out so-called \acf{NGD} \citep{amari98,hoffman13,barfoot_arxiv20}, which exploits the information geometry to make the update more efficient than regular gradient descent.  To see this, we stack our variational parameters into a single column, $\mbs{\alpha}$, using the $\mbox{vec}(\cdot)$ operator, which converts a matrix to a vector by stacking its columns:
\begin{gather}
\mbs{\alpha} = \bbm \mbs{\mu} \\ \mbox{vec}\left( \mbs{\Sigma}^{-1} \right) \ebm, \quad \delta\mbs{\alpha} = \bbm \delta\mbs{\mu} \nonumber \\ \mbox{vec}\left( \delta\mbs{\Sigma}^{-1} \right) \ebm, \\ \quad \frac{\partial V(q)}{\partial \mbs{\alpha}^T} = \bbm \frac{\partial V(q)}{\partial \mbs{\mu}^T} \\ \mbox{vec}\left(\frac{\partial V(q)}{\partial \mbs{\Sigma}^{-1}} \right) \ebm. 
\end{gather}
The last expression is the gradient of the loss functional with respect to $\mbs{\alpha}$.

The \ac{NGD} update scheme can then be defined as
\begin{equation}
\delta\mbs{\alpha} = -\mbs{\mathcal{I}}_{\mbs{\alpha}}^{-1} \frac{\partial V(q)}{\partial \mbs{\alpha}^T},
\end{equation}
where $\mbs{\mathcal{I}}_{\mbs{\alpha}}$ is the \ac{FIM} \citep{fisher22} for the variational parameter, $\mbs{\alpha}$, and its calculation can be found in Appendix~\ref{sec:AppFisher}.  Inserting the details of the components of the above we have
\begin{multline}
\bbm \delta\mbs{\mu} \\ \mbox{vec}\left( \delta\mbs{\Sigma}^{-1} \right) \ebm \\ = - \bbm \mbs{\Sigma}^{-1} & \mbf{0} \\ \mbf{0} & \frac{1}{2} \left( \mbs{\Sigma} \otimes \mbs{\Sigma} \right)\ebm^{-1} \bbm \frac{\partial V(q)}{\partial \mbs{\mu}^T} \\ \mbox{vec}\left(\frac{\partial V(q)}{\partial \mbs{\Sigma}^{-1}} \right) \ebm,
\end{multline}
where $\otimes$ is the Kronecker product.  Extracting the individual updates we see
\begin{subequations}
\begin{eqnarray}
\delta\mbs{\mu} & = & -\mbs{\Sigma} \frac{\partial V(q)}{\partial \mbs{\mu}^T}, \\
\mbox{vec}\left( \delta\mbs{\Sigma}^{-1} \right) & = & -2 \left( \mbs{\Sigma}^{-1} \otimes \mbs{\Sigma}^{-1} \right) \mbox{vec}\left(\frac{\partial V(q)}{\partial \mbs{\Sigma}^{-1}} \right). \nonumber \\ & & 
\end{eqnarray}
\end{subequations}
Finally, using that $\mbox{vec}(\mbf{A}\mbf{B}\mbf{C}) \equiv (\mbf{C}^T \otimes \mbf{A}) \,\mbox{vec}(\mbf{B})$, we have
\begin{subequations}
\begin{eqnarray}
\mbs{\Sigma}^{-1} \, \delta\mbs{\mu} & = & -\frac{\partial V(q)}{\partial \mbs{\mu}^T}, \\
\delta\mbs{\Sigma}^{-1}  & = & -2 \mbs{\Sigma}^{-1} \frac{\partial V(q)}{\partial \mbs{\Sigma}^{-1}} \mbs{\Sigma}^{-1},
\end{eqnarray}
\end{subequations}
which is the same set of updates as in the previous subsection.

\subsection{Stein's Lemma}

While our iterative scheme could be implemented as is, it will be expensive (i.e., $O(N^3)$ per iteration) for large problems.
The next section will show how to exploit sparsity to make the scheme efficient and, in preparation for that, we will manipulate our update equations into a slightly different form using {\em Stein's lemma} \citep{stein81}.  In our notation, the lemma says
\begin{equation} \label{eq:stein1}
\mathbb{E}_q[ (\mbf{x} - \mbs{\mu}) f(\mbf{x})] \equiv \mbs{\Sigma} \, \mathbb{E}_q\left[ \frac{\partial f(\mbf{x})}{\partial \mbf{x}^T}\right],
\end{equation}
where $q(\mbf{x}) = \mathcal{N}(\mbs{\mu}, \mbs{\Sigma})$ is a Gaussian random variable and $f(\cdot)$ is any nonlinear differentiable function. A double application of Stein's lemma also reveals
\begin{multline} \label{eq:stein2}
\mathbb{E}_q [ (\mbf{x} - \mbs{\mu}) (\mbf{x} - \mbs{\mu})^T f(\mbf{x})] \\ \equiv \mbs{\Sigma} \, \mathbb{E}_q\left[ \frac{\partial^2 f(\mbf{x})}{\partial \mbf{x}^T \partial \mbf{x}}\right] \, \mbs{\Sigma} + \mbs{\Sigma} \, \mathbb{E}_q [ f(\mbf{x}) ], 
\end{multline}
assuming $f(\cdot)$ is twice differentiable.  Combining Stein's lemma with \change{our loss derivatives in}~\eqref{eq:deriv1a}, \eqref{eq:deriv1b}, and~\eqref{eq:deriv2}, we have the useful identities
\begin{subequations}
\label{eq:derivident}
\begin{eqnarray}
\frac{\partial}{\partial \mbs{\mu}^T}\mathbb{E}_q[f(\mbf{x})] & \equiv & \mathbb{E}_q\left[\frac{\partial f(\mbf{x})}{\partial \mbf{x}^T}\right], \label{eq:derivident1}\\ \frac{\partial^2}{\partial \mbs{\mu}^T\partial\mbs{\mu}}\mathbb{E}_q[f(\mbf{x})] & \equiv & \mathbb{E}_q\left[\frac{\partial^2 f(\mbf{x})}{\partial \mbf{x}^T \partial \mbf{x}}\right] \label{eq:derivident2} \\ & \equiv & -2 \mbs{\Sigma}^{-1} \left(\frac{\partial}{\partial \mbs{\Sigma}^{-1}} \mathbb{E}_q[f(\mbf{x})]  \right)\mbs{\Sigma}^{-1}, \nonumber  
\end{eqnarray}
\end{subequations}
which we will have occasion to use later on; \change{Appendix~\ref{sec:AppDerivIdent} provides the derivations.}

We can apply \change{Stein's lemma from~\eqref{eq:stein1} and~\eqref{eq:stein2} to our optimization scheme in~\eqref{eq:iter2} and~\eqref{eq:iter1} } to write the iterative updates compactly as
\begin{subequations}\label{eq:iterstein}
\begin{eqnarray}
\left(\mbs{\Sigma}^{-1}\right)^{(i+1)} & = &   \mathbb{E}_{q^{(i)}}\left[ \frac{\partial^2}{\partial \mbf{x}^T \partial \mbf{x}} \phi(\mbf{x})\right], \label{eq:iterstein1} \\
\left(\mbs{\Sigma}^{-1}\right)^{(i+1)} \, \delta\mbs{\mu} & = & - \mathbb{E}_{q^{(i)}}\left[ \frac{\partial}{\partial \mbf{x}^T} \phi(\mbf{x})\right],  \label{eq:iterstein2}   \\
\mbs{\mu}^{(i+1)} & = & \mbs{\mu}^{(i)} + \delta\mbs{\mu}.  \label{eq:iterstein3}
\end{eqnarray}
\end{subequations}
\citet[App. C]{alaluhtala15} also make use of Stein's lemma in this way in the context of Gaussian variational smoothers.  In general, this iterative scheme will still be expensive for large problems and so we will look to exploit structure to make \ac{GVI} more efficient.  As only the first and second derivatives of $\phi(\mbf{x})$ are required, we can drop any constant terms (i.e., the normalization constant of $p(\mbf{x},\mbf{z})$).

\change{Notably, our optimization scheme in~\eqref{eq:iterstein} is identical to the \ac{MAP} approach (with the Laplace covariance approximation) if we approximate the expectations using only the mean of $q(\mbf{x})$.  Thus, \ac{MAP} with Laplace can be viewed as an approximation of the more general approach we discuss in this paper.}

\subsection{Recovery of the RTS Smoother}

Before moving on, we briefly show that our \ac{GVI} formulation produces the discrete-time \ac{RTS} smoother result in the linear case.  As is shown by \citet[\S3, p. 44]{barfoot17}, the batch linear state estimation problem can be written in {\em lifted form} (i.e., at the trajectory level):
\begin{subequations}
\begin{eqnarray}
\mbf{x} & = & \mbf{A} (\mbf{B}\mbf{u} + \mbf{w}), \\
\mbf{y} & = & \mbf{C} \mbf{x} + \mbf{n},
\end{eqnarray}
\end{subequations}
where $\mbf{x}$ is the entire trajectory (states over time), $\mbf{u}$ are the control inputs, $\mbf{y}$ are the sensor outputs, $\mbf{w} \sim \mathcal{N}(\mbf{0}, \mbf{Q})$ is process noise, $\mbf{n} \sim \mathcal{N}(\mbf{0}, \mbf{R})$ is measurement noise, $\mbf{A}$ is the lifted transition matrix, $\mbf{B}$ is the lifted control matrix, and $\mbf{C}$ is the lifted observation matrix.  We then have
\begin{multline}
\phi(\mbf{x}) = \frac{1}{2} \left(\mbf{B}\mbf{u} - \mbf{A}^{-1} \mbf{x}\right)^T \mbf{Q}^{-1}  \left(\mbf{B}\mbf{u} - \mbf{A}^{-1} \mbf{x}\right) \\ + \frac{1}{2} \left( \mbf{y} - \mbf{C} \mbf{x} \right)^T \mbf{R}^{-1} \left( \mbf{y} - \mbf{C} \mbf{x} \right).
\end{multline}
\change{The expected derivatives can be calculated analytically for this linear problem:}
\begin{subequations}
\begin{eqnarray}
& & \hspace*{-0.3in} \mathbb{E}_{q}\left[ \frac{\partial^2}{\partial \mbf{x}^T \partial \mbf{x}} \phi(\mbf{x})\right]  \label{eq:RTScov} \\ & = & \mbf{A}^{-T} \mbf{Q}^{-1}  \mbf{A}^{-1} + \mbf{C}^T \mbf{R}^{-1} \mbf{C}, \nonumber  \\
& & \hspace*{-0.3in} \mathbb{E}_{q}\left[ \frac{\partial}{\partial \mbf{x}^T} \phi(\mbf{x})\right] \label{eq:RTSmean} \\ & = & -\mbf{A}^{-T} \mbf{Q}^{-1} \left(\mbf{B}\mbf{u} - \mbf{A}^{-1} \mbs{\mu}\right) - \mbf{C}^T \mbf{R}^{-1} \left( \mbf{y} - \mbf{C} \mbs{\mu} \right). \nonumber
\end{eqnarray} 
\end{subequations}
At convergence, \eqref{eq:RTSmean} must be zero, so we have
\begin{subequations}
\begin{eqnarray}
\mbs{\Sigma}^{-1} & = & \underbrace{\mbf{A}^{-T} \mbf{Q}^{-1}  \mbf{A}^{-1} + \mbf{C}^T \mbf{R}^{-1} \mbf{C}}_{\mbox{block-tridiagonal}}, \\
\mbs{\Sigma}^{-1} \mbs{\mu}  & = & \mbf{A}^{-T} \mbf{Q}^{-1} \mbf{B}\mbf{u}  + \mbf{C}^T \mbf{R}^{-1} \mbf{y}, 
\end{eqnarray}
\end{subequations}
which can be solved efficiently for $\mbs{\mu}$ due to the block-tridiagonal nature of $\mbs{\Sigma}^{-1}$; from here, \citet[\S3, p.55]{barfoot17} shows the algebraic equivalence of this form to the canonical \ac{RTS} smoother.  Thus, our \ac{GVI} approach still reproduces the classic linear result.  However, we can also now address nonlinear problems more completely than the \ac{MAP} case.

\section{Exact Sparsity}

\change{This section shows how to exploit application-specific structure to make the optimization scheme of the previous section efficient for large-scale problems.  We first show that when the joint likelihood of the state and data can be factored, the calculation of the required expectations in our optimization scheme is exactly sparse, meaning we only need the marginals of the covariance associated with each factor.  We then discuss how we can calculate these marginals efficiently from the inverse covariance,  for any \ac{GVI} problem.  Finally, we show how to use sigmapoints drawn from these marginals to implement the full optimization scheme.}

\label{sec:exactsparsity}

\subsection{Factored Joint Likelihood}

We have seen in the previous section that the iterative update scheme relies on calculating three expectations:
\begin{equation}
\underbrace{\mathbb{E}_q[\phi(\mbf{x})]}_{\mbox{scalar}}, \quad \underbrace{\mathbb{E}_q\left[ \frac{\partial}{\partial \mbf{x}^T} \phi(\mbf{x})\right]}_{\mbox{column}}, \quad \underbrace{\mathbb{E}_q\left[ \frac{\partial^2}{\partial \mbf{x}^T \partial \mbf{x}} \phi(\mbf{x})\right]}_{\mbox{matrix}}, \label{eq:expectations}
\end{equation}
where we drop the iteration index for now.  Let us now assume that the joint state/data likelihood can be factored such that we can write its negative log-likelihood as
\begin{equation}
\phi(\mbf{x}) =  \sum_{k=1}^K \phi_k( \mbf{x}_k ), 
\end{equation}
where $\phi_k(\mbf{x}_k) = - \ln p(\mbf{x}_k,\mbf{z}_k)$ is the $k$th (negative log) factor expression, $\mbf{x}_k$ is a {\em subset} of variables in $\mbf{x}$ associated with the $k$th factor, and $\mbf{z}_k$ is a subset of the data in $\mbf{z}$ associated with the $k$th factor.  

Let us consider the first (scalar) expectation in~\eqref{eq:expectations}.  We can insert the factored likelihood and see what happens:
\begin{multline}\label{eq:expectation1}
\mathbb{E}_q[\phi(\mbf{x})] = \mathbb{E}_q\left[\sum_{k=1}^K \phi_k( \mbf{x}_k )\right] \\ = \sum_{k=1}^K \mathbb{E}_q[\phi_k( \mbf{x}_k )] = \sum_{k=1}^K \mathbb{E}_{q_k}[\phi_k( \mbf{x}_k )],
\end{multline}
where the last step is subtle but paramount: the expectation simplifies from being over $q = q(\mbf{x})$, the full Gaussian estimate, to being over $q_k = q_k(\mbf{x}_k)$, the {\em marginal} of the estimate for just the variables in each factor.  This is not an approximation and the implications are many.  

The other two expectations \change{(column and matrix) } in~\eqref{eq:expectations} enjoy similar simplifications and more, but require a bit more explanation.  Let $\mbf{P}_k$ be a projection matrix such that it extracts $\mbf{x}_k$ from $\mbf{x}$:
\begin{equation}
\mbf{x}_k = \mbf{P}_k \mbf{x}.
\end{equation}
Then inserting the factored expression into the second (column) expectation we have
\begin{eqnarray}
\mathbb{E}_q\left[ \frac{\partial}{\partial \mbf{x}^T}\phi(\mbf{x})\right]  & = & \mathbb{E}_q\left[ \frac{\partial}{\partial \mbf{x}^T} \sum_{k=1}^K \phi_k( \mbf{x}_k )\right] \nonumber  \\ & = & \sum_{k=1}^K \mathbb{E}_q\left[ \frac{\partial}{\partial \mbf{x}^T} \phi_k( \mbf{x}_k )\right]  \nonumber \\ & = & \sum_{k=1}^K \mbf{P}_k^T \, \mathbb{E}_q \left[ \frac{\partial}{\partial \mbf{x}_k^T} \phi_k( \mbf{x}_k )\right] \nonumber  \\ & = & \sum_{k=1}^K \mbf{P}_k^T \, \mathbb{E}_{q_k} \left[ \frac{\partial}{\partial \mbf{x}_k^T} \phi_k( \mbf{x}_k )\right]. \quad \label{eq:expectation2}
\end{eqnarray}
For factor $k$, we are able to simplify the derivative from being with respect to $\mbf{x}$, to being with respect to $\mbf{x}_k$, since there is no dependence on the variables not in $\mbf{x}_k$ and hence the derivative with respect to those variables is zero; we use the projection matrix (as a dilation matrix) to map the derivative back into the appropriate rows of the overall result.  After this, the expectation again simplifies to being with respect to $q_k = q_k(\mbf{x}_k)$, the marginal of the estimate for just the variables in factor $k$.  For the last (matrix) expectation we have a similar result:
\begin{eqnarray}
& & \hspace*{-0.5in} \mathbb{E}_q\left[ \frac{\partial^2}{\partial \mbf{x}^T \partial \mbf{x}} \phi(\mbf{x})\right] \nonumber \\ & = & \mathbb{E}_q\left[ \frac{\partial^2}{\partial \mbf{x}^T \partial \mbf{x}} \sum_{k=1}^K \phi_k( \mbf{x}_k )\right] \nonumber \\ & = & \sum_{k=1}^K \mathbb{E}_q\left[ \frac{\partial^2}{\partial \mbf{x}^T \partial \mbf{x}} \phi_k( \mbf{x}_k )\right] \nonumber \\ & = & \sum_{k=1}^K \mbf{P}_k^T \, \mathbb{E}_q \left[ \frac{\partial^2}{\partial \mbf{x}_k^T \partial \mbf{x}_k} \phi_k( \mbf{x}_k )\right] \mbf{P}_k \nonumber \\ & = & \sum_{k=1}^K \mbf{P}_k^T \,\mathbb{E}_{q_k} \left[ \frac{\partial^2}{\partial \mbf{x}_k^T \partial \mbf{x}_k} \phi_k( \mbf{x}_k )\right] \mbf{P}_k. \label{eq:expectation3}
\end{eqnarray}
\change{The simplified expectations in}~\eqref{eq:expectation1},~\eqref{eq:expectation2}, and~\eqref{eq:expectation3} are the key tools that enable our \ac{ESGVI} approach and we now make several remarks about them:
\begin{enumerate}
\item We do not require the full Gaussian estimate, $q(\mbf{x})$, to evaluate the three expectations involved in our iterative scheme but rather we only require the marginals associated with each factor, $q_k(\mbf{x}_k)$.  This can represent a huge computational and storage savings in practical problems because it means that we never need to fully construct and store the (usually dense) covariance matrix, $\mbs{\Sigma}$.  \citet{schon11,gavsperin11,kokkala16} also show how the required expectations are simplified to being over the marginals specifically for the smoother problem, but here we have generalized that result to any factorization of the joint likelihood.

\item Looking to \change{the covariance update in}~\eqref{eq:iterstein1} and now the simplification in~\eqref{eq:expectation3}, we know that $\mbs{\Sigma}^{-1}$ will be exactly sparse (with the pattern depending on the nature of the factors) and that {\em the sparsity pattern will remain constant as we iterate}.  A  fixed sparsity pattern ensures that we can build a custom sparse solver \change{for the mean}~\eqref{eq:iterstein2} and use it safely at each iteration; for example, in the batch state estimation problem, $\mbs{\Sigma}^{-1}$ is block-tridiagonal (under a chronological variable ordering).

\item As a reminder, marginalization of a Gaussian amounts to projection such that
\begin{equation}
q_k(\mbf{x}_k) = \mathcal{N}\left( \mbs{\mu}_k, \mbs{\Sigma}_{kk} \right) = \mathcal{N}\left( \mbf{P}_k\mbs{\mu}, \mbf{P}_k\mbs{\Sigma} \mbf{P}_k^T \right),
\end{equation}
so that it is just specific sub-blocks of the full covariance matrix that are ever required.

\item The {\em only} sub-blocks of $\mbs{\Sigma}$ that we require are precisely the ones corresponding to the non-zero sub-blocks of $\mbs{\Sigma}^{-1}$ (which is typically highly sparse).  We can see this more plainly by writing
\begin{equation}
\mbs{\Sigma}^{-1} = \sum_{k=1}^K \mbf{P}_k^T \,\mathbb{E}_{q_k} \left[ \frac{\partial^2}{\partial \mbf{x}_k^T \partial \mbf{x}_k} \phi_k( \mbf{x}_k )\right] \mbf{P}_k,
\end{equation}
where we can see that each factor uses some sub-blocks, $\mbs{\Sigma}_{kk} = \mbf{P}_k\mbs{\Sigma} \mbf{P}_k^T$, to evaluate the expectation, and then the results are inserted back into the same elements of $\mbs{\Sigma}^{-1}$.

\item It turns out that we can extract the required sub-blocks of $\mbs{\Sigma}$ very efficiently.  For example, for batch state estimation, with a block-tridiagonal $\mbs{\Sigma}^{-1}$, we can piggyback the calculation of the required blocks (i.e., the three main block diagonals of $\mbs{\Sigma}$) onto the solution \change{for the mean in}~\eqref{eq:iterstein2} \citep{meurant92, barfoot17} while keeping the complexity of the solver the same.  However, we can also compute the required blocks of $\mbs{\Sigma}$ efficiently in the general case \citep{takahashi73}, and the next section is devoted to discussion of this topic.

\end{enumerate}

Some of these remarks may seem familiar to those used to working with a \ac{MAP} approach to batch state estimation (e.g., the sparsity pattern of $\mbs{\Sigma}^{-1}$ exists and is constant across iterations).   But now we are performing \ac{GVI} that iterates over a full Gaussian \ac{PDF} (i.e., mean and covariance) not just a point estimate (i.e., mean only).

At this point, the only approximation that we have made is that our estimate of the posterior is Gaussian.  However, to implement the scheme in practice, we need to choose a method to actually compute the (marginal) expectations in~\eqref{eq:expectation1},~\eqref{eq:expectation2}, and~\eqref{eq:expectation3}.  There are many choices including linearization, Monte Carlo sampling, and also deterministic sampling.  We will show how to use sampling methods in a later section.

\newcounter{mytempeqncnt}
\begin{table*}[t]
\vspace*{-0.2in}
\normalsize
\setcounter{mytempeqncnt}{\value{equation}}
\setcounter{equation}{44}
\begin{multline}
\bbm  \ddots &  &  &  \\ \hdots & \mbs{\Sigma}_{K-2,K-2} &  &  \\ \hdots & \mbs{\Sigma}_{K-1,K-2} & \mbs{\Sigma}_{K-1,K-1} &  \\ \hdots & \mbs{\Sigma}_{K,K-2} & \mbs{\Sigma}_{K,K-1}  & \mbs{\Sigma}_{K,K} \ebm =  \bbm \ddots &  & & \\ \hdots & \mbf{D}^{-1}_{K-2,K-2}  & & \\ \hdots & \mbf{0} & \mbf{D}^{-1}_{K-1,K-1} & \\ \hdots & \mbf{0} & \mbf{0} & \mbf{D}^{-1}_{K,K} \ebm  \\ - \bbm  \ddots & \vdots & \vdots & \vdots \\ \hdots & \mbs{\Sigma}_{K-2,K-2} & \mbs{\Sigma}_{K-2,K-1} & \mbs{\Sigma}_{K-2,K} \\ \hdots & \mbs{\Sigma}_{K-1,K-2} & \mbs{\Sigma}_{K-1,K-1} & \mbs{\Sigma}_{K-1,K} \\ \hdots & \mbs{\Sigma}_{K,K-2} & \mbs{\Sigma}_{K,K-1} & \mbs{\Sigma}_{K,K} \ebm\bbm \ddots &  &  &  \\ \hdots & \mbf{0} &  & \\ \hdots & \mbf{L}_{K-1,K-2} & \mbf{0} &  \\ \hdots &  \mbf{L}_{K,K-2} & \mbf{L}_{K,K-1} & \mbf{0} \ebm \label{eq:takalong}
\end{multline} 
\setcounter{equation}{\value{mytempeqncnt}}
\vspace*{-0.2in}
\hrulefill
\end{table*}

\subsection{Partial Computation of the Covariance}

For completeness, we briefly summarize how it is possible to compute the blocks of $\mbs{\Sigma}$ (typically dense) corresponding to the non-zero sub-blocks of $\mbs{\Sigma}^{-1}$ (typically very sparse) in an efficient manner.  This idea was first proposed by \citet{takahashi73} in the context of circuit theory and was later used by \citet{broussolle78} in a state estimation context where the matrix of interest was a covariance matrix like ours.  \citet{erisman75} provide a proof of the closure of the Takahashi et al. procedure and also discuss algorithmic complexity.  \change{More recently, \citet[App. B.4]{triggs00} and \citet{kaess09} discuss methods to calculate specific blocks of the covariance matrix efficiently from the inverse covariance for computer vision and robotics applications, but do not discuss doing so for the complete set of covariance blocks corresponding to the non-zero blocks of the inverse covariance matrix.}

At each iteration of our \ac{GVI} approach, we are required to solve a system of linear equations for the change in the mean:
\begin{equation}
\mbs{\Sigma}^{-1} \, \delta\mbs{\mu} = \mbf{r},
\end{equation}
where $\mbf{r}$ is the right-hand side in~\eqref{eq:iterstein2}.  We start by carrying out a sparse lower-diagonal-upper decomposition,
\begin{equation}
\label{eq:LDL}
\mbs{\Sigma}^{-1} = \mbf{L} \mbf{D} \mbf{L}^T,
\end{equation}
where $\mbf{D}$ is diagonal and $\mbf{L}$ is lower-triangular with ones on the main diagonal (and sparse).  The cost of this decomposition will depend on the nature of the prior and measurement factors. The key thing is that the sparsity pattern of $\mbf{L}$ is a direct function of the factors' variable dependencies and can be determined in advance; more on this below.  We can then solve the following two systems of equations for the change in the mean:
\begin{eqnarray}
\left(\mbf{L} \mbf{D}\right) \, \mbf{v} & = & \mbf{r}, \quad \mbox{(sparse forward substitution)} \\
\mbf{L}^T \, \delta\mbs{\mu} & = & \mbf{v}. \quad \mbox{(sparse backward substitution)}  \;\;\;
\end{eqnarray}

To solve for the required blocks of $\mbs{\Sigma}$, we notice that
\begin{equation}
\mbf{L} \mbf{D} \mbf{L}^T \mbs{\Sigma} = \mbf{1},
\end{equation}
where $\mbf{1}$ is the identity matrix.  We can premultiply by the inverse of $\mbf{L}\mbf{D}$ to arrive at
\begin{equation}
\mbf{L}^T \mbs{\Sigma} =  \mbf{D}^{-1} \mbf{L}^{-1},
\end{equation}
where $\mbf{L}^{-1}$ will in general no longer be sparse.  Taking the transpose and adding $\mbs{\Sigma} - \mbs{\Sigma}\mbf{L}$ to both sides we have \citep{takahashi73}
\begin{equation}
\mbs{\Sigma} =  \mbf{L}^{-T} \mbf{D}^{-1} +  \mbs{\Sigma} \left( \mbf{1} - \mbf{L} \right).
\end{equation}
Since $\mbs{\Sigma}$ is symmetric, we only require (at most) calculation of the main diagonal and the lower-half blocks and, as it turns out, this can also be done through a backward substitution pass.  To see this we expand the lower-half blocks as in~\eqref{eq:takalong} at the top of the page,
where we only show the blocks necessary for the calculation of the lower-half of $\mbs{\Sigma}$; critically, $\mbf{L}^{-T}$ is unnecessary since it only affects the upper-half blocks of $\mbs{\Sigma}$ and is therefore dropped.  Temporarily ignoring the need to exploit sparsity, we see that we can calculate the lower-half blocks of $\mbs{\Sigma}$ through backward substitution:
\stepcounter{equation}
\begin{subequations}
\begin{eqnarray}
\mbs{\Sigma}_{K,K} & = & \mbf{D}_{K,K}^{-1}, \\
\mbs{\Sigma}_{K,K-1} & = & -\mbs{\Sigma}_{K,K} \mbf{L}_{K,K-1}, \\
\mbs{\Sigma}_{K-1,K-1} & = & \mbf{D}_{K-1,K-1}^{-1} -  \mbs{\Sigma}_{K-1,K}  \mbf{L}_{K,K-1}, \\
& \vdots & \nonumber \\
\mbs{\Sigma}_{j,k} & = &  \delta(j,k) \,\mbf{D}_{j,k}^{-1} - \sum_{\ell=k+1}^K \mbs{\Sigma}_{j,\ell} \mbf{L}_{\ell,k}, \; (j\geq k) \nonumber \\ & & \label{eq:tak1} 
\end{eqnarray}
\end{subequations}
where $\delta(\cdot,\cdot)$ is the Kronecker delta function.  

\begin{table*}[t]
\caption{Example sparsity patterns of $\mbs{\Sigma}^{-1}$ and the corresponding sparsity patterns of factor $\mbf{L}$, \change{where $*$ indicates non-zero}.  The set of zero entries of the lower-half of $\mbf{L}$ is a subset of the zero entries of the lower-half of $\mbs{\Sigma}^{-1}$.  There are some extra non-zero entries of $\mbf{L}$, shown as $+$, that arise from completing the `four corners of a box'.  \change{The box rule is shown in light grey for the leftmost example; the bottom-right corner of the box is zero in $\mbs{\Sigma}^{-1}$ but non-zero in $\mbf{L}$}.}
\begin{tikzpicture}
\fill[white] (0,0) rectangle (1,7);
\fill[black!10!white] (2.25,0.45) rectangle (3.65,1.65);
\fill[black!10!white] (2.27,3.46) rectangle (3.67,4.62);
\end{tikzpicture}
\vspace*{-2.8in}
\begin{center}
\begin{tabular}{ccc}
\hspace{0.4in}basic sparsity constraint & \hspace{0.5in}trajectory example & \hspace{0.5in}SLAM example \\ 
\hspace{0.4in}(note fill in at $(5,3)$ in $\mbf{L}$) & \hspace{0.5in}(6 robot poses) & \hspace{0.5in}(3 poses, 3 landmarks) \\ \\
$\mbs{\Sigma}^{-1} = \bbm * & & * & & * &  \\ & * & & & &  \\ * & & * & & &  \\ & & & * & &  \\ * & & & & * &   \\ & & & & & * \ebm$ &
$\mbs{\Sigma}^{-1} = \bbm * & * &  & &  &  \\ * & * & * & & &  \\ & * & * & * & &  \\ & & * & * & * &  \\  & & & * & * & *   \\ & & & &  * & * \ebm$ &
$\mbs{\Sigma}^{-1} = \left[ \;\; \begin{matrix} * & * & &\vline &  * & *  & * \\ * & * & *&\vline & * & * & *  \\ & * & * &\vline & * &  * & *  \\ \hline * & * & * &\vline & *  &   &   \\  * & * & *&\vline &  & * &    \\ * & * & *&\vline&  &   & *\end{matrix}\;\;\right]$\\ \\
$\;\;\;\;\mbf{L} = 
\bbm * & &  & &  &  \\ & * & & & &  \\ * & & * & & &  \\ & & & * & &  \\ * & & + & & * &   \\ & & & & & * \ebm$ &
$\;\;\;\;\mbf{L} = \bbm * & &  & &  &  \\  * & * & & & &  \\  & * & * & & &  \\ & & * & * & &  \\  & &  & * & * &   \\ & & & & * & * \ebm$ &
$\;\;\;\;\mbf{L} = \left[ \;\; \begin{matrix} * & & &\vline & &  &  \\  * & * & &\vline& & &  \\  & * & *&\vline & & &  \\ \hline  * &  * & * &\vline& * & &  \\ * & * & *&\vline & + & * &   \\  * & * & *&\vline & + & + & * \end{matrix}\;\;\right]$ 
\end{tabular}

\end{center}
\label{tab:sparsity}
\vspace*{-0.2in}
\end{table*}

In general, blocks that are zero in $\mbf{L}$ will also be zero in $\mbs{\Sigma}^{-1}$, but not the other way around.  Therefore, it is sufficient (but not necessary) to calculate the blocks of $\mbf{\Sigma}$ that are non-zero in $\mbf{L}$ and it turns out this can always be done.  Table~\ref{tab:sparsity} shows some example sparsity patterns for $\mbs{\Sigma}^{-1}$ and the corresponding sparsity pattern of $\mbf{L}$.  The sparsity of the lower-half of $\mbf{L}$ is the same as the sparsity of the lower-half of $\mbs{\Sigma}^{-1}$ except that $\mbf{L}$ can have a few more non-zero entries to ensure that when multiplied together the sparsity of $\mbs{\Sigma}^{-1}$ is produced.  Specifically, if $\mbf{L}_{k,i} \neq \mbf{0}$ and $\mbf{L}_{j,i} \neq \mbf{0}$ then we must have $\mbf{L}_{j,k} \neq \mbf{0}$ \citep{erisman75}; this can be visualized as completing the `four corners of a box', as shown in the example in the first column of Table~\ref{tab:sparsity}.

Table~\ref{tab:sparsity} also shows some typical robotics examples.  In batch trajectory estimation, $\mbs{\Sigma}^{-1}$ is block-tridiagonal and in this case the $\mbf{L}$ matrix requires no extra non-zero entries.  In \ac{SLAM}, $\mbf{\Sigma}^{-1}$ is an `arrowhead' matrix with the upper-left partition (corresponding to the robot's trajectory) as block-tridiagonal and the lower-right partition (corresponding to landmarks) as block-diagonal.  Using an $\mbf{L} \mbf{D} \mbf{L}^T$ decomposition, we can exploit the sparsity of the upper-left partition, as shown in the example.  If we wanted to exploit the sparsity of the lower-right, we could reverse the order of the variables or do a $\mbf{L}^T \mbf{D} \mbf{L}$ decomposition instead.  In this \ac{SLAM} example, each of the three landmarks is observed from each of the three poses so the upper-right and lower-left partitions are dense and this causes some extra entries of $\mbf{L}$ to be non-zero.  

Finally, to understand why we do not need to calculate all of the blocks of $\mbs{\Sigma}$, we follow the explanation of \citet{erisman75}.  We aim to compute all the blocks of the lower-half of $\mbs{\Sigma}$ corresponding to the non-zero blocks of $\mbf{L}$.  Looking to equation~\eqref{eq:tak1}, we see that if $\mbf{L}_{p,k}$ is non-zero, then we require $\mbs{\Sigma}_{j,p}$ for the calculation of non-zero block $\mbs{\Sigma}_{j,k}$.  But if $\mbs{\Sigma}_{j,k}$ is non-zero, so must be $\mbf{L}_{j,k}$ and then using our `four corners of a box' rule, this implies $\mbf{L}_{j,p}$ must be non-zero and so we will have $\mbs{\Sigma}_{j,p}$ and $\mbs{\Sigma}_{p,j} = \mbs{\Sigma}_{j,p}^T$ on our list of blocks to compute already.  This shows the calculation of the desired blocks is closed under the scheme defined by~\eqref{eq:tak1}, which in turn implies there will always exist an efficient algorithm to calculate the blocks of $\mbf{\Sigma}$ corresponding to the non-zero blocks of $\mbs{\Sigma}^{-1}$, plus a few more according to the `four corners of a box' rule.

It is worth noting that variable reordering and other schemes such as Givens rotations \citep{golub96} can be combined with the Takahashi et al. approach to maximize the benefit of sparsity in $\mbs{\Sigma}^{-1}$ \citep{kaess08}.  In this section, we have simply shown that in general, the calculation of the required blocks of $\mbs{\Sigma}$ (corresponding to the non-zero block of $\mbs{\Sigma}^{-1}$) can be piggybacked efficiently onto the solution of~\eqref{eq:iterstein2}, with the details depending on the specific problem.  In fact, the bottleneck in terms of computational complexity is the original lower-diagonal-upper decomposition, which is typically required even for \ac{MAP} approaches.  We therefore claim that our \ac{ESGVI} approach has the same order of computational cost (as a function of the state size, $N$) as \ac{MAP} for a given problem, but will have a higher coefficient due to the extra burden of using the marginals to compute expectations.

\subsection{Marginal Sampling}

We have seen in the previous section that we actually only need to calculate the marginal expectations (for each factor),
\begin{gather}
\underbrace{\mathbb{E}_{q_k}[\phi_k( \mbf{x}_k )]}_{\mbox{scalar}}, \quad \underbrace{\mathbb{E}_{q_k} \left[ \frac{\partial}{\partial \mbf{x}_k^T} \phi_k( \mbf{x}_k )\right]}_{\mbox{column}}, \nonumber \\ \underbrace{\mathbb{E}_{q_k} \left[ \frac{\partial^2}{\partial \mbf{x}_k^T \partial \mbf{x}_k} \phi_k( \mbf{x}_k )\right]}_{\mbox{matrix}}, \label{eq:margexpectations}
\end{gather}
which can then be reassembled back into the larger expectations of~\eqref{eq:expectations}. 

\change{As a quick aside, an additional use for the `scalar' expression above is to evaluate the loss functional,
\begin{equation}
\label{eq:computeV}
V(q) = \sum_{k=1}^K \underbrace{\mathbb{E}_{q_k}[\phi_k( \mbf{x}_k )]}_{V_k(q_k)} + \underbrace{\frac{1}{2} \ln \left(| \mbs{\Sigma}^{-1} | \right)}_{V_0},
\end{equation}
which is used to test for convergence and to perform backtracking during optimization.  The $V_0$ term can be evaluated by noting that,
\begin{equation}
\ln\left(| \mbs{\Sigma}^{-1} |\right) = \ln\left(| \mbf{L} \mbf{D} \mbf{L}^T| \right) = \ln\left(|\mbf{D}| \right)= \sum_{i=1}^N \ln(d_{ii}),
\end{equation}
where $d_{ii}$ are the diagonal elements of $\mbf{D}$; this exploits the lower-diagonal-upper decomposition from~\eqref{eq:LDL} that we already compute during the optimization.}

The computation of each \change{expectation } in \eqref{eq:margexpectations} looks, on the surface, rather intimidating. The first and second derivatives suggests each factor must be twice differentiable, and somehow the expectation over $q_k(\mbf{x}_k)$ must be computed. So far we have made no assumptions on the specific form of the factors $\phi_k$, and we would like to keep it that way, avoiding the imposition of differentiability requirements. Additionally, recalling how sampling-based filters, such as the unscented Kalman filter \citep{julier96}, the cubature Kalman filter \citep{Arasaratnam09}, and the Gauss-Hermite Kalman filter \citep{ItoXiong00}\citep{Wu06}, approximate terms involving expectations, a cubature approximation of the associated expectations in \eqref{eq:margexpectations} appears appropriate. This section considers the use of  Stein's lemma and cubature methods to derive an alternative means to compute the terms in \eqref{eq:margexpectations} that is derivative-free. 

To avoid the need to compute derivatives of $\phi_k$, we can once again apply Stein's lemma, but in the opposite direction from our previous use.  Using~\eqref{eq:stein1} we have
\begin{equation}\label{eq:margexpectation1}
\mathbb{E}_{q_k} \left[ \frac{\partial}{\partial \mbf{x}_k^T} \phi_k( \mbf{x}_k )\right] = \mbs{\Sigma}_{kk}^{-1} \mathbb{E}_{q_k}[ (\mbf{x}_k - \mbs{\mu}_k) \phi_k( \mbf{x}_k )],
\end{equation}
and using~\eqref{eq:stein2} we have
\begin{multline}\label{eq:margexpectation2}
\mathbb{E}_{q_k} \left[ \frac{\partial^2}{\partial \mbf{x}_k^T \partial \mbf{x}_k} \phi_k( \mbf{x}_k )\right] \\ =  \mbs{\Sigma}_{kk}^{-1} \mathbb{E}_{q_k}[ (\mbf{x}_k - \mbs{\mu}_k) (\mbf{x}_k - \mbs{\mu}_k)^T \phi_k( \mbf{x}_k )] \mbs{\Sigma}_{kk}^{-1} \\ - \; \mbs{\Sigma}_{kk}^{-1} \mathbb{E}_{q_k}[\phi_k( \mbf{x}_k )].
\end{multline}
Thus, an alternative means to computing the three expectations in \eqref{eq:margexpectations}, without explicit computation of derivatives, involves first computing
\begin{gather}
\underbrace{\mathbb{E}_{q_k}[\phi_k( \mbf{x}_k )]}_{\mbox{scalar}}, \quad \underbrace{\mathbb{E}_{q_k} \left[ (\mbf{x}_k - \mbs{\mu}_k) \phi_k( \mbf{x}_k )\right]}_{\mbox{column}}, \nonumber \\ \underbrace{\mathbb{E}_{q_k} \left[ (\mbf{x}_k - \mbs{\mu}_k)(\mbf{x}_k - \mbs{\mu}_k)^T\phi_k( \mbf{x}_k )\right]}_{\mbox{matrix}}, \label{eq:margexpectations3}
\end{gather}
then computing \eqref{eq:margexpectation1} and \eqref{eq:margexpectation2} using the results of \eqref{eq:margexpectations3}.
The reverse application of Stein's lemma has not destroyed the sparsity that we unveiled earlier because we have now applied it at the marginal level, not the global level. 

Of interest next is how to actually compute the three expectations given in \eqref{eq:margexpectations3} in an efficient yet accurate way.  As integrals, the  expectations in \eqref{eq:margexpectations3} are 
\begin{subequations}\label{eq:expectations_to_approx}
\begin{eqnarray}
&& \hspace*{-0.4in}	\mathbb{E}_{q_k}[\phi_k( \mbf{x}_k )]  \\ & = & \int_{-\mbs{\infty}}^{\mbs{\infty}} \phi_k( \mbf{x}_k ) q_k(\mbf{x}_k) d \mbf{x}_k , \nonumber\\
&& \hspace*{-0.4in}	\mathbb{E}_{q_k} \left[ (\mbf{x}_k - \mbs{\mu}_k) \phi_k( \mbf{x}_k )\right]  \\ & = & \int_{-\mbs{\infty}}^{\mbs{\infty}} (\mbf{x}_k - \mbs{\mu}_k) \phi_k( \mbf{x}_k ) q_k(\mbf{x}_k) d \mbf{x}_k , \nonumber \\
&& \hspace*{-0.4in}	\mathbb{E}_{q_k} \left[ (\mbf{x}_k - \mbs{\mu}_k)(\mbf{x}_k - \mbs{\mu}_k)^T\phi_k( \mbf{x}_k )\right]  \\ & = & \int_{-\mbs{\infty}}^{\mbs{\infty}} (\mbf{x}_k - \mbs{\mu}_k) (\mbf{x}_k - \mbs{\mu}_k)^T \phi_k( \mbf{x}_k ) q_k(\mbf{x}_k) d \mbf{x}_k , \nonumber
\end{eqnarray}
\end{subequations}
where $q_k(\mbf{x}_k) = \mathcal{N}( \mbs{\mu}_k , \mbs{\Sigma}_{kk})$. 
Computing these integrals analytically is generally not possible, and as such, a numerical approximation is sought. 
There are many ways of approximating the integrals in \eqref{eq:expectations_to_approx}, the most popular type being multi-dimensional {\em Gaussian quadrature}, commonly referred to as Gaussian cubature or simply {\em cubature}  \citep{Cools97}\citep{Sarmavuori12}\citep{kokkala16}\citep{Sarkka16}\citep[\S6, p. 100]{sarkka13}. Using cubature, each of the integrals in \eqref{eq:expectations_to_approx} is approximated as \citep{kokkala16}\citep{Sarkka16}\citep[\S6, p. 99-106]{sarkka13}
\begin{subequations}\label{eq:sampledmargexpectations}
\begin{eqnarray}
& & \hspace*{-0.4in} \mathbb{E}_{q_k}[\phi_k( \mbf{x}_k )] \\ & \approx & \sum_{\ell=1}^L w_{k,\ell} \, \phi_k( \mbf{x}_{k,\ell} ), \nonumber \\
& & \hspace*{-0.4in}\mathbb{E}_{q_k}[ (\mbf{x}_k - \mbs{\mu}_k) \phi_k( \mbf{x}_k )] \\ & \approx & \sum_{\ell=1}^L w_{k,\ell} \, (\mbf{x}_{k,\ell} - \mbs{\mu}_k) \phi_k( \mbf{x}_{k,\ell} ),  \nonumber \\
& & \hspace*{-0.4in}\mathbb{E}_{q_k}[ (\mbf{x}_k - \mbs{\mu}_k)(\mbf{x}_k - \mbs{\mu}_k)^T \phi_k( \mbf{x}_k )] \\ & \approx & \sum_{\ell=1}^L w_{k,\ell} \, (\mbf{x}_{k,\ell} - \mbs{\mu}_k)(\mbf{x}_{k,\ell} - \mbs{\mu}_k)^T \phi_k( \mbf{x}_{k,\ell} ), \nonumber
\end{eqnarray}
\end{subequations}
where $w_{k,\ell}$ are weights,  $\mbf{x}_{k,\ell} = \mbs{\mu}_k + \sqrt{\mbs{\Sigma}_{kk}} \mbs{\xi}_{k,\ell} $ are sigmapoints, and $\mbs{\xi}_{k,\ell}$ are unit sigmapoints. Both the weights and unit sigmapoints are specific to the cubature method. For example, the popular unscented transformation \citep{julier96}\citep{Sarkka16}\citep[\S6, p. 109-110]{sarkka13} uses weights
\begin{equation}
 	w_{k,0} = \frac{\kappa}{N_k + \kappa} , \quad w_{k,\ell} = \frac{1}{2(N_k + \kappa)} , \quad \ell = 1 , \ldots , 2 N_k 
\end{equation}
and sigmapoints
\begin{equation}
	\mbs{\xi}_{k,\ell}
	=
	\left\{
	\begin{array}{ll}
		\mbf{0}  & \ell = 0  \\
		\sqrt{N_k + \kappa} \mbf{1}_\ell  & \ell = 1 , \ldots , N_k   \\
		-\sqrt{N_k + \kappa} \mbf{1}_{\ell - N_k}  & \ell = N_k + 1 , \ldots , 2 N_k  
	\end{array}
	\right. ,
\end{equation}
where $N_k$ is the dimension of $\mbf{x}_k$.  On the other hand, the spherical-cubature rule \citep{Arasaratnam09}\citep{kokkala16}\citep[\S6, p. 106-109]{sarkka13} uses weights 
\begin{equation} 
	w_{k,\ell} = \frac{1}{2 N_k} , \quad  \ell = 1, \ldots ,  2 N_k , 
\end{equation}
and sigmapoints
\begin{equation}
	\mbs{\xi}_{k,\ell}
	=
	\left\{
	\begin{array}{ll}
		\sqrt{N_k} \mbf{1}_\ell  & \ell = 1 , \ldots , N_k   \\
		-\sqrt{N_k} \mbf{1}_{\ell - N_k}  & \ell = N_k + 1 , \ldots , 2 N_k  
	\end{array}
	\right. ,
\end{equation} %
where $\mbf{1}_i$ is a $N_k \times 1$ column matrix with $1$ at row $i$ and zeros everywhere else. Gauss-Hermite cubature is yet another  method that can be used to compute the approximations in \eqref{eq:sampledmargexpectations} \citep{ItoXiong00}\citep{Wu06}\citep[\S6 p. 99-106]{sarkka13}. As discussed in \citet[\S6 p. 103]{sarkka13}, given an integrand composed of a linear combination of monomials of the form $x_1^{d_1}, x_2^{d_2}, \ldots , x_{N_k}^{d_{N_k}} $, the $M$th order Gauss-Hermite cubature rule is exact when $d_i \leq 2 M - 1$. However, for an $M$th-order Gauss-Hermite cubature approximation, $M^{N_k}$ sigmapoints are needed, which could be infeasible in practise when $N_k$ is large \citep[\S6 p. 103]{sarkka13}. Fortunately, the approximations of \eqref{eq:expectations_to_approx} given in \eqref{eq:sampledmargexpectations} are at the factor level (i.e., at the level of $\mbf{x}_k$, not $\mbf{x}$), and at the factor level $N_k$ is often a manageable size in most robotics problems.  For this reason, Gauss-Hermite cubature is used in our numerical work presented in Sections \ref{sec:1dstereoexp}, \ref{sec:stereoKexp}, and \ref{sec:robexp}, yielding accurate yet reasonably efficient approximations of \eqref{eq:expectations_to_approx}.

Some additional remarks are as follows: 
\begin{enumerate}

\item The accuracy of the \change{cubature } approximations in \eqref{eq:sampledmargexpectations} will depend on the specific method and the severity of the nonlinearity in $\phi_k$. Alternative means to approximate \eqref{eq:sampledmargexpectations}, such as cubature methods that are exact for specific algebraic and trigonometric polynomials \citep{Cools97}\citep{kokkala16}, Gaussian-process cubature \citep{ohagan90}\citep{Sarkka16}, or even adaptive cubature methods \citep[\S4, p. 194]{numrec07}, can be employed. In the case where computational complexity is of concern, a high-degree cubature rule that is an efficient alternative to Gauss-Hermite cubature is presented in \citet{Jia13}. 

\item We are proposing quite a different way of using a cubature method (or any sampling method) than is typical in the state estimation literature; we consider the entire factor expression, $\phi_k$, to be the nonlinearity, not just the observation or motion models, as is common.  This means, for example, that if there is a robust cost function incorporated in our factor expression \citep[\S5, p. 163]{barfoot17}\citep{mactavish_crv15}, it is handled automatically and does not need to be implemented as iteratively reweighted least squares \citep{holland77}.  

\item Because we have `undone' Stein's lemma at this point (it was a temporary step to exploit the sparsity only), it may not even be necessary to have $\phi_k$ differentiable anymore.  \change{Appendix~\ref{sec:AppDirectESGVI} shows how to get to the derivative-free version of ESGVI directly without the double application of Stein's lemma. }  This opens the door to some interesting possibilities including the use of the $H_\infty$ (worst case) norm, hard constraints on some or all of the states, or the aforementioned use of a robust cost function, within the factor $\phi_k$.  An appropriate sampling method would be required.

\item We see in~\eqref{eq:sampledmargexpectations} that the scalars, $\phi_k$, serve to reweight each sample, but that otherwise the expressions are simply those for the first three moments of a distribution.

\item \change{We also use cubature to evaluate $V(q)$ according to~\eqref{eq:computeV}.  This is required to test for convergence of the optimization scheme.}

\end{enumerate}

The approach that we have presented up to this point is extremely general and can benefit any \ac{GVI} problem where $p(\mbf{x},\mbf{z})$ \change{can be factored}.  In computer vision and robotics, some examples include \ac{BA} \citep{brown58} and \ac{SLAM} \citep{durrantwhyte06a}.  In Section~\ref{sec:evaluation}, we will demonstrate the technique first on controlled toy problems, then on a batch \ac{SLAM} problem.

\section{Extensions}

\label{sec:extensions}

\change{Before moving on to our experiments,  we pause to elaborate two extensions of the main paper.   First, we discuss an alternate loss functional that leads to a modified optimization scheme similar to a Gauss-Newton solver and offers some computational savings.  Second, we discuss how to extend our approach beyond estimation of the latent state to include estimation of unknown parameters in our models.}

\subsection{Alternate Loss Functional}
\label{sec:alternateloss}

We can consider an alternate variational problem that may offer computational advantages over the main \ac{ESGVI} approach of this paper.  We consider the special case where the negative-log-likelihood takes the form
\begin{equation}
\phi(\mbf{x}) = \frac{1}{2} \mbf{e}(\mbf{x})^T \mbf{W}^{-1}  \mbf{e}(\mbf{x}).
\end{equation}
Substituting this into the loss functional, we have
\begin{equation}
V(q) = \frac{1}{2} \mathbb{E}_q \left[ \mbf{e}(\mbf{x})^T \mbf{W}^{-1}  \mbf{e}(\mbf{x}) \right] + \frac{1}{2} \ln( | \mbs{\Sigma}^{-1}| ). 
\end{equation}
Owing to the convexity of the quadratic expression, $\mbf{e}^T\mbf{W}^{-1}\mbf{e}$, we can apply {\em Jensen's inequality} \citep{jensen06} directly to write
\begin{equation}
\mathbb{E}_q [ \mbf{e}(\mbf{x})]^T \mbf{W}^{-1} \mathbb{E}_q [ \mbf{e}(\mbf{x}) ] \leq \mathbb{E}_q \left[ \mbf{e}(\mbf{x})^T \mbf{W}^{-1}  \mbf{e}(\mbf{x}) \right].
\end{equation}
The {\em Jensen gap} is the (positive) difference between the right and left sides of this inequality and will generally tend to be larger the more nonlinear is $\mbf{e}(\mbf{x})$ and less concentrated is $q(\mbf{x})$.  Motivated by this relationship, we can define a new loss functional as
\begin{equation}
V^\prime(q) = \frac{1}{2} \mathbb{E}_q [ \mbf{e}(\mbf{x})]^T \mbf{W}^{-1} \mathbb{E}_q [ \mbf{e}(\mbf{x}) ] + \frac{1}{2} \ln( | \mbs{\Sigma}^{-1}| ),
\end{equation}
which may be thought of as a (conservative) approximation of $V(q)$ that is appropriate for mild nonlinearities and/or concentrated posteriors; the conservative aspect will be discussed a bit later on.  We will now show that we can minimize $V^\prime(q)$ by iteratively updating $q(\mbf{x})$ and continue to exploit problem sparsity arising from a factored likelihood.

We begin by noting that we can directly approximate the expected error as
\begin{eqnarray}
& & \hspace*{-0.5in} \mathbb{E}_{q^{(i+1)}} [ \mbf{e}(\mbf{x})] \nonumber \\ 
& \approx & \mathbb{E}_{q^{(i)}} [ \mbf{e}(\mbf{x})]  + \frac{\partial}{\partial \mbs{\mu}} \mathbb{E}_{q^{(i)}} [ \mbf{e}(\mbf{x})] \, \underbrace{\left( \mbs{\mu}^{(i+1)} - \mbs{\mu}^{(i)} \right)}_{\delta\mbs{\mu}} \nonumber \\ & = & \underbrace{\mathbb{E}_{q^{(i)}} [ \mbf{e}(\mbf{x})]}_{\bar{\mbf{e}}^{(i)}} +  \underbrace{\mathbb{E}_{q^{(i)}} \left[ \frac{\partial}{\partial \mbf{x}}\mbf{e}(\mbf{x})\right]}_{\bar{\mbf{E}}^{(i)}} \, \delta\mbs{\mu} \nonumber \\ & = & \bar{\mbf{e}}^{(i)} + \bar{\mbf{E}}^{(i)} \, \delta\mbs{\mu},
\end{eqnarray}
where we have employed \change{the derivative identity in}~\eqref{eq:derivident1}.

We can then approximate the loss functional as
\begin{multline}
V^\prime(q) \approx \frac{1}{2} \left( \bar{\mbf{e}}^{(i)} + \bar{\mbf{E}}^{(i)} \, \delta\mbs{\mu}\right)^T \mbf{W}^{-1} \left( \bar{\mbf{e}}^{(i)} + \bar{\mbf{E}}^{(i)} \, \delta\mbs{\mu} \right) \\ + \frac{1}{2} \ln( | \mbs{\Sigma}^{-1}| ),
\end{multline}
which is now exactly quadratic in $\delta\mbs{\mu}$.  This specific approximation leads directly to a Gauss-Newton estimator, bypassing Newton's method, as we have implicitly approximated the Hessian \citep[p.131]{barfoot17}.  Taking the first and second derivatives with respect to  $\delta\mbs{\mu}$, we have
\begin{subequations}
\begin{eqnarray}
\frac{\partial V^\prime(q)}{\partial \, \delta\mbs{\mu}^T} & = & \bar{\mbf{E}}^{(i)^T} \mbf{W}^{-1}  \left( \bar{\mbf{e}}^{(i)} + \bar{\mbf{E}}^{(i)} \, \delta\mbs{\mu}\right), \quad \label{eq:meanGN} \\
\frac{\partial^2 V^\prime(q)}{\partial \, \delta\mbs{\mu}^T \partial \, \delta\mbs{\mu}} & = & \bar{\mbf{E}}^{(i)^T} \mbf{W}^{-1} \bar{\mbf{E}}^{(i)} .
\end{eqnarray}
\end{subequations}
For the derivative with respect to $\mbs{\Sigma}^{-1}$, we have
\begin{equation}
\frac{\partial V^\prime(q)}{\partial \mbs{\Sigma}^{-1}} \approx -\frac{1}{2} \mbs{\Sigma} \,\bar{\mbf{E}}^{(i)^T} \mbf{W}^{-1} \bar{\mbf{E}}^{(i)} \, \mbs{\Sigma} + \frac{1}{2} \mbs{\Sigma},
\end{equation}
where the approximation enforces the relationship in~\eqref{eq:derivident2}, which does not hold exactly anymore due to the altered nature of $V^\prime(q)$.  Setting this to zero for a critical point we have
\begin{equation}
(\mbs{\Sigma}^{-1})^{(i+1)} = \bar{\mbf{E}}^{(i)^T} \mbf{W}^{-1} \bar{\mbf{E}}^{(i)},
\end{equation}
where we have created an iterative update analogous to that in the main \ac{ESGVI} approach.

For the mean, we set~\eqref{eq:meanGN} to zero and then for the optimal update we have
\begin{equation}
\underbrace{\bar{\mbf{E}}^{(i)^T} \mbf{W}^{-1} \bar{\mbf{E}}^{(i)}}_{(\mbs{\Sigma}^{-1})^{(i+1)}} \, \delta\mbs{\mu} = -\bar{\mbf{E}}^{(i)^T} \mbf{W}^{-1}\bar{\mbf{e}}^{(i)}.
\end{equation}
Solving for $\delta\mbs{\mu}$ provides a Gauss-Newton update, which we will refer to as \ac{ESGVI-GN}.  This is identical to how Gauss-Newton is normally carried out, but now we calculate $\bar{\mbf{e}}$ and $\bar{\mbf{E}}$ not just at a single point but rather as an expectation over our Gaussian posterior estimate.  We again make a number of remarks about the approach:
\begin{enumerate}
\item The sparsity of the inverse covariance matrix, $\mbs{\Sigma}^{-1}$, will be identical to the full \ac{ESGVI} approach.  This can be seen by noting that
\begin{multline}
\phi(\mbf{x}) =  \sum_{k=1}^K \phi_k( \mbf{x}_k ) = \frac{1}{2}\sum_{k=1}^K \mbf{e}_k( \mbf{x}_k )^T \mbf{W}_k^{-1} \mbf{e}_k( \mbf{x}_k ) \\ = \frac{1}{2} \mbf{e}(\mbf{x})^T \mbf{W}^{-1}  \mbf{e}(\mbf{x}),
\end{multline}
where
\begin{equation}
\mbf{e}(\mbf{x}) = \bbm \mbf{e}_1(\mbf{x}_1) \\ \vdots \\ \mbf{e}_K(\mbf{x}_K)\ebm, \quad \mbf{W} = \mbox{diag}(\mbf{W}_1, \ldots, \mbf{W}_K).
\end{equation}
Then we have
\begin{multline}
\mbs{\Sigma}^{-1} = \mathbb{E}_q \left[ \frac{\partial}{\partial \mbf{x}}\mbf{e}(\mbf{x})\right]^T \mbf{W}^{-1}  \mathbb{E}_q \left[ \frac{\partial}{\partial \mbf{x}}\mbf{e}(\mbf{x})\right] \\ = \sum_{k=1}^K \mbf{P}_k^T \mathbb{E}_{q_k} \left[\frac{\partial}{\partial \mbf{x}_k} \mbf{e}_k(\mbf{x}_k) \right]^T \mbf{W}_k^{-1} \\ \times \; \mathbb{E}_{q_k} \left[\frac{\partial}{\partial \mbf{x}_k} \mbf{e}_k(\mbf{x}_k) \right] \mbf{P}_k,
\end{multline}
which will have zeros wherever an error term does not depend on the variables.  We also see, just as before, that the expectations can be reduced to being over the marginal, $q_k(\mbf{x}_k)$, meaning we still only require the blocks of $\mbf{\Sigma}$ corresponding to the non-zero blocks of $\mbs{\Sigma}^{-1}$.

\item We can still use Stein's lemma to avoid the need to compute any derivatives:
\begin{multline}
 \mathbb{E}_{q_k} \left[\frac{\partial}{\partial \mbf{x}_k} \mbf{e}_k(\mbf{x}_k) \right] \\ =  \mathbb{E}_{q_k} \left[  \mbf{e}_k(\mbf{x}_k) (\mbf{x}_k -\mbs{\mu}_k)^T \right] \mbs{\Sigma}_{kk}^{-1}. 
\end{multline}
This is sometimes referred to as a statistical Jacobian and this usage is very similar to the filtering and smoothing approaches described by \citet{sarkka13}, amongst others, as cubature can be applied at the measurement model level rather than the factor level.  Because we are iteratively recomputing the statistical Jacobian about our posterior estimate, this is most similar to \citet{sibley06} and \citet{garcia15}, although some details are different as well as the fact that we started from our loss functional, $V^\prime(q)$.

\item The number of cubature points required to calculate $\mathbb{E}_{q_k} \left[  \mbf{e}_k(\mbf{x}_k) (\mbf{x}_k -\mbs{\mu}_k)^T \right]$ will be lower than our full \ac{ESGVI} approach described earlier as the order of the expression in the integrand is half that of $\mathbb{E}_{q_k} \left[  (\mbf{x}_k -\mbs{\mu}_k)(\mbf{x}_k -\mbs{\mu}_k)^T \phi_k(\mbf{x}_k) \right]$.  Since the number of cubature points goes up as $M^{N_k}$, cutting $M$ in half is significant and could be the difference between tractable and not for some problems.  This was the main motivation for exploring this alternate approach.

\item It is known that minimizing $\mbox{KL}(q||p)$, which our $V(q)$ is effectively doing, can result in a Gaussian that is too confident (i.e., inverse covariance is too large) \citep{bishop06,alaluhtala15}.  A side benefit of switching from $V(q)$ to $V^\prime(q)$ is that the resulting inverse covariance will be more conservative.  This follows from Jensen's inequality once again.  For an arbitrary non-zero vector, $\mbf{a}$, we have
\begin{multline}
0 \; < \; \mbf{a}^T   \underbrace{\mathbb{E}_q \left[\frac{\partial \mbf{e}(\mbf{x})}{\partial \mbf{x}}\right]^T \mbf{W}^{-1}  \mathbb{E}_q \left[ \frac{\partial \mbf{e}(\mbf{x})}{\partial \mbf{x}}\right]}_{\mbox{$\mbs{\Sigma}^{-1}$ from $V^\prime(q)$}}  \mbf{a} \;\;  \\ \stackrel{\mbox{Jensen}}{\leq} \;\;\; \mbf{a}^T \mathbb{E}_q \left[\frac{\partial \mbf{e}(\mbf{x})}{\partial \mbf{x}}^T \mbf{W}^{-1} \frac{\partial \mbf{e}(\mbf{x})}{\partial \mbf{x}} \right] \mbf{a} \;\; \\ \stackrel{\mbox{Gauss-Newton}}{\approx} \;\; \mbf{a}^T \underbrace{\mathbb{E}_q \left[\frac{\partial^2 \phi(\mbf{x})}{\partial \mbf{x}^T \partial \mbf{x}}  \right]}_{\mbox{$\mbs{\Sigma}^{-1}$ from $V(q)$}} \mbf{a}, 
\end{multline}
which ensures that not only do we have a positive definite inverse covariance but that it is conservative compared to the full \ac{ESGVI} approach.
\end{enumerate}
Due to the extra approximations made in \ac{ESGVI-GN} compared to \ac{ESGVI}, it remains to be seen whether it \change{offers an improvement } over \ac{MAP} approaches.  However, as  \ac{ESGVI-GN} provides a batch option that does not require any derivatives, it can be used as a less expensive preprocessor for the derivative-free version of full \ac{ESGVI}.

\subsection{Parameter Estimation}
\label{sec:paramest}

Although it is not the main focus of our paper, we use this section to provide a sketch of how parameters may also be estimated using our \ac{ESGVI} framework.  We introduce some unknown parameters, $\mbs{\theta}$, to our loss functional,
\begin{equation}
V(q | \mbs{\theta}) = \mathbb{E}_q[ \phi(\mbf{x} | \mbs{\theta})] + \frac{1}{2} \ln( | \mbs{\Sigma}^{-1} | ),
\end{equation}
and recall that $V(q|\mbs{\theta})$ is the negative of the so-called \acf{ELBO}, which can be used in an \ac{EM} framework to estimate parameters when there is a latent state \citep{neal98,ghahramani99}.  The expectation, or E-step, is already accomplished by \ac{ESGVI}; we simply hold $\mbs{\theta}$ fixed and run the inference to convergence to solve for $q(\mbf{x})$, our Gaussian approximation to the posterior.  In the M-step, which is actually a minimization in our case, we hold $q(\mbf{x})$ fixed and find the value of $\mbs{\theta}$ that minimizes the loss functional.  By alternating between the E- and M-steps, we can solve for the best value of the parameters to minimize $-\ln p(\mbf{z} | \mbs{\theta})$, the negative log-likelihood of the measurements given the parameters.  

\begin{table*}[t]
\caption{Descriptions of variants of our \ac{ESGVI} algorithm tested in our experiments.  \change{The first two algorithms in the table are \ac{MAP} comparisons, which can be recovered in our framework by evaluating all the expectations using a single quadrature point at the mean of the current estimate.  The only extra feature added to the all the methods beyond the `plain vanilla' implementation described in the theory section is that all algorithms are allowed to backtrack when updating the mean and inverse covariance in order to ensure that the loss functional actually decreases at each iteration.}}
\vspace*{-0.05in}
\begin{center}
\begin{tabular}{l|c|c}
algorithm & method to evaluate  & $M$, number of quadrature points  \\ 
label         & expectations in~\eqref{eq:margexpectations} & (per dimension) \\\hline\hline
\ac{MAP} Newton & analytical Jacobian and Hessian & 1  \\
\ac{MAP} GN & analytical Jacobian and approximate Hessian & 1  \\
\ac{ESGVI} deriv M=2 & analytical Jacobian and Hessian + quadrature & 2  \\
\ac{ESGVI} deriv M=3 &  analytical Jacobian and Hessian + quadrature & 3  \\
\ac{ESGVI} deriv-free M=3 & Stein's lemma + quadrature & 3  \\
\ac{ESGVI-GN} deriv-free M=3 & Stein's lemma + quadrature & 3  \\
\ac{ESGVI} deriv-free M=4 & Stein's lemma + quadrature & 4  \\
\ac{ESGVI} deriv-free M=10 & Stein's lemma + quadrature & 10  \\
\end{tabular}
\end{center}
\label{tab:alglabels}
\vspace*{-0.25in}
\end{table*}

As we have done in the main part of the paper, we assume the joint likelihood of the state and measurements (given the parameters) factors so that
\begin{equation}
 \phi(\mbf{x} | \mbs{\theta}) = \sum_{k=1}^K  \phi_k(\mbf{x}_k | \mbs{\theta}),
\end{equation}
where for generality we have each factor being affected by the entire parameter set, $\mbs{\theta}$, but in practice it could be a subset.  Taking the derivative of the loss functional with respect to $\mbs{\theta}$, we have
\begin{eqnarray}
\frac{\partial V(q|\mbs{\theta})}{\partial \mbs{\theta}} & = & \frac{\partial}{\partial \mbs{\theta}} \mathbb{E}_q[ \phi(\mbf{x} | \mbs{\theta})] \nonumber \\ & = & \frac{\partial}{\partial \mbs{\theta}} \mathbb{E}_q\left[ \sum_{k=1}^K  \phi_k(\mbf{x}_k | \mbs{\theta}) \right] \nonumber \\ & = & \sum_{k=1}^K \mathbb{E}_{q_k}\left[  \frac{\partial}{\partial \mbs{\theta}} \phi_k(\mbf{x}_k | \mbs{\theta}) \right],
\end{eqnarray}
where in the last expression the expectation simplifies to being over the marginal, $q_k(\mbf{x}_k)$, rather than the full Gaussian, $q(\mbf{x})$.  As with the main \ac{ESGVI} approach, this means that we only need the blocks of the covariance, $\mbs{\Sigma}$, corresponding to the non-zero blocks of $\mbs{\Sigma}^{-1}$, which we are already calculating as part of the E-step.  Furthermore, we can easily evaluate the marginal expectations using cubature.

To make this more tangible, consider the example of
\begin{equation}
 \phi(\mbf{x} | \mbf{W}) = \frac{1}{2} \sum_{k=1}^K \left( \mbf{e}_k(\mbf{x}_k)^T \mbf{W}^{-1} \mbf{e}_k(\mbf{x}_k) - \ln(|\mbf{W}^{-1}|) \right),
\end{equation}
where the unknown parameter is $\mbf{W}$, the measurement covariance matrix.  Then taking the derivative with respect to $\mbf{W}^{-1}$ we have
\begin{equation}
\frac{\partial V(q|\mbf{W})}{\partial \mbf{W}^{-1}} = \frac{1}{2} \sum_{k=1}^K \mathbb{E}_{q_k}\left[  \mbf{e}_k(\mbf{x}_k)  \mbf{e}_k(\mbf{x}_k)^T \right] - \frac{K}{2} \mbf{W}.
\end{equation}
Setting this to zero for a minimum we have
\begin{equation}
\mbf{W} = \frac{1}{K} \sum_{k=1}^K \mathbb{E}_{q_k}\left[  \mbf{e}_k(\mbf{x}_k)  \mbf{e}_k(\mbf{x}_k)^T \right],
\end{equation}
where we can use cubature to evaluate the marginal expectations.  Reiterating, we never require the full covariance matrix, $\mbs{\Sigma}$, implying that our exactly sparse framework extends to parameter estimation.

\section{Evaluation}

\label{sec:evaluation}

\change{We compared several variants of our algorithm on three different test problems:  (i) a scalar toy problem motivated by a stereo camera to show that \ac{ESGVI} achieves a better estimate than \ac{MAP}, (ii) a multi-dimensional \ac{SLAM} problem that shows we can carry out \ac{ESGVI} in a tractable way, and (iii) a \ac{SLAM} problem using a real robotics dataset that shows the new methods work well in practice.

Table~\ref{tab:alglabels} lists the variants of \ac{ESGVI} that we studied as well as two \ac{MAP} variants for comparison.  The methods differ in the way the expectations of~\eqref{eq:margexpectations} are evaluated.  `Analytical' means that we calculate derivatives in closed form whereas `derivative-free' means we used Stein's lemma to avoid derivatives.   All the methods attempt to minimize $V(q)$ (i.e., they are running our Newton-style optimizer) unless they have the `GN' (Gauss-Newton) designation, which indicates that they are using the alternate loss functional, $V^\prime(q)$, from Section~\ref{sec:alternateloss}.  All the methods use cubature to calculate the expectations, but the number of points per dimension is varied.  The \ac{MAP} methods are simply special cases of the \ac{ESGVI} approach in which we use analytical derivatives and a single quadrature point located at the mean of the current estimate.

All the methods use the `plain vanilla' optimization scheme exactly as described in the paper with the exception of one extra feature.  When updating the mean and inverse covariance, we allow backtracking if the loss functional is not reduced.  In other words, we attempt to update according to
\begin{subequations}
\begin{eqnarray}
\mbs{\mu}^{(i+1)} & \leftarrow & \mbs{\mu}^{(i)} + \alpha^B \, \delta\mbs{\mu}, \\
\mbs{\Sigma}^{-1^{(i+1)}} & \leftarrow & \mbs{\Sigma}^{-1^{(i)}} + \alpha^B \, \delta\mbs{\Sigma}^{-1},
\end{eqnarray}
\end{subequations}
where $\alpha = 0.95$ and with $B=0,1,2,\ldots$ increasing until the loss functional goes down from the previous iteration.  All the methods tested were allowed to do this in the same way and used~\eqref{eq:computeV} to calculate the loss functional.
}

\subsection{Experiment 1:  Stereo One-Dimensional Simulation}
\label{sec:1dstereoexp}

Our first simulation is a simple one-dimensional, nonlinear estimation problem motivated by the type of inverse-distance nonlinearity found in a stereo camera model.  As this problem is only one-dimensional, we cannot demonstrate the ability to exploit sparsity in the problem, leaving this to the next two subsections.  Here our aim is to show that indeed our proposed iterative scheme converges to the minimum of our loss functional and also that we offer an improvement over the usual \ac{MAP} approach.  

This same experiment (with the same parameter settings) was used as a running example by \citet[\S4]{barfoot17}.  We assume that our true state is drawn from a Gaussian prior:
\begin{equation}
x  \sim \mathcal{N}(\mu_p, \sigma_p^2).
\end{equation}
We then generate a measurement according to
\begin{equation}
y = \frac{fb}{x} + n, \qquad n \sim \mathcal{N}(0,\sigma_r^2),
\end{equation}
where $n$ is measurement noise.  The numerical values of the parameters used in our trials were
\begin{gather}
\label{eq:stereoexpparam}
\mu_p = 20 \mbox{ [m]}, \quad \sigma_p^2 = 9 \mbox{ [m$^2$]}, \\
f = 400 \mbox{ [pixel]}, \quad b = 0.1 \mbox{ [m]}, \quad \sigma_r^2 = 0.09 \mbox{ [pixel$^2$]}. \nonumber
\end{gather}
The two factors are defined as
\begin{equation}
\phi = \frac{1}{2} \frac{(x - \mu_p)^2}{\sigma_p^2}, \qquad \psi = \frac{1}{2} \frac{\left(y - \frac{fb}{x}\right)^2}{\sigma_r^2},
\end{equation}
so that $-\ln p(\mbf{x},\mbf{z}) = \phi + \psi + \mbox{constant}$.  Our loss functional is therefore
\begin{equation}
V(q) = \mathbb{E}_q[\phi] + \mathbb{E}_q[\psi] + \frac{1}{2} \ln ( \sigma^{-2}),
\end{equation}
where $q = \mathcal{N}(\mu, \sigma^2)$ is our estimate of the posterior.  
We seek to find the $q$ to minimize $V(q)$.  This problem can also be viewed as the correction step of the Bayes filter \citep{jazwinsky70}:  start from a prior and correct it based on the latest (nonlinear) measurement.

\begin{figure*}[p]
\centering
\vspace*{1.0in}
\includegraphics[width=\textwidth]{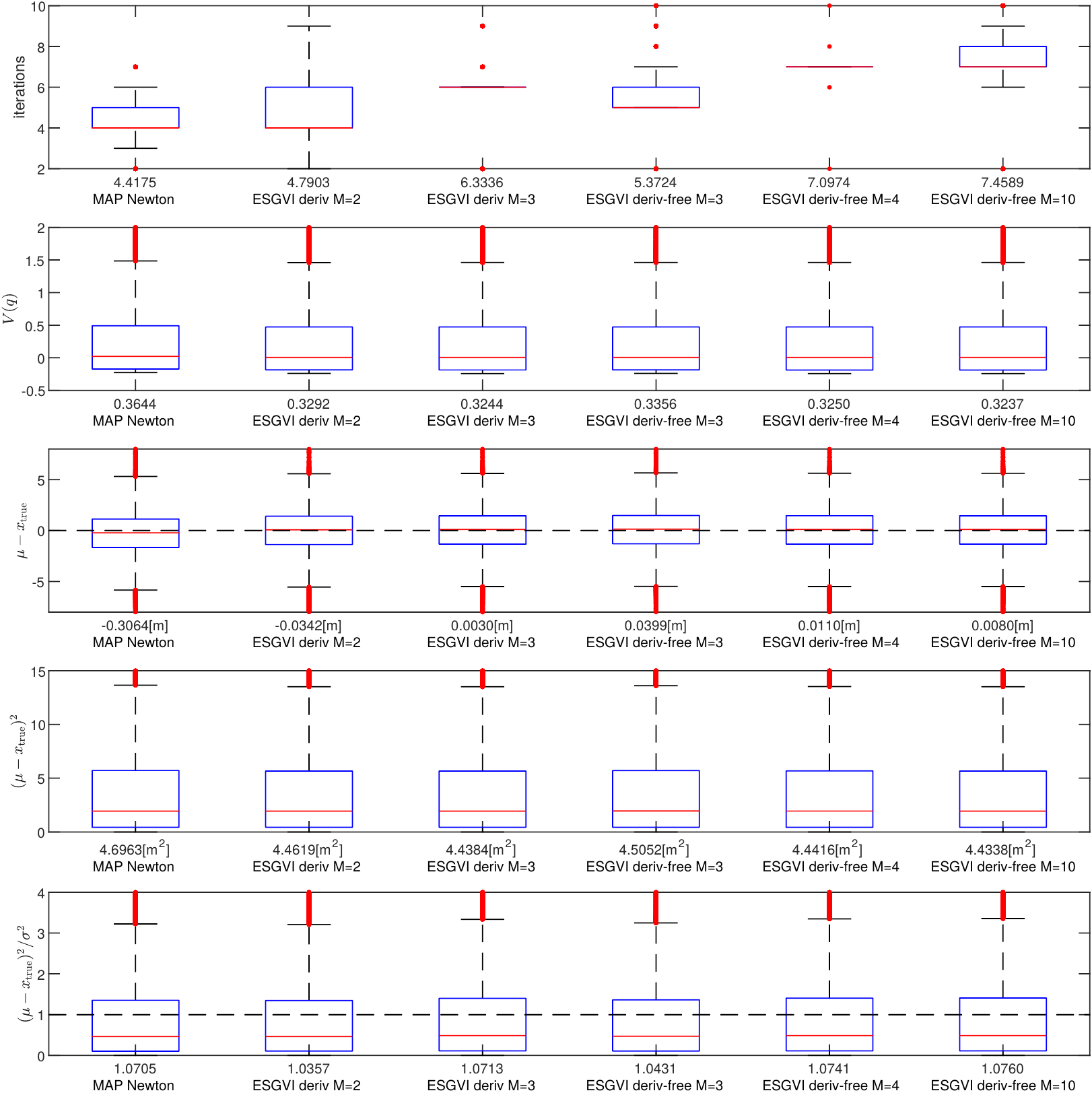}
\caption{(Experiment 1) Statistical results of $100,000$ trials of the one-dimensional stereo camera simulation shown as standard boxplots.  The different rows show different performance metrics for the different variants of our algorithm (columns).  Table~\ref{tab:alglabels} provides details of the different algorithms tested.  The number above an algorithm label is its mean performance on that metric.  Further details are discussed in the text.}
\label{fig:esgvi_exp1b}
\vspace*{1.0in}
\end{figure*}

\begin{figure*}[p]
\centering
\begin{turn}{90}\hspace*{0.8in}MAP Newton\end{turn} \hspace{0.1in}
\includegraphics[height=0.23\textheight]{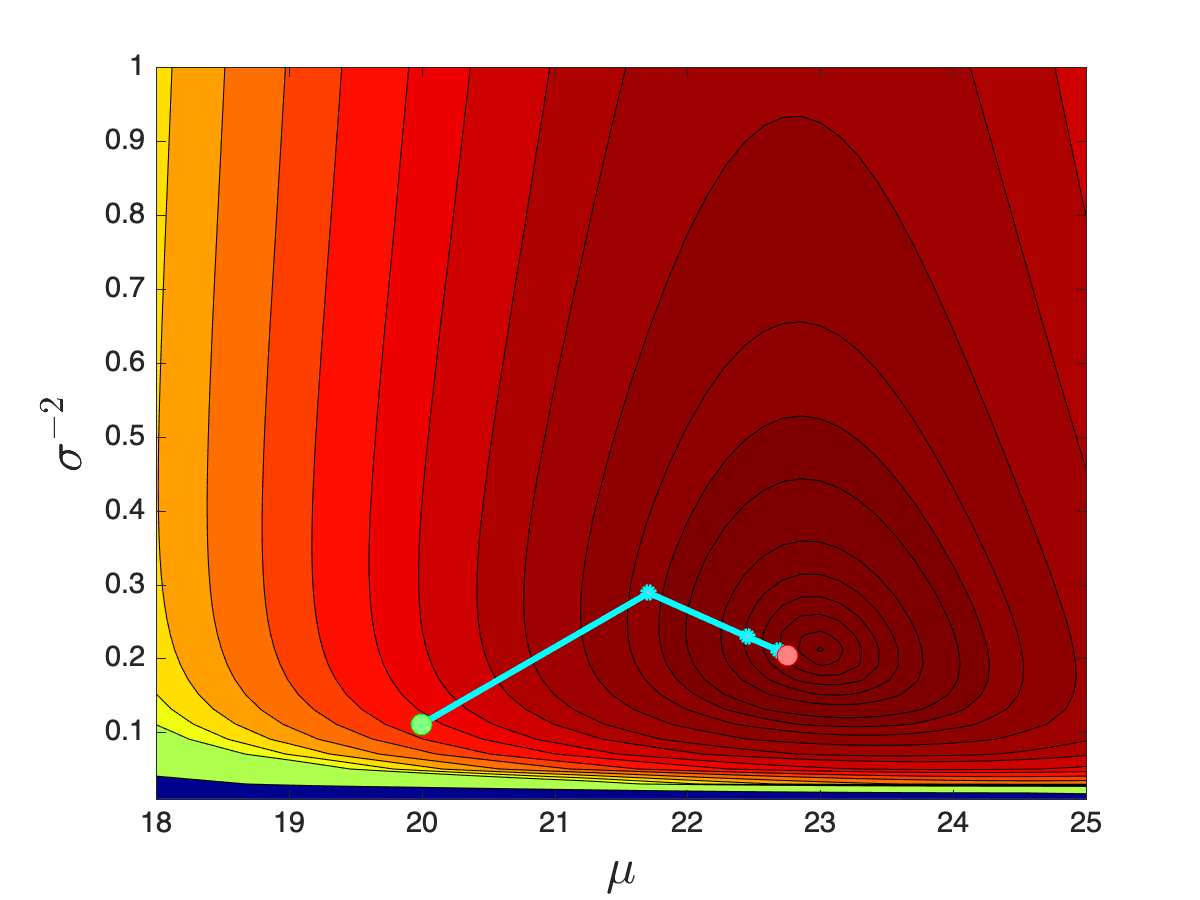} 
\includegraphics[height=0.22\textheight]{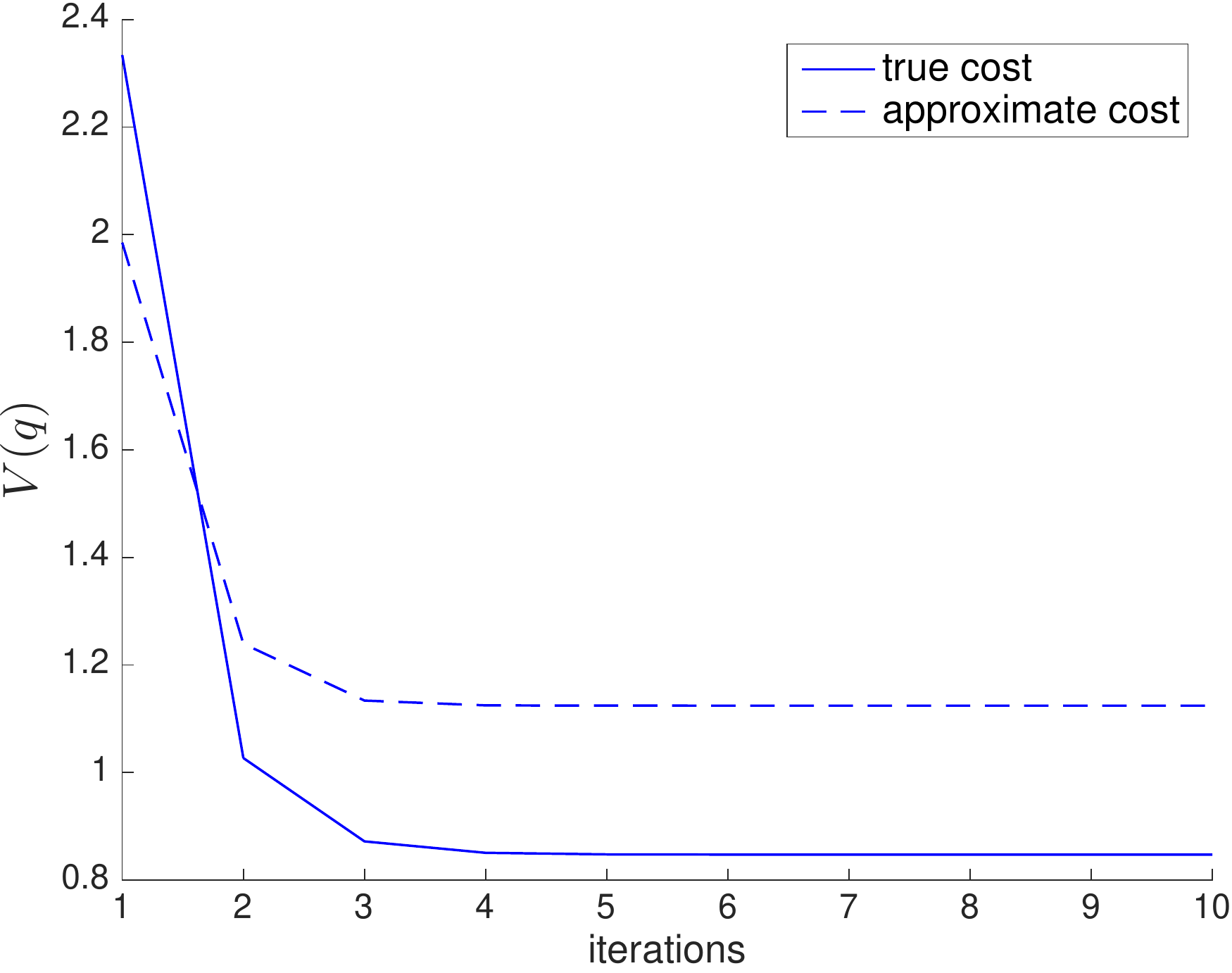} \\

\vspace*{-0.03in}
\begin{turn}{90}\hspace*{0.7in}ESGVI deriv M=2\end{turn} \hspace{0.1in}
\includegraphics[height=0.23\textheight]{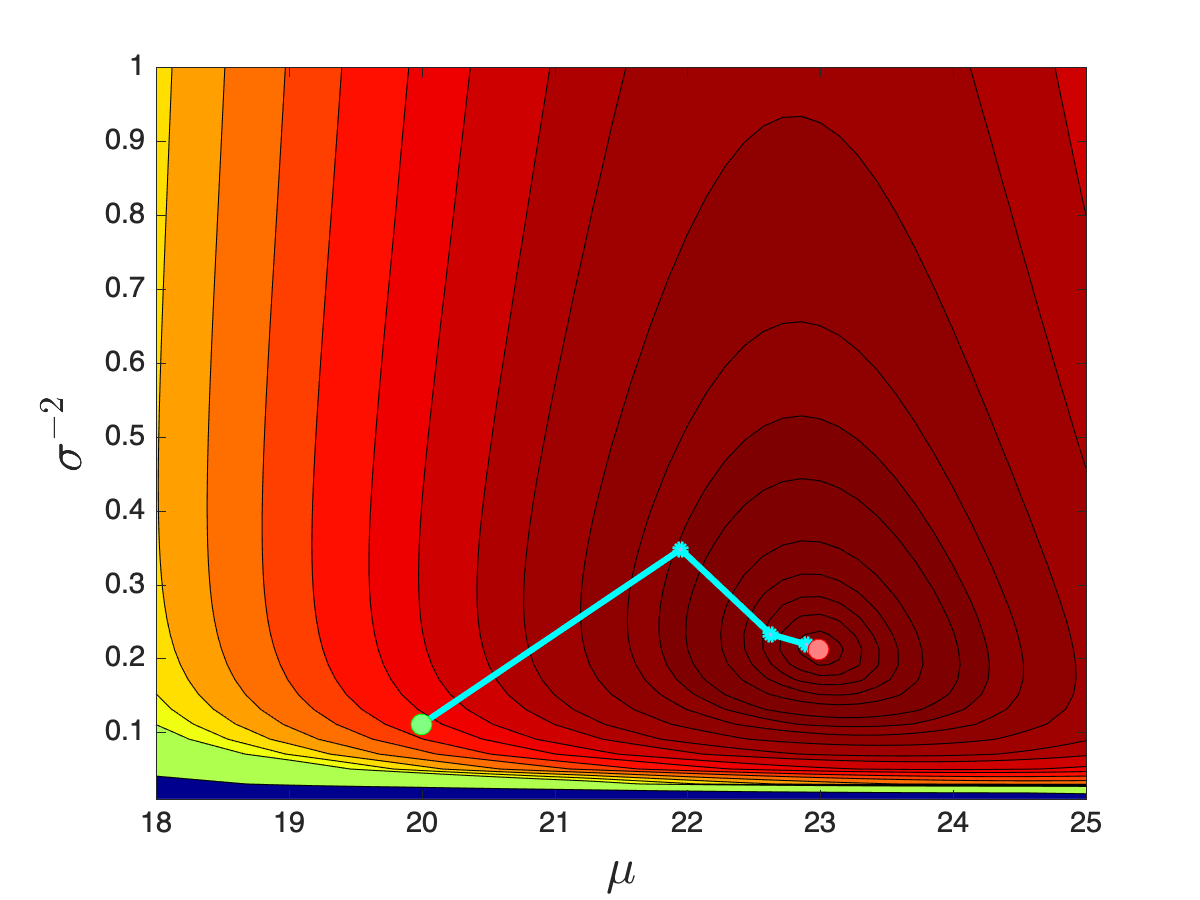} 
\includegraphics[height=0.22\textheight]{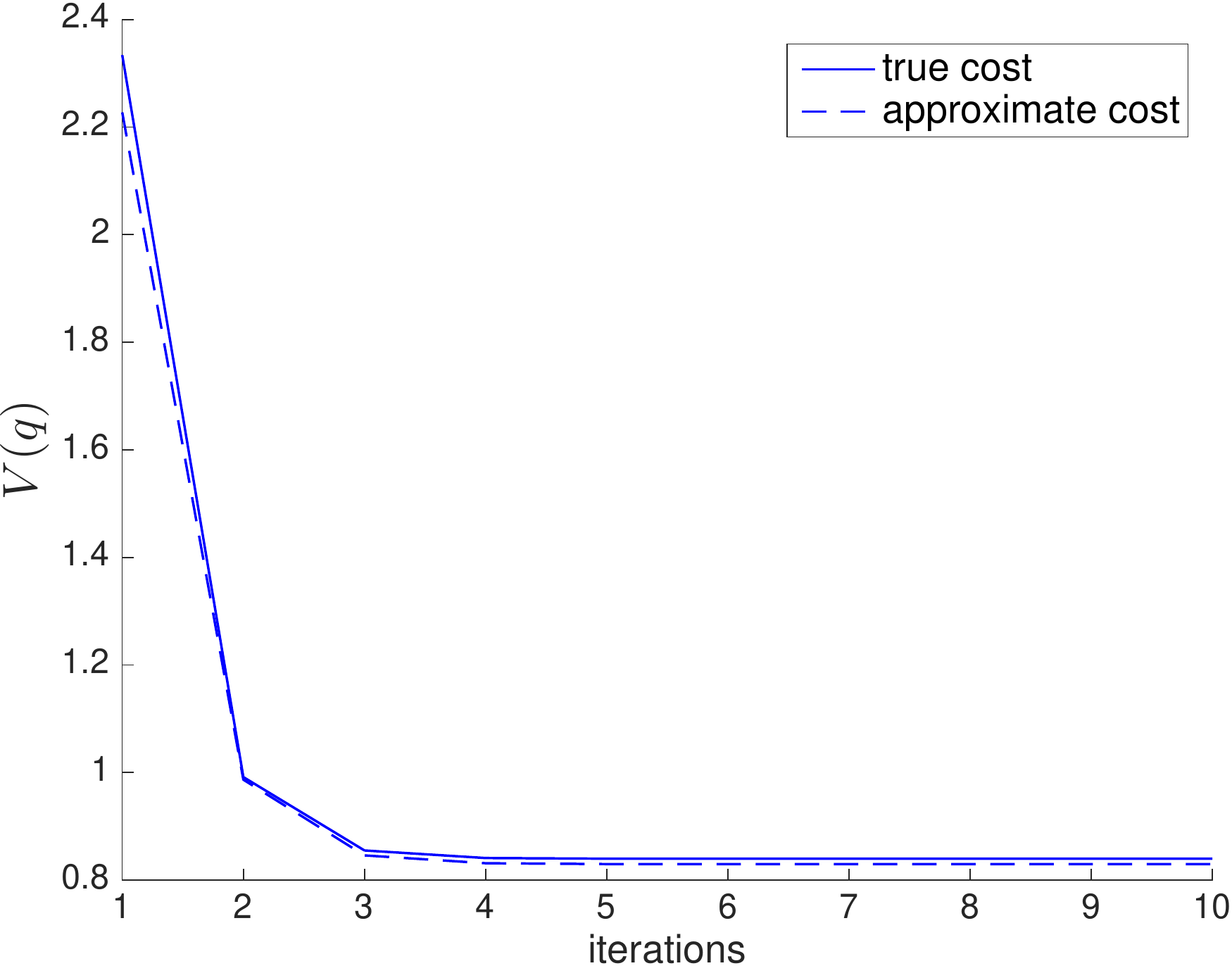} \\

\vspace*{-0.03in}
\begin{turn}{90}\hspace*{0.7in}ESGVI deriv M=3\end{turn} \hspace{0.1in}
\includegraphics[height=0.23\textheight]{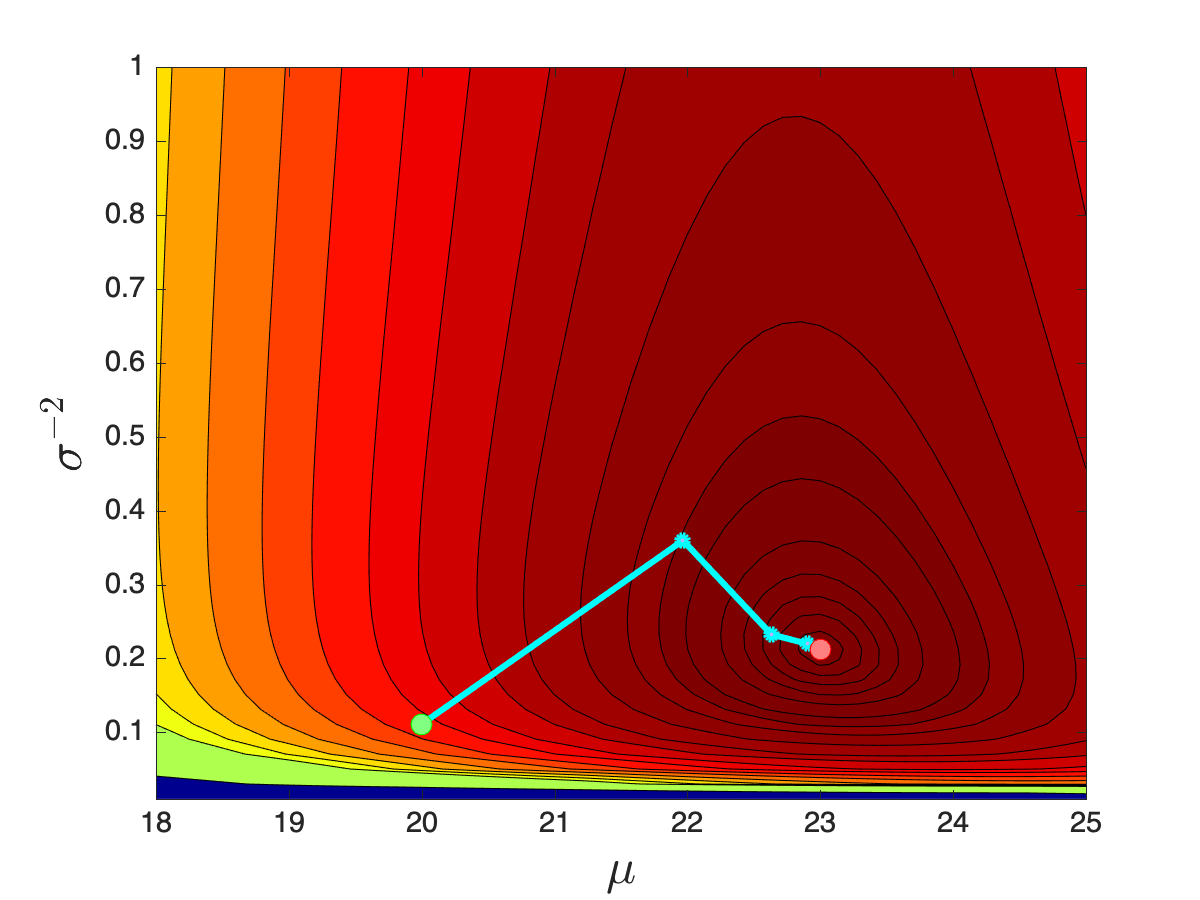} 
\includegraphics[height=0.22\textheight]{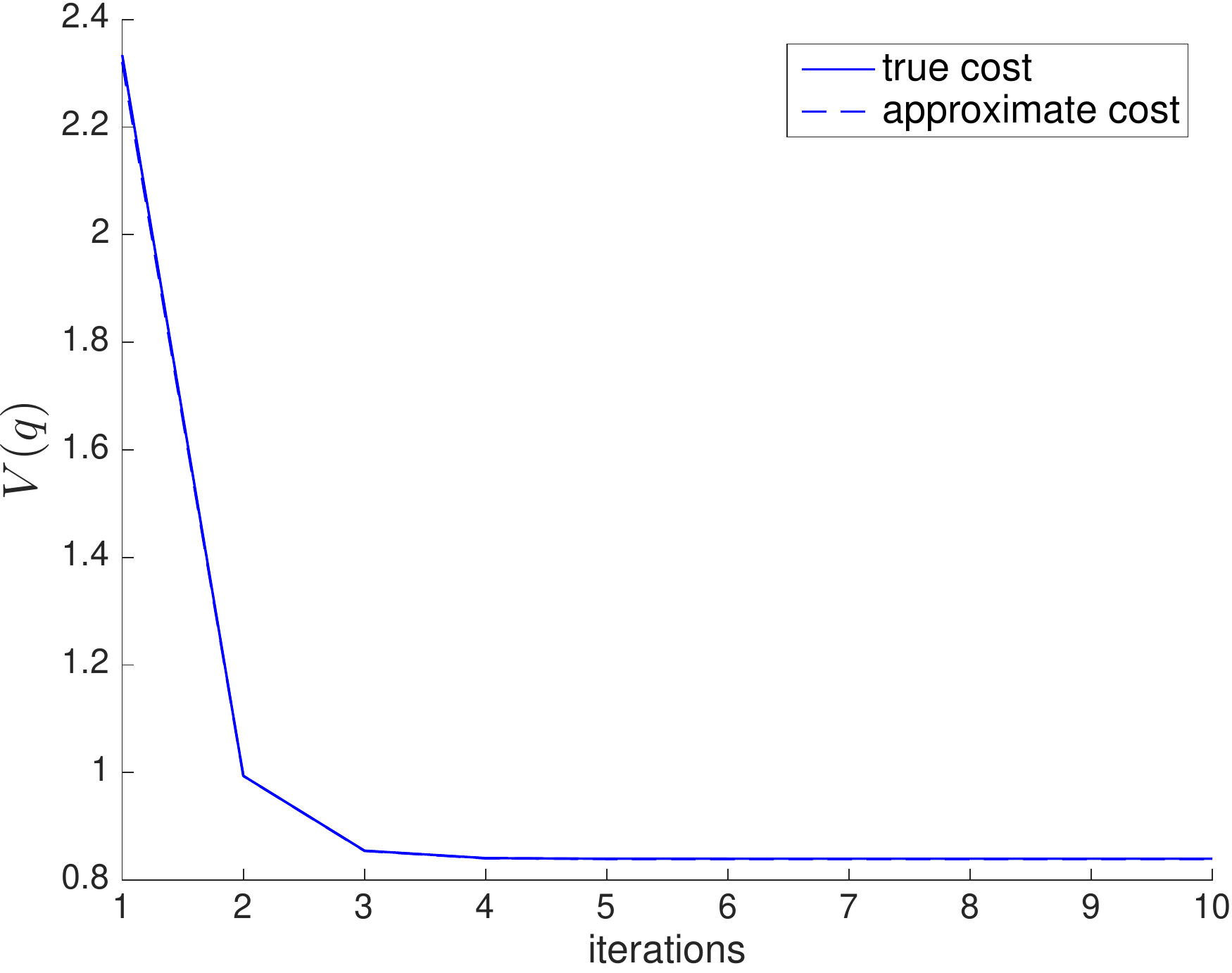} \\

\vspace*{-0.03in}
\begin{turn}{90}\hspace*{0.5in}ESGVI deriv-free M=4\end{turn} \hspace{0.1in}
\includegraphics[height=0.23\textheight]{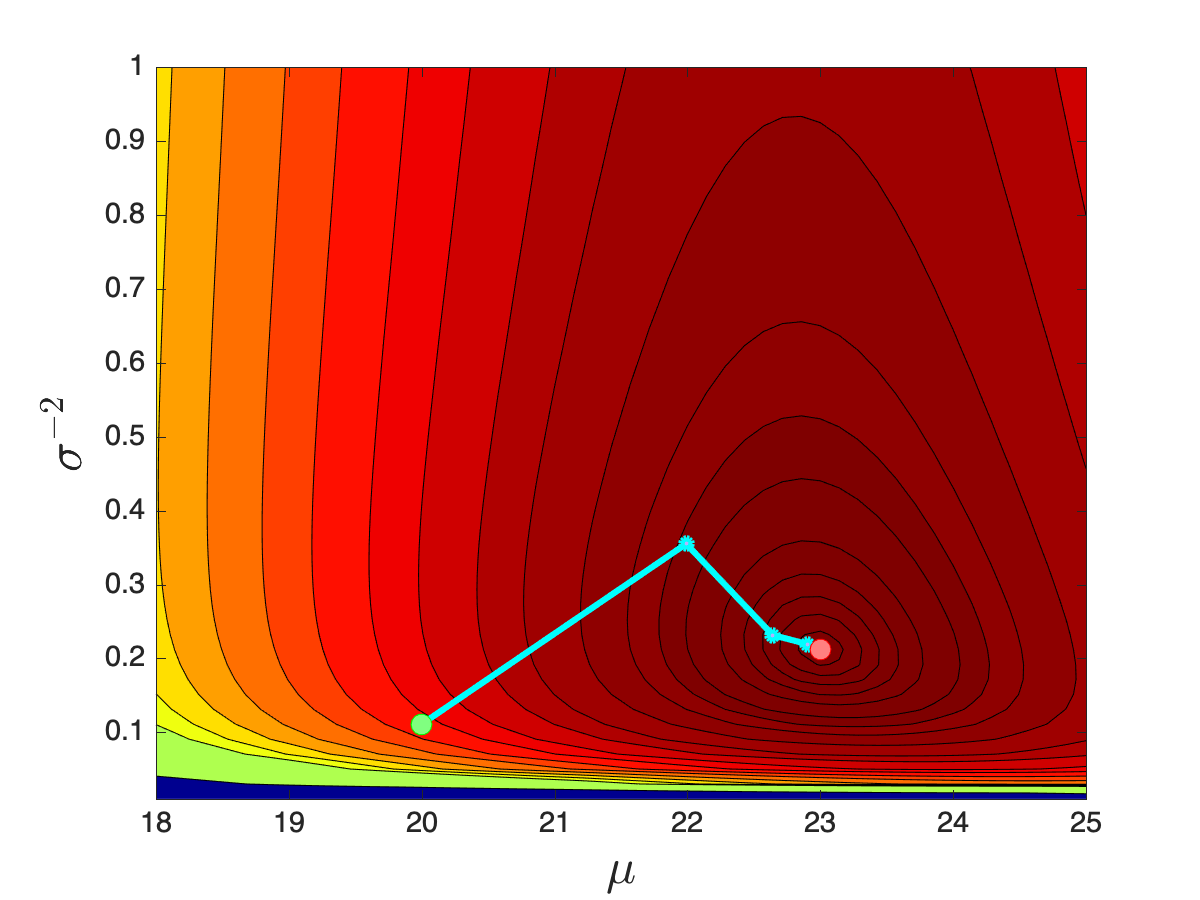} 
\includegraphics[height=0.22\textheight]{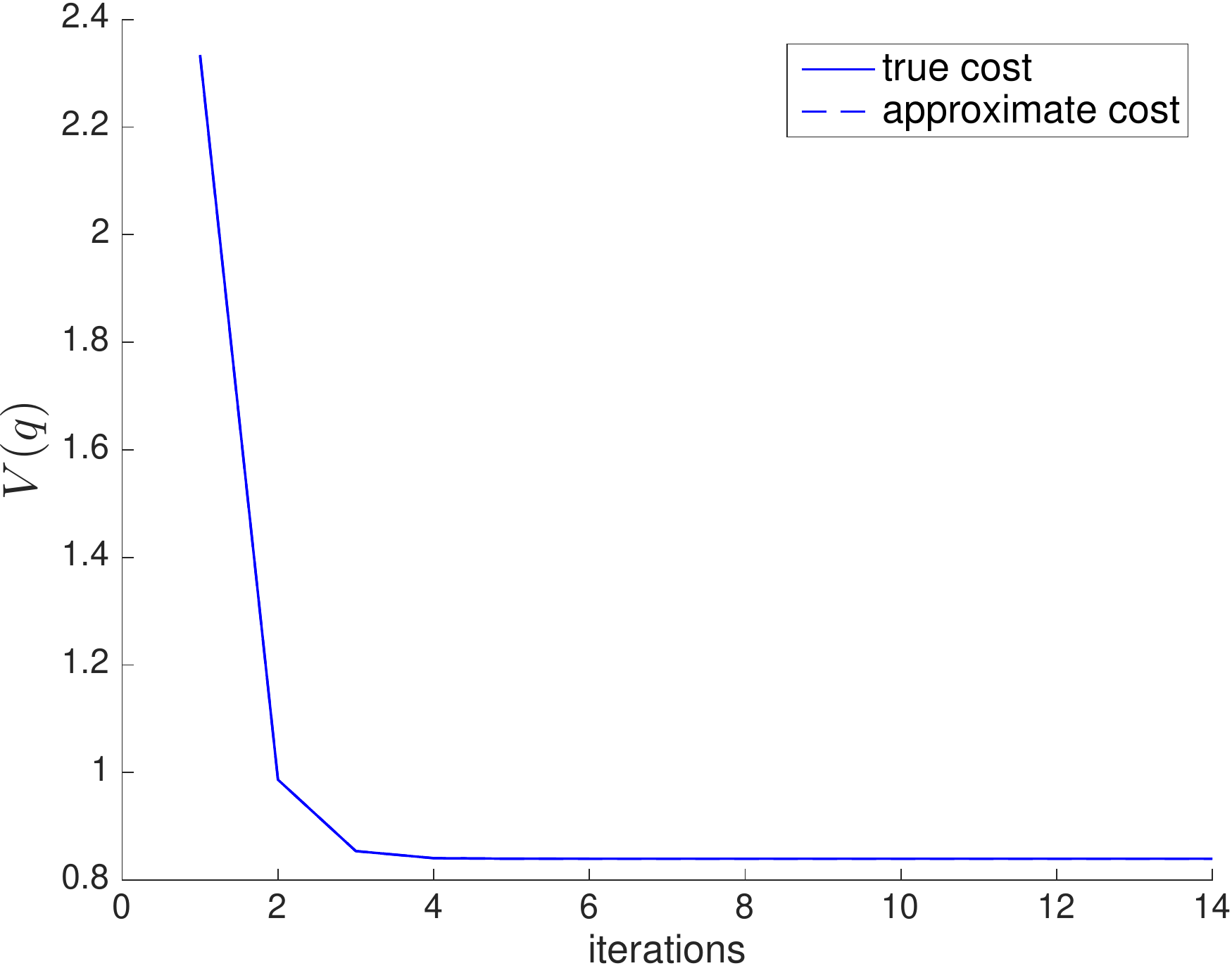}
\vspace*{-0.1in}
\caption{(Experiment 1) One trial of the one-dimensional stereo camera simulation showing the convergence history for four different algorithms shown in each row.  The left column shows a contour map of the loss functional, $V(q)$, with the steps the optimizer took starting from the prior (green dot) to its converged value (red dot).  The right column shows the loss at each iteration; as each algorithm makes different approximations to the loss during execution, we show the loss that each algorithm used to make decisions and the actual loss at each step.}
\label{fig:esgvi_exp1a}
\end{figure*}

To conduct a proper Bayesian experiment, we ran $100,000$ trials where each one consisted of drawing the latent state from the prior, then producing a noisy measurement given that state.  To stay clear of edge cases (e.g., negative distance), we only accepted a draw of the latent state if it was within $4$  standard deviations of the mean, resulting in $6$ out of the $100,000$ experiments to not be accepted and the state redrawn.  We then ran several variants of our algorithm summarized in Table~\ref{tab:alglabels}.  Everything else to do with the experiment was the same for all algorithms, allowing a fair comparison.  Figure~\ref{fig:esgvi_exp1b} shows the statistical results of our $100,000$ trials as boxplots.  The columns correspond to the different versions of our algorithm while the rows are different performance metrics.  The first column (analytical Hessian and Jacobian with a single quadrature point at the mean) is equivalent to a standard \ac{MAP} approach.  We can see that our new algorithms do require a few more iterations (first row) to converge than \ac{MAP}, which is to say that it takes more computation to arrive at a better approximation to the posterior.  We also see that the new algorithms do find a lower final value of the loss functional, $V(q)$, which is what we asked them to minimize (second row).  

We also wanted to see if the new algorithms were less biased and more consistent than \ac{MAP}, and so calculated the bias as the sum of errors (third row), squared error (fourth row), and squared Mahalanobis / \ac{NEES} (fifth row).  \change{We cannot ask the estimator to minimize these quantities (because they are based on knowledge of the groundtruth values of the latent state) } but our hypothesis has been that by minimizing $V(q)$, we should also do better on these metrics.  Looking at the third row, all the \ac{GVI} variants are less biased than \ac{MAP} by \change{an order } of magnitude or more.  Our \ac{MAP} error of $-30.6$ cm is consistent with the result reported by \citet[\S4]{barfoot17}.  The best algorithm reported there, the \ac{ISPKF} \citep{sibley06}, had a bias of $-3.84$ cm. Our best algorithm had a bias of $0.3$ cm.  Squared error (fourth row) is also slightly improved compared to \ac{MAP}.  

The squared Mahalanobis / \ac{NEES} error should be close to $1$ for a one-dimensional problem; here the results are mixed, with some of our approaches doing better than \ac{MAP} and some not.  It seems that our choice of $\mbox{KL}(q||p)$ rather than $\mbox{KL}(p||q)$ results in a slightly overconfident covariance.  \citet[fig 10.1]{bishop06} shows a similar situation for the same choice of $\mbox{KL}(q||p)$ as does \citet{alaluhtala15}.  As discussed earlier, it may be possible to overcome this by changing the relative weighting between the two main terms in $V(q)$ through the use of a \change{metaparameter } that is optimized for a particular situation.

\begin{figure*}[t]
\centering
\includegraphics[width=0.6\textwidth]{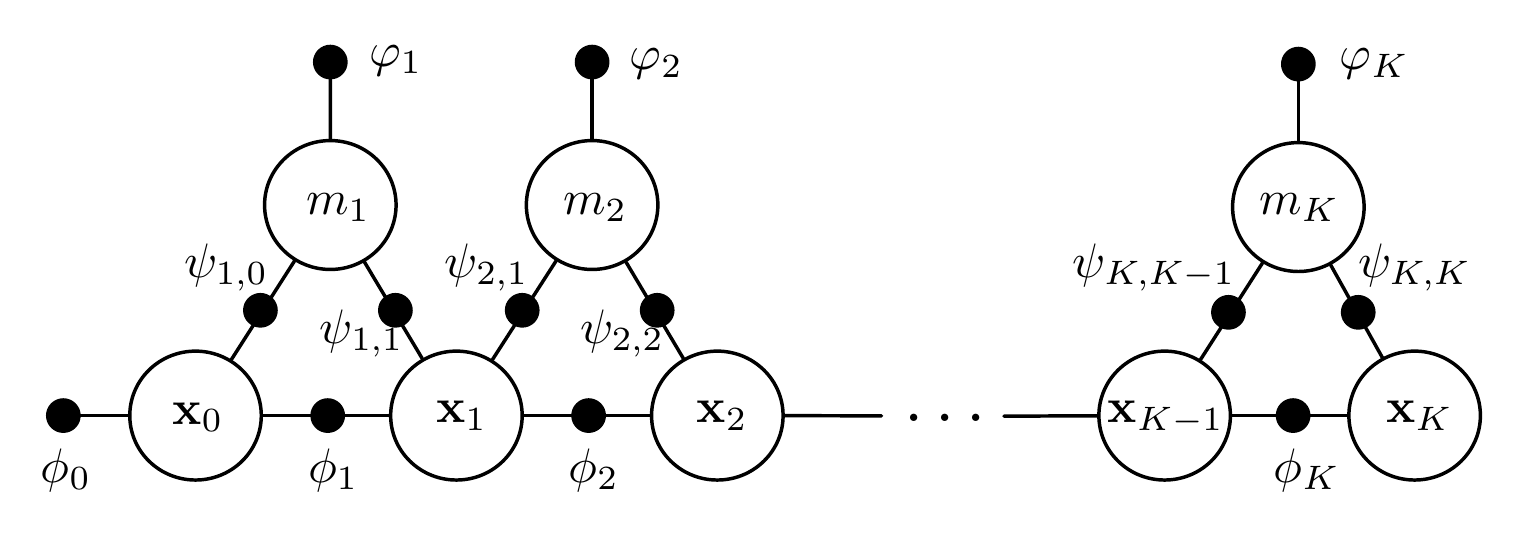}
\vspace*{-0.1in}
\caption{(Experiment 2) Factor graph for the stereo $K$-dimensional simulation.  White circles represent variables to be estimated (both robot positions and landmark positions).  Small black dots represent factors in the joint likelihood of the state and data.  }
\label{fig:exp2_fg}
\end{figure*}

Figure~\ref{fig:esgvi_exp1a} shows the details of a single trial of the $100,000$ that we ran.  We show only a subset of the algorithms (rows) in the interest of space.  The left column provides a contour plot of $V(q)$ ($\mu$ on the horizontal and $\sigma^{-2}$ on the vertical) and the path the optimizer actually took to arrive at its minimum (red dot) starting from the prior (green dot).  The right column shows the value of $V(q)$ at each iteration.  It is worth noting that we show the true value of the loss \change{(calculated using~\eqref{eq:computeV} with a large number of quadrature points) } as well as the approximation of the loss that the algorithm had access to during its iterations (each algorithm used a different number of quadrature points, $M$).  We see that the \ac{MAP} approach clearly does not terminate at the minimum of $V(q)$; its approximation of the required expectations is too severe to converge to the minimum.  The other algorithms end up very close to the true minimum, in a similar number of iterations.  

In the next section, we introduce time and allow our simulated robot to move along the $x$-axis, with the same nonlinear stereo camera model.  Our aim is to show that we can exploit the sparse structure of the problem in higher dimensions.  \change{We deliberately chose a \ac{SLAM} problem to showcase that our \ac{ESGVI} approach can work even when the probabilistic graphical model has loops. }

\subsection{Experiment 2:  Stereo $K$-Dimensional Simulation}

\label{sec:stereoKexp}

This simulation was designed to show that we can scale up to a more realistic problem size, while still deriving benefit from our variational approach.  We extend the stereo camera problem from the previous section to the time domain by allowing a robot to move along the $x$-axis.  In order to continue to carry out a proper Bayesian comparison of algorithms, we introduce a prior both on the robot motion and on the landmark positions in this \ac{SLAM} problem.  The {\em factor graph} for the problem can be seen in Figure~\ref{fig:exp2_fg}.

The state to be estimated is
\begin{equation}
\mbf{x} = \bbm \mbf{x}_0 \\ \mbf{x}_1 \\  \vdots  \\\mbf{x}_K \\ m_1 \\ \vdots \\ m_K \ebm, \quad \mbf{x}_k = \bbm p_k \\ v_k \ebm,
\end{equation}
where $p_k$ is a robot position, $v_k$ a robot speed, and $m_k$ a landmark position.  The problem is highly structured as each landmark is seen exactly twice, from two consecutive robot positions.

\begin{figure*}[t]
\centering
\includegraphics[width=\textwidth]{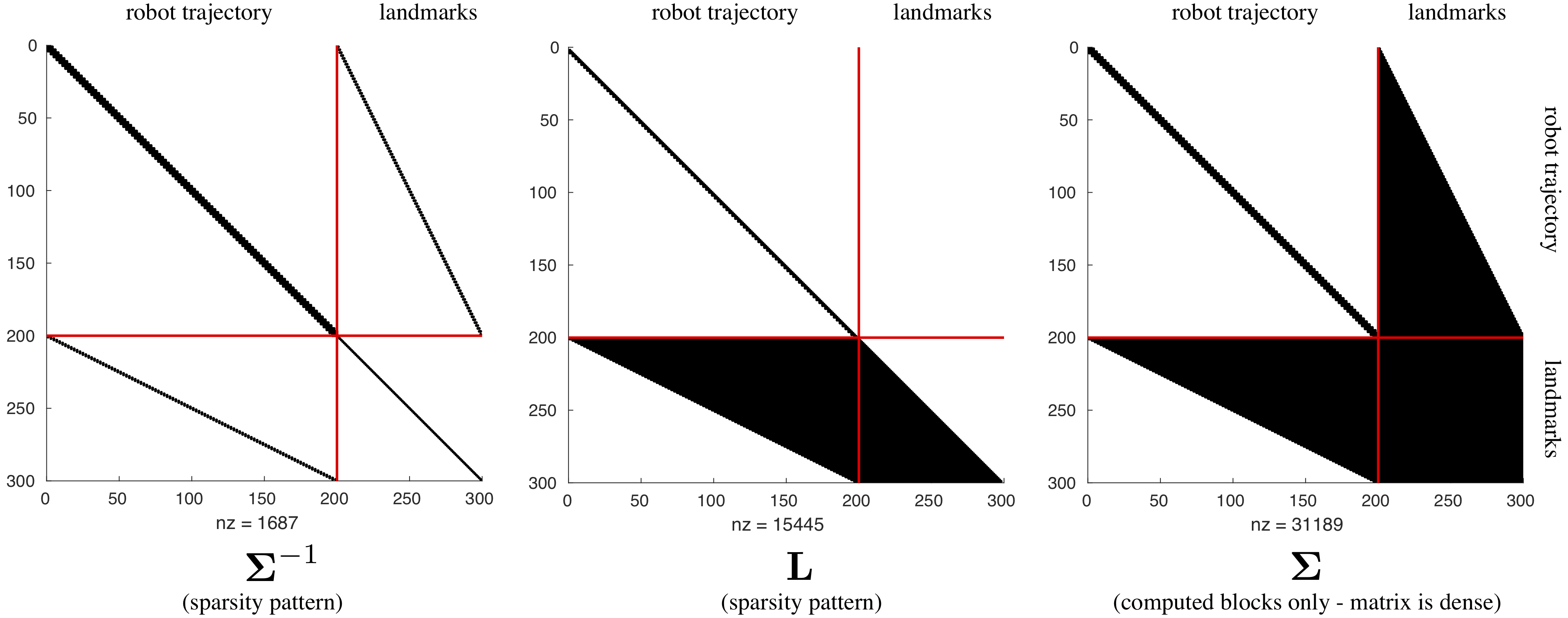}
\vspace*{-0.1in}
\caption{(Experiment 2) Sparsity patterns for the stereo $K$-dimensional simulation.  The red partition lines separate the robot state variables from the landmark variables.  The inverse covariance, $\mbs{\Sigma}^{-1}$, is highly sparse owing to the factor graph pattern in Figure~\ref{fig:exp2_fg}; only 1,687/89,401 = 1.9\% of entries are nonzero. After performing a lower-diagional-upper decomposition, the lower factor, $\mbf{L}$, becomes more filled in owing to the `four corners of a box' rule; 15,445/89,401 = 17.3\% of entries are non-zero.  Finally, we see that only a fraction of the entries of $\mbs{\Sigma}$ are required despite the fact that this matrix is actually dense; since $\mbs{\Sigma}$ is symmetric, we only need to calculate 17.3\% of it as well.}
\label{fig:exp2_sparsity}
\end{figure*}

\begin{figure*}[p]
\centering
\vspace*{1.0in}
\includegraphics[width=\textwidth]{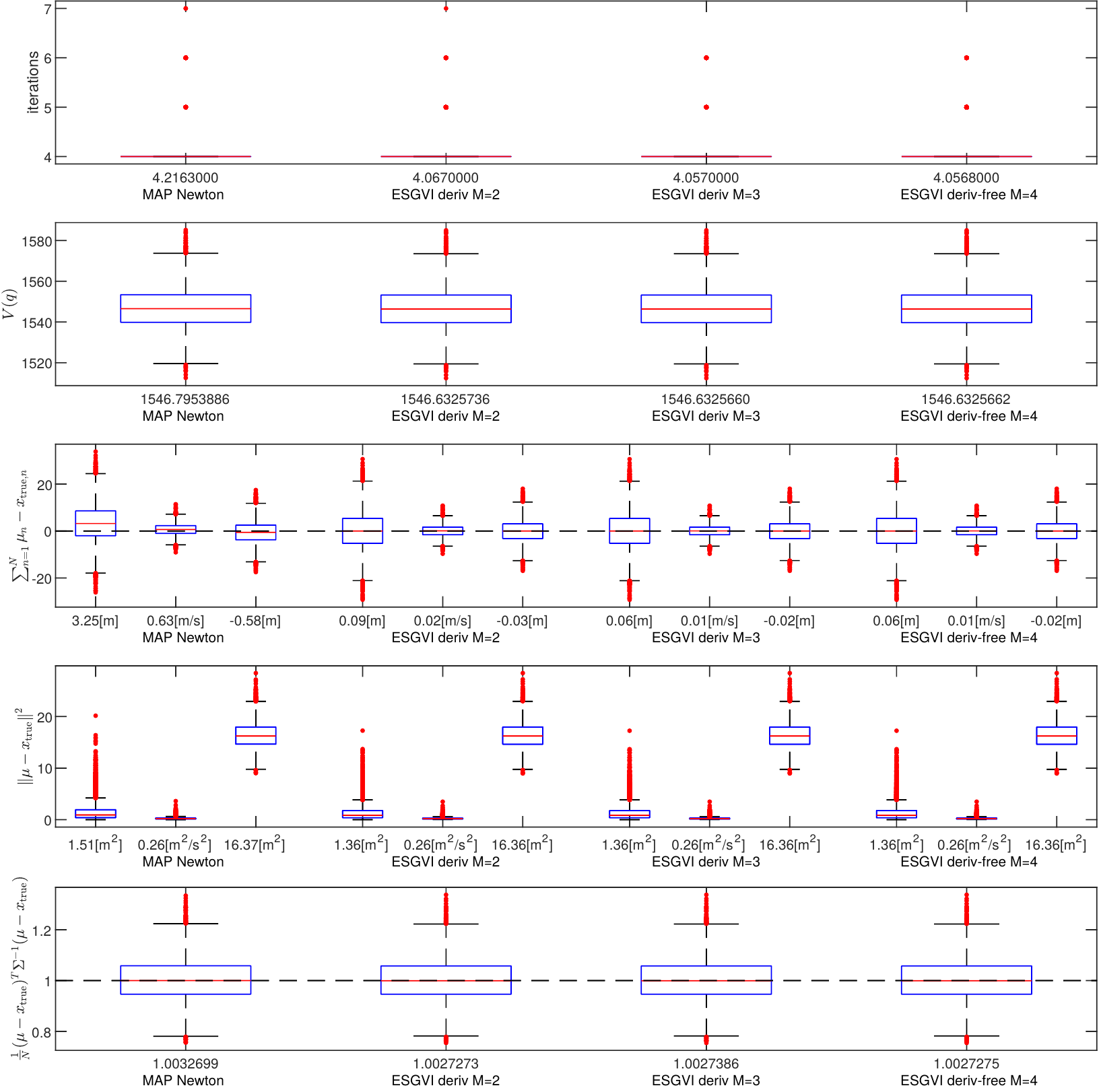}
\caption{(Experiment 2) Statistical results of $10,000$ trials of the $K$-dimensional stereo camera simulation shown as standard boxplots.  The different rows show different performance metrics for the different variants of our algorithm (columns).  Table~\ref{tab:alglabels} provides details of the different algorithms tested.  \change{For the bias (row 3) and squared error (row 4) metrics, we show the results separately for (left to right) robot position, robot velocity, and landmark position to avoid combining quantities with different units. } The number above an algorithm label is its mean performance on that metric.   Further details are discussed in the text.}
\label{fig:exp2_results}
\vspace*{1.0in}
\end{figure*}

For the (linear) prior factors we have
\begin{subequations}
\begin{eqnarray}
\phi_k & = & \left\{ 
\begin{array}{ll}
\frac{1}{2} (\mbf{x}_0 - \pri{\mbf{x}}_0)^T \pri{\mbf{P}}^{-1} (\mbf{x}_0 - \pri{\mbf{x}}_0)  & k=0 \\
\frac{1}{2} (\mbf{x}_k - \mbf{A} \mbf{x}_{k-1})^T \mbf{Q}^{-1}  \\ \;\qquad\qquad \times \; (\mbf{x}_k - \mbf{A} \mbf{x}_{k-1})                  & k>0
\end{array}   \right. ,\qquad \\
\varphi_k & = & \frac{1}{2} \frac{(m_k - \mu_{m,k})^2}{\sigma_m^2} ,
\end{eqnarray}
\end{subequations}
with
\begin{gather}
\pri{\mbf{P}} = \mbox{diag}(\sigma_p^2, \sigma_v^2), \quad \mbf{A} = \bbm 1 & T \\ 0 & 1 \ebm, \nonumber \\ \mbf{Q} = \bbm \frac{1}{3} T^3 Q_C & \frac{1}{2} T^2 Q_C \\ \frac{1}{2} T^2 Q_C  & T Q_C  \ebm,
\end{gather}
where $T$ is the discrete-time sampling period, $Q_C$ is a power spectral density, and $\sigma_p^2$, $\sigma_v^2$, $\sigma_m^2$ are variances.  The robot state prior encourages constant velocity \cite[\S 3, p.85]{barfoot17}.  The landmark prior is simply a Gaussian centered at the true landmark location, $\mu_{m,k}$.

For the (nonlinear) measurement factors we have
\begin{equation}
\psi_{\ell,k} = \frac{1}{2} \frac{ \left( y_{\ell,k} - \frac{fb}{m_\ell - p_k} \right)^2}{\sigma_r^2},
\end{equation}
where $f$ and $b$ are the camera parameters (same as the previous experiment), $y_{\ell,k}$ is the disparity measurement of the $\ell$th landmark from the $k$th position, and $\sigma_r^2$ is the measurement noise variance.

The negative log-likelihood of the state and data is then
\begin{multline}
-\ln p(\mbf{x}, \mbf{z})  = \sum_{k=0}^K \phi_k + \sum_{k=1}^K \varphi_k  \\ + \sum_{k=1}^K (\psi_{k,k-1} + \psi_{k,k})  + \mbox{constant}.
\end{multline}
We set the maximum number of timesteps to be $K=99$ for this problem, resulting in an overall state dimension of $299$.
Figure~\ref{fig:exp2_sparsity} shows the sparsity patterns of $\mbs{\Sigma}^{-1}$, $\mbf{L}$, and the blocks of $\mbs{\Sigma}$ that get computed by the method of \citet{takahashi73}.  This can very likely be improved further using modern sparsity techniques but the point is that we have a proof-of-concept scheme that can compute the subset of blocks of $\mbs{\Sigma}$ required to carry out \ac{GVI}.

We ran $10,000$ trials of this simulation.  In each trial, we drew the latent robot trajectory and landmark states from the Bayesian prior, then simulated the nonlinear measurements with a random draw of the noise.  We estimated the full state using four different algorithms from Table~\ref{tab:alglabels}: `MAP Newton', `ESGVI deriv M=2', `ESGVI deriv M=3', and `ESGVI deriv-free M=4'.  

Figure~\ref{fig:exp2_results} shows the statistical results of the $10,000$ trials.  
All the algorithms converge well in a small number of iterations (usually $4$).  Increasing the number of cubature points for the derivative-based methods does result in reducing the overall value of the loss functional, $V(q)$; the `ESGVI deriv-free M=4' method does about as well as the `ESGVI deriv M=3' method but requires no analytical derivatives of the factors.  

\begin{table}[t]
\caption{(Experiment 2) Wall-clock time per iteration for tested algorithms.}
\change{
\begin{center}
\begin{tabular}{l|r}
algorithm  & wall-clock time \\
label & per iteration [s]  \\ \hline\hline
\ac{MAP} Newton & 0.0262  \\
\ac{ESGVI} deriv M=2 & 0.0745   \\
\ac{ESGVI} deriv M=3 & 0.1836   \\
\ac{ESGVI} deriv-free M=4 & 0.2653 
\end{tabular}
\end{center}
}
\label{tab:iterationtime2}
\vspace*{-0.3in}
\end{table}

As in the one-dimensional simulation, the bias in the estimate is significantly reduced in the \ac{ESGVI} approaches compared to the \ac{MAP} approach (row 3 of Figure~\ref{fig:exp2_results}). 
This is important since this result can be achieved in a tractable way for large-scale problems and even without analytical derivatives.  \change{Note that we chose to show this metric separately for (left to right) robot position, robot velocity, and landmark position to avoid combining quantities with different units. We can see that the improvements offered by \ac{ESGVI} are mostly due to the robot position variables but robot velocity and landmark positions are also improved. }

The \ac{ESGVI} methods also do slightly better than \ac{MAP} on squared error (row 4 of Figure~\ref{fig:exp2_results}). \change{Again, we chose to show this metric separately for (left to right) robot position, robot velocity, and landmark position to avoid combining quantities with different units.  We can see that the improvements offered by \ac{ESGVI} are mostly on the robot position variables.  For squared Mahalanobis distance / \ac{NEES} (row 5 of Figure~\ref{fig:exp2_results}), all algorithms perform well. }

\change{Table~\ref{tab:iterationtime2} shows the wall-clock time per iteration for the different algorithms.  Naturally, as we use more cubature points, the computational cost increases.  The derivative-free version of \ac{ESGVI} is about an order of magnitude slower than the \ac{MAP} approach.  We believe that there may be applications where this increased computational cost is worthwhile in terms of increased performance or the convenience of avoiding the calculation of derivatives.}

\begin{figure*}[t]
\centering
\includegraphics[width=\textwidth]{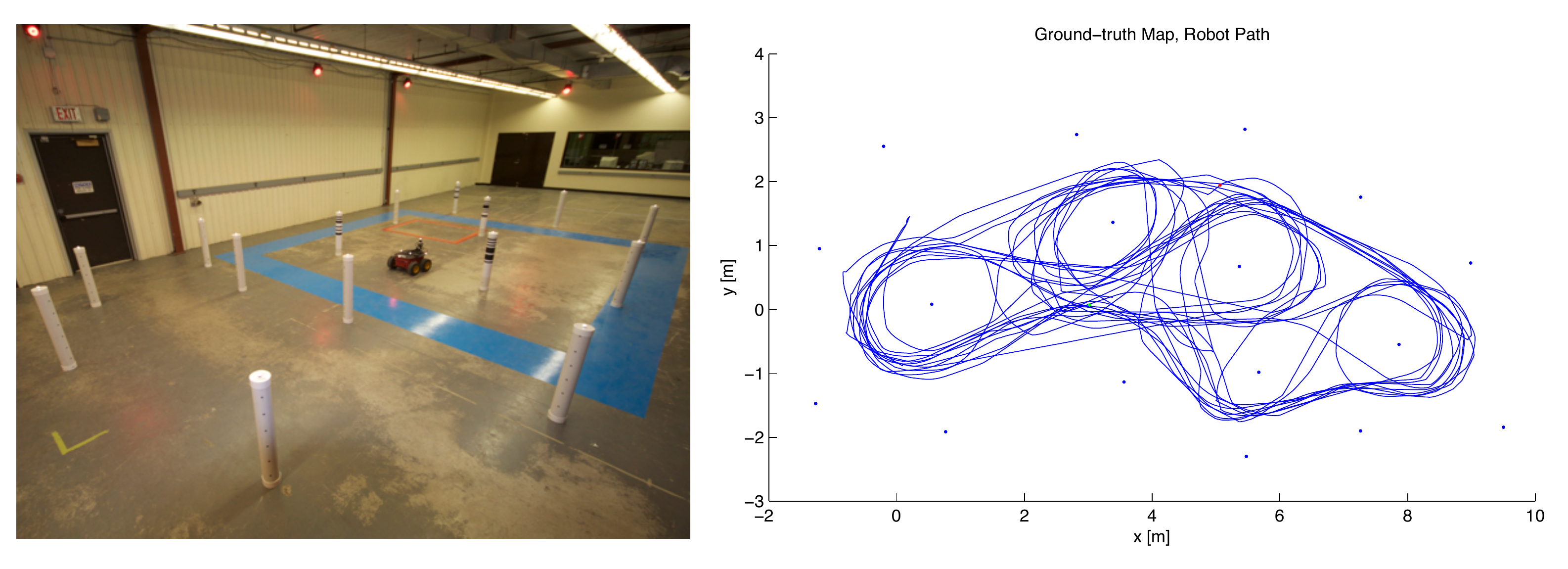}
\caption{(Experiment 3) Setup: (left) a mobile robot navigates amongst a map of landmarks; it receives bearing measurements to some landmarks as well as wheel odometry.  (right) the ground-truth path of the robot and landmark map as measured by an overhead camera system.}
\label{fig:dataset2}
\vspace*{-0.2in}
\end{figure*}

\begin{figure*}[t]
\centering
\includegraphics[width=0.7\textwidth]{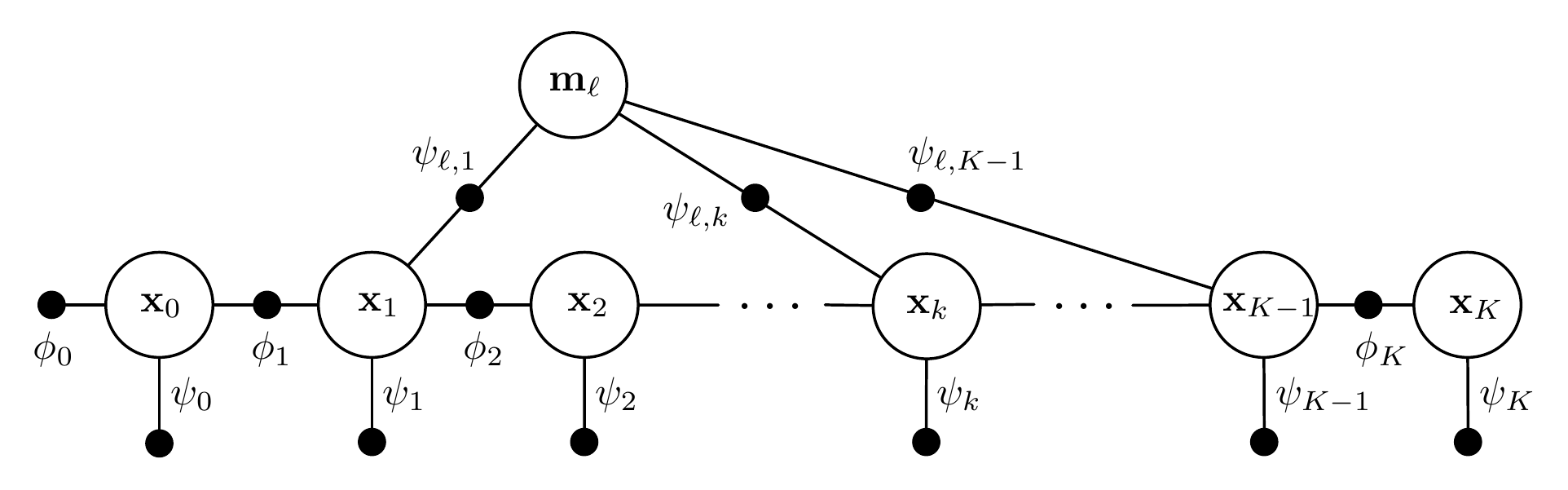}
\caption{(Experiment 3) Factor graph for our robot dataset.  White circles indicate variables and small black circles indicate factors involving variables.}
\label{fig:exp3_fg}
\end{figure*}

\subsection{Experiment 3: Robot Dataset}

\label{sec:robexp}

Finally, we consider a batch \ac{SLAM} problem with a robot driving around and building a map of landmarks as depicted in Figure~\ref{fig:dataset2}.  \change{The robot is equipped with a laser rangefinder and wheel odometers and must estimate its own trajectory and the locations of a number of tubular landmarks. }This dataset has been used previously by \citet{barfoot14} to test \ac{SLAM} algorithms.  Groundtruth for both the robot trajectory and landmark positions \change{(this is a unique aspect of this dataset) } is provided by a Vicon motion capture system.  The whole dataset is $12,000$ timesteps long, which we broke into six subsequences of $2000$ timestamps; statistical performance reported below is an average over these six subsequences.  We assume that the data association (i.e., which measurement corresponds to which landmark) is known in this experiment to restrict testing to the state estimation part of the problem.

The state to be estimated is
\begin{equation}
\mbf{x} = \bbm \mbf{x}_0 \\ \mbf{x}_1 \\  \vdots  \\\mbf{x}_K \\ \mbf{m}_1 \\ \vdots \\ \mbf{m}_L \ebm, \quad \mbf{x}_k = \bbm x_k \\ y_k \\ \theta_k \\ \dot{x}_k \\ \dot{y}_k \\ \dot{\theta}_k \ebm, \quad \mbf{m}_\ell = \bbm x_\ell \\ y_\ell \ebm ,
\end{equation}
where $\mbf{x}_k$ is a robot state and $\mbf{m}_\ell$ a landmark position.  For each of our six subsequences we have $K=2000$ and $L=17$. 

\begin{figure*}[t]
\centering
\includegraphics[width=\textwidth]{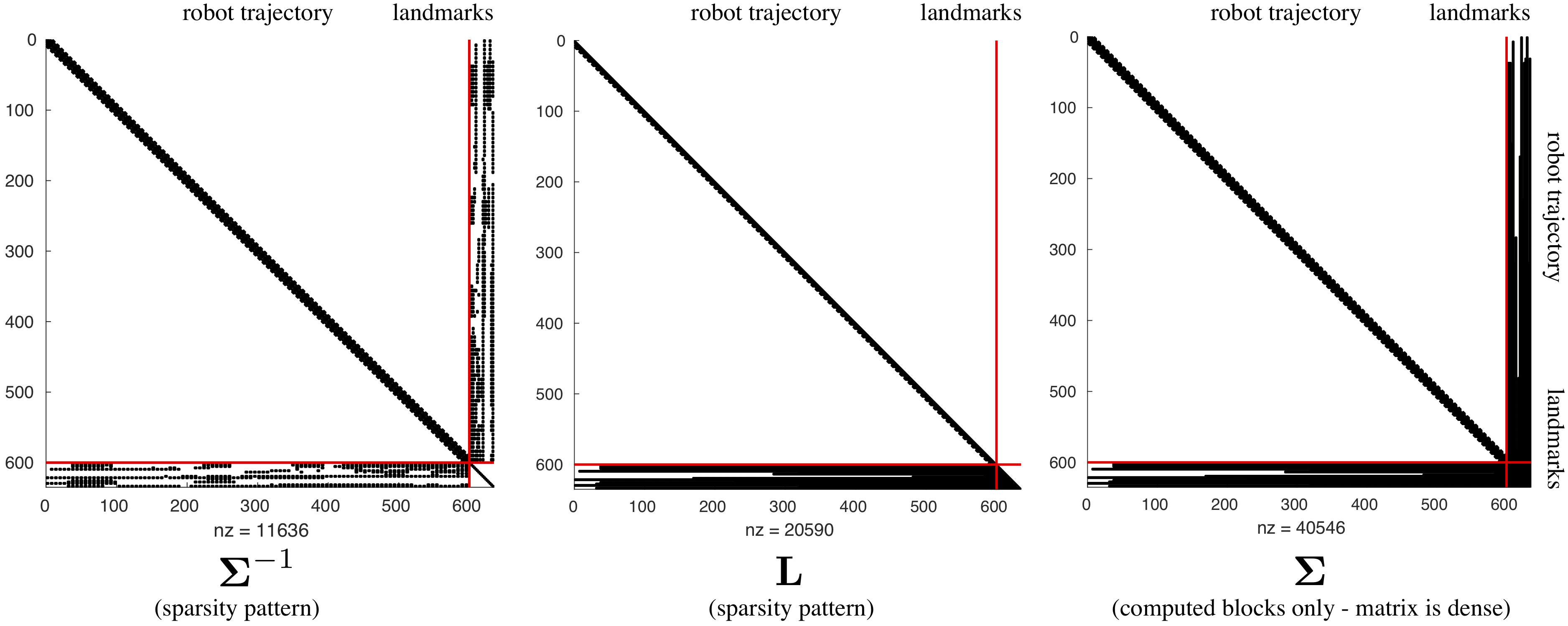}
\caption{(Experiment 3) Sparsity patterns for the first $100$ of $2000$ timestamps of the robot dataset.  The red partition lines separate the robot state variables from the landmark variables.  The inverse covariance, $\mbs{\Sigma}^{-1}$, is highly sparse owing to the factor graph pattern in Figure~\ref{fig:exp3_fg}; only 11,636/401,956 = 2.9\% of entries are nonzero. After performing a lower-diagional-upper decomposition, the lower factor, $\mbf{L}$, becomes more filled in owing to the `four corners of a box' rule; 20,590/401,956 = 5.1\% of entries are non-zero.  Finally, we see that only a fraction of the entries of $\mbs{\Sigma}$ are required despite the fact that this matrix is actually dense; since $\mbs{\Sigma}$ is symmetric, we only need to calculate 5.1\% of it as well.  For the full $2000$-timestamp dataset the sparsity is even more favourable for \ac{ESGVI}, but the landmark part of the pattern becomes difficult to visualize due to its small size relative to the trajectory part.}
\label{fig:exp3_sparsity}
\vspace*{-0.1in}
\end{figure*}

Figure~\ref{fig:exp3_fg} shows the factor graph for this experiment and Figure~\ref{fig:exp3_sparsity} shows the corresponding sparsity patterns.  For the (linear) prior factor on the robot states we have
\begin{equation}
\phi_k = \left\{ 
\begin{array}{ll}
\frac{1}{2} (\mbf{x}_0 - \pri{\mbf{x}}_0)^T \pri{\mbf{P}}^{-1} (\mbf{x}_0 - \pri{\mbf{x}}_0)  & k=0 \\
\frac{1}{2} (\mbf{x}_k - \mbf{A} \mbf{x}_{k-1})^T \mbf{Q}^{-1}  (\mbf{x}_k - \mbf{A} \mbf{x}_{k-1})                  & k>0
\end{array}   \right. ,
\end{equation}
with
\begin{gather}
\pri{\mbf{P}} = \mbox{diag}(\sigma_x^2, \sigma_y^2,\sigma_\theta^2, \sigma_{\dot{x}}^2, \sigma_{\dot{y}}^2, \sigma_{\dot{\theta}}^2), \;\; \mbf{A} = \bbm \mbf{1} & T \mbf{1} \\ \mbf{0} & \mbf{1} \ebm, \nonumber \\ \mbf{Q} = \bbm \frac{1}{3} T^3 \mbf{Q}_C  & \frac{1}{2} T^2 \mbf{Q}_C \\ \frac{1}{2} T^2 \mbf{Q}_C & T \mbf{Q}_C \ebm, \nonumber \\ \mbf{Q}_C = \mbox{diag}(Q_{C,1}, Q_{C,2},Q_{C,3}),
\end{gather}
where $T$ is the discrete-time sampling period, $Q_{C,i}$ are power spectral densities, and $\sigma_x^2$, $\sigma_y^2$,$\sigma_\theta^2$, $\sigma_{\dot{x}}^2$, $\sigma_{\dot{y}}^2$, $\sigma_{\dot{\theta}}^2$ are variances.  The robot state prior encourages constant velocity \cite[\S 3, p.85]{barfoot17}.   Unlike the previous experiment, we do not have a prior on the landmark positions; this was necessary when conducting a proper Bayesian evaluation in the previous experiment, but here we simply have a standard \ac{SLAM} problem.

The (nonlinear) odometry factors, derived from wheel encoder measurements, are 
\begin{equation}
\psi_k = \frac{1}{2} \left( \mbf{v}_k - \mbf{C}_k \mbf{x}_k \right)^T \mbf{S}^{-1}  \left( \mbf{v}_k - \mbf{C}_k \mbf{x}_k \right),
\end{equation}
where
\begin{gather}
\mbf{v}_k = \bbm u_k \\ v_k \\ \omega_k \ebm, \quad \mbf{C}_k = \bbm 0 & 0 & 0 & \cos\theta_k & \sin\theta_k & 0 \\ 0 & 0 & 0 & -\sin\theta_k & \cos\theta_k & 0 \\ 0 & 0 & 0 & 0 & 0 & 1 \ebm, \nonumber \\ \mbf{S} = \mbox{diag}\left(\sigma^2_u, \sigma_v^2, \sigma_\omega^2 \right).
\end{gather}
The $\mbf{v}_k$ consists of measured forward, lateral, and rotational speeds in the robot frame, derived from wheel encoders; we set $v_k = 0$, which enforces the nonholonomy of the wheels as a soft constraint.  The $\sigma_u^2$, $\sigma_v^2$, and $\sigma_\omega^2$ are measurement noise variances. 

The (nonlinear) bearing measurement factors, derived from a laser rangefinder, are
\begin{equation}
\psi_{\ell,k} = \frac{1}{2} \frac{\left( \beta_{\ell,k} - g(\mbf{m}_{\ell},\mbf{x}_k) \right)^2}{\sigma_r^2},
\end{equation}
with
\begin{multline}
g(\mbf{m}_{\ell},\mbf{x}_k) = \mbox{atan2}(y_\ell - y_k - d \sin\theta_k,  \\ x_\ell - x_k - d \cos \theta_k) - \theta_k,
\end{multline}
where $\beta_{\ell,k}$ is a bearing measurement from the $k$th robot pose to the $\ell$th landmark, $d$ is the offset of the laser rangefinder from the robot center in the longitudinal direction, and $\sigma_r^2$ is measurement noise variance.  Although the dataset provides range to the landmarks as well, we chose to neglect these measurements to accentuate the differences between the various algorithms.  Our setup is similar to a monocular camera situation, which is known to be a challenging \ac{SLAM} problem.

Putting these together, our joint state/data likelihood in this case is of the form
\begin{multline}
- \ln p(\mbf{x}, \mbf{z}) =  \sum_{k=0}^K \phi_k + \sum_{k=0}^K \psi_k \\ + \sum_{k=1}^K \sum_{\ell=1}^L \psi_{\ell,k} + \mbox{constant},
\end{multline}
where it is understood that not all $L = 17$ landmarks are actually seen at each timestep and thus we must remove the factors for unseen landmarks.  

\begin{figure*}[p]
\centering
\vspace*{1.0in}
\includegraphics[width=\textwidth]{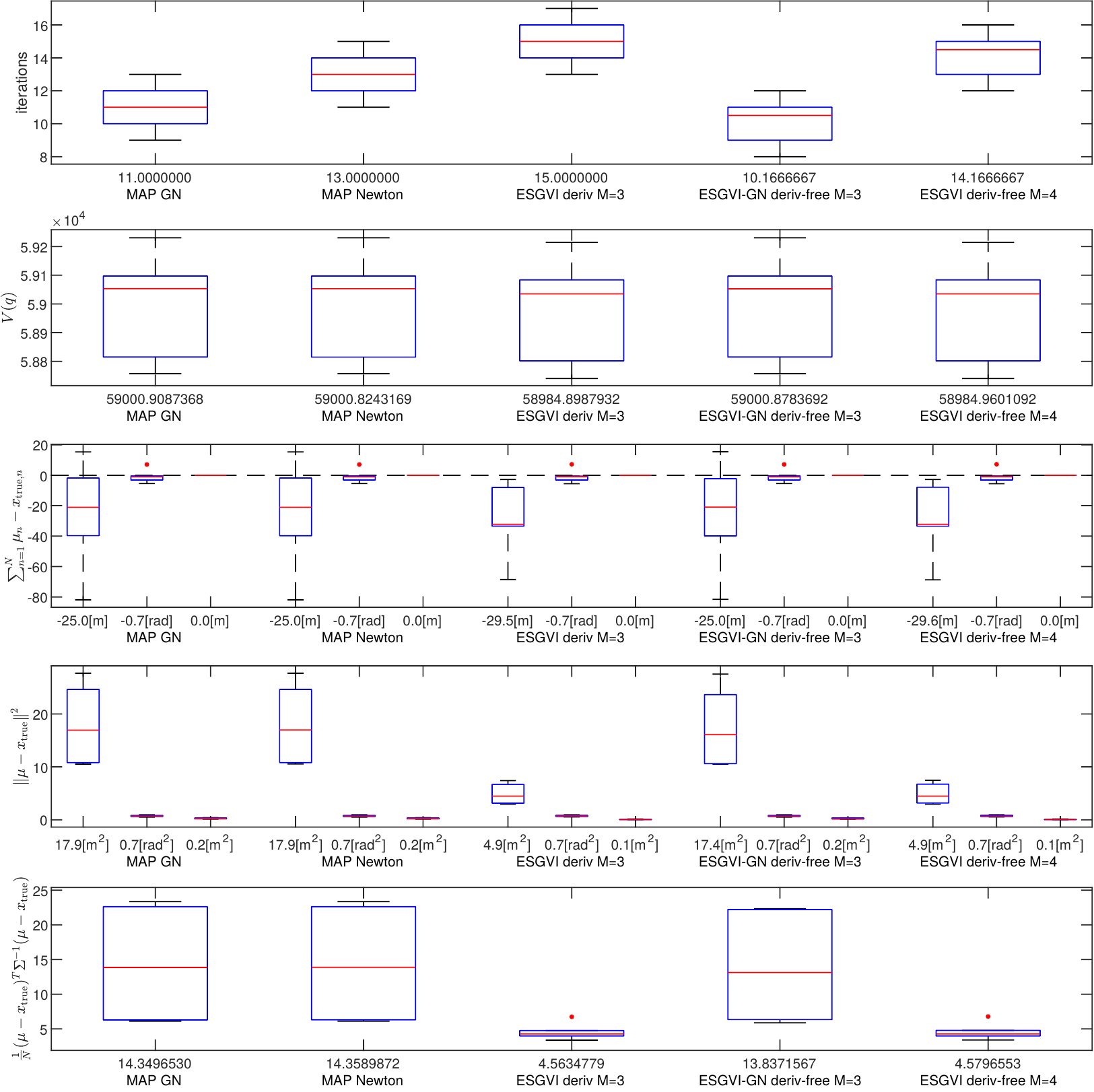}
\caption{(Experiment 3) Statistical results of robot dataset shown as standard boxplots.  The different rows show different performance metrics for the different variants of our algorithm (columns).  Table~\ref{tab:alglabels} provides details of the different algorithms tested.  \change{For the bias (row 3) and squared error (row 4) metrics, we show the results separately for (left to right) robot position, robot orientation, and landmark position to avoid combining quantities with different units. } The number above an algorithm label is its mean performance on that metric, averaged over all $2000$ timestamps and six subsequences.  Further details are discussed in the text. }
\label{fig:exp3_results}
\vspace*{1.0in}
\end{figure*}

By using only bearing measurements, this proved to be a challenging dataset.  We initialized our landmark locations using the bearing-only \ac{RANSAC} \citep{fischler81} strategy described by \citet{mcgarey17}.  We attempted to initialize our robot states using only the wheel odometry information, but this proved too difficult for methods making use of the full Hessian (i.e., Newton's method).  To remedy this problem, we used wheel odometry to initialize Gauss-Newton and then used this to initialize Newton's method.  Specifically, we used \ac{MAP} Gauss-Newton to initialize \ac{MAP} Newton and \ac{ESGVI-GN} to initialize \ac{ESGVI}.  To compare our results to groundtruth, we aligned the resulting landmark map to the groundtruth map since it is well established that \ac{SLAM} produces a relative solution; reported errors are calculated after this alignment.  

\begin{figure*}[p]
\centering
\includegraphics[width=0.73\textwidth]{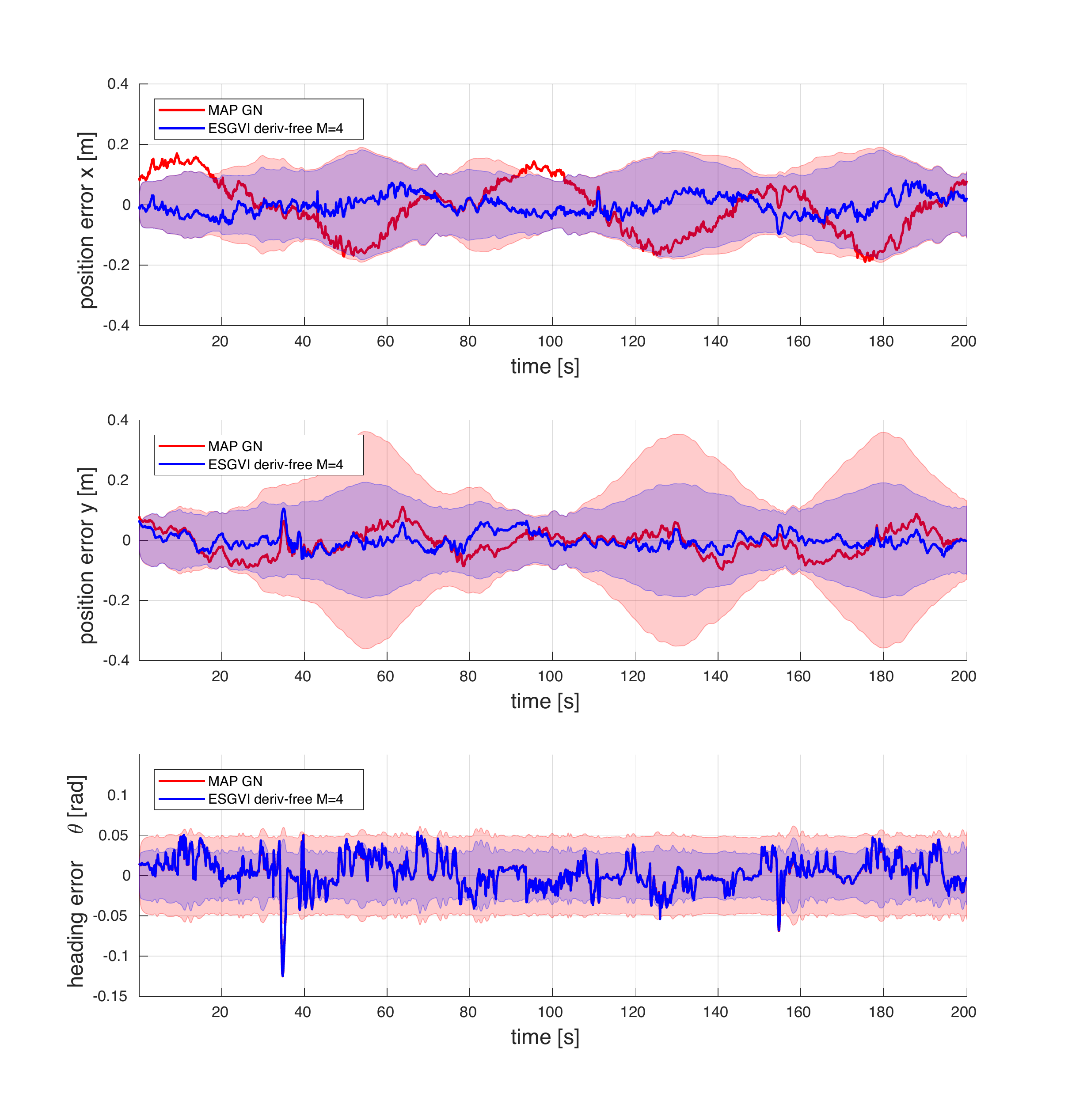} \\
\includegraphics[width=0.69\textwidth]{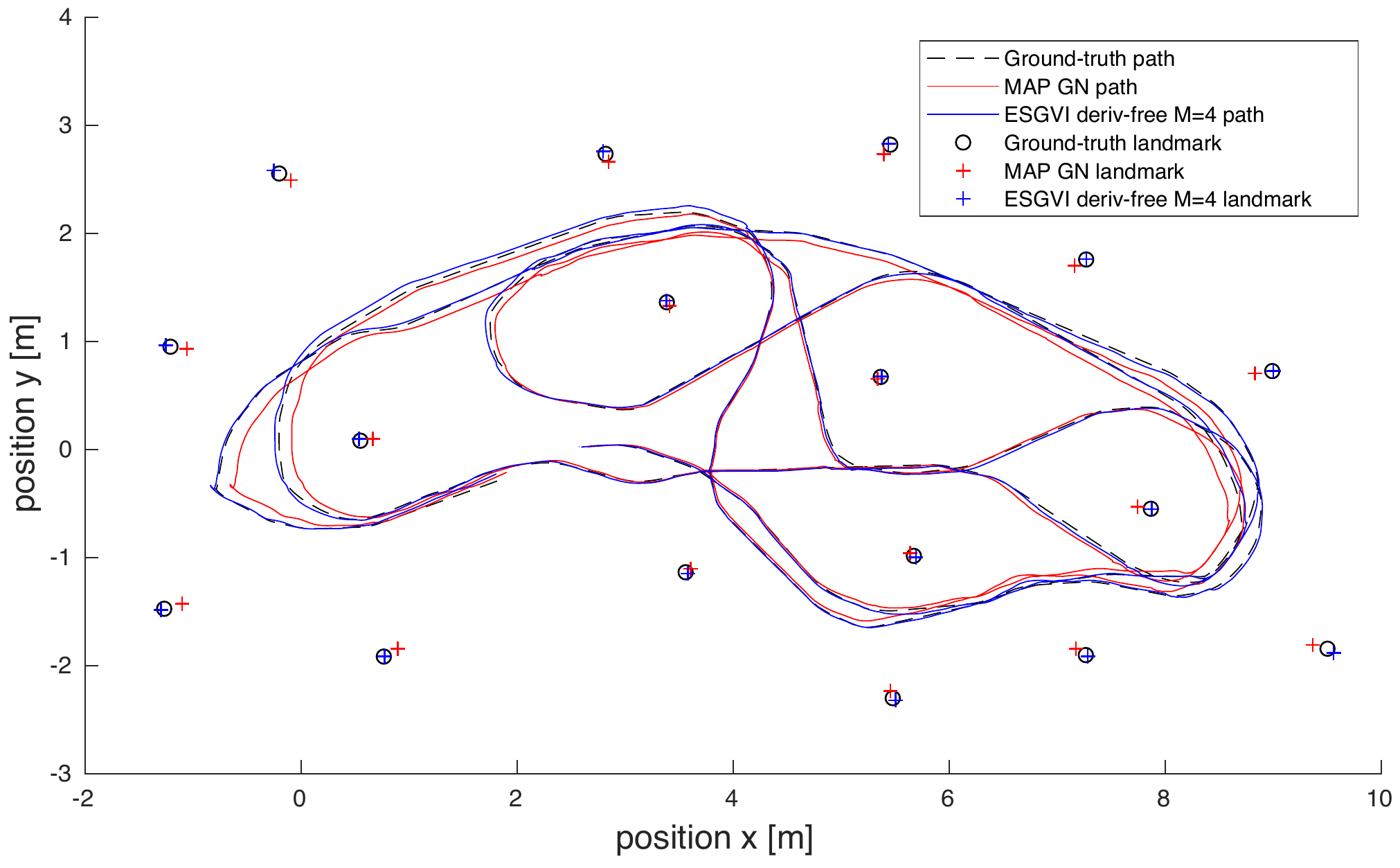}
\caption{(Experiment 3) A comparison of `\ac{MAP} GN' and `\ac{ESGVI} deriv-free M=4' on one of the six subsequences of $2000$ timestamps. Above, we see the individual error plots (with $3\sigma$ covariance envelopes) for the $x$, $y$, and $\theta$ components of the robot state as compared to groundtruth.  Below, we have an overhead view of the robot path and landmark map for the two algorithms as well as groundtruth. \change{The total landmark squared error for `\ac{MAP} GN' was 0.2214 [m$^2$] while for `ESGVI deriv-free M=4' it was 0.0168 [m$^2$].}}
\label{fig:exp3_run1}
\end{figure*}

Figure~\ref{fig:exp3_results} provides the statistical results of several variants of our \ac{ESGVI} algorithms.  We see that the number of iterations required to converge is higher than in the previous experiments, with the  \ac{ESGVI} variants requiring a few more than the corresponding \ac{MAP} algorithms.  Again, we see the \ac{ESGVI} variants reducing the loss functional, $V(q)$, further than the \ac{MAP} methods. 

\change{The bias is further away from zero for the derivative-free \ac{ESGVI} method than \ac{MAP}, which could simply be related to the relatively small number of trials compared to the previous two experiments.  Note, as in the previous experiment, we show the bias separately for (left to right) robot position, robot orientation, and landmark position to avoid combining quantities with different units.  We do not have groundtruth for velocity so do not report errors for this part of the state.}

The squared error and squared Mahalanobis distance metrics are drastically improved for the full \ac{ESGVI} methods compared to the \ac{MAP} method and even the \ac{ESGVI-GN} method.  \change{Again, we show the squared error separately for (left to right) robot position, robot orientation, and landmark position to avoid combining quantities with different units.  We can see that the improvements offered by \ac{ESGVI} on squared error are due to the robot and landmark position variables rather than robot orientation.}

Figure~\ref{fig:exp3_run1} shows the detailed error plots for one of the six subsequences for the `\ac{MAP} GN' and `\ac{ESGVI} deriv-free M=4' algorithms.  The \ac{ESGVI} path is visibly better than \ac{MAP} in most sections.  \ac{MAP} seems to have underestimated the scale of the whole solution, resulting in much worse performance on all translational variables, while performing similarly on heading error.  Both algorithms are fairly consistent, with \ac{ESGVI} being both more confident and more consistent.  The other five subsequences have similar results.

\begin{table}[t]
\caption{(Experiment 3) Wall-clock time per iteration for tested algorithms (averages over the 6 sequences).}
\begin{center}
\begin{tabular}{l|r}
algorithm  & wall-clock time \\
label & per iteration [s]  \\ \hline\hline
\ac{MAP} Newton & 5.8079  \\
\ac{MAP} GN &  1.2035 \\
\ac{ESGVI} deriv M=3 & 80.2644   \\
\ac{ESGVI-GN} deriv-free M=3 & 41.4672  \\
\ac{ESGVI} deriv-free M=4 & 164.0719 
\end{tabular}
\end{center}
\label{tab:iterationtime}
\vspace*{-0.3in}
\end{table}

Figure~\ref{fig:exp3_sparsity} shows the sparsity patterns of $\mbs{\Sigma}^{-1}$ and $\mbf{L}$, as well as the blocks of $\mbs{\Sigma}$ that are computed (the matrix is actually dense); only the patterns for the first $100$ timestamps are shown for clarity, but each subsequence is actually $2000$ timestamps long.  In terms of computational complexity, all of the algorithms for this \ac{SLAM} problem are $O(L^3 + L^2K)$ per iteration, where $L$ is the number of landmarks and $K$ is the number of timesteps.  However, the wall-clock time required by the different algorithms varies significantly due to different numbers of iterations and the accuracy with which the required expectations in~\eqref{eq:margexpectations} are computed.  Table~\ref{tab:iterationtime} reports how long each algorithm took per iteration; the \ac{ESGVI} methods come at a cost, but this may be acceptable for batch (i.e., offline) applications.  It is also worth noting that we have made little attempt to optimize our implementation.  We used a brute-force cubature method requiring $M^{N_k}$ sigmapoints where $N_k$ is the number of state variables involved in a factor.  More efficient options could be swapped in to speed up the evaluation of each factor.  Additionally, parallelization could be employed at the factor level quite easily in our approach by evaluating the expectations in~\eqref{eq:margexpectations} in parallel across several \change{compute elements}.

\section{Conclusion and Future Work}

\label{sec:conclusion}

We presented our \acf{ESGVI} approach and demonstrated that it is possible to compute a Gaussian that is `best' in terms of \ac{KL} divergence from the full Bayesian posterior, even for large-scale problems.  We exploited the fact that the joint likelihood of the state and data factors, a property of most common robotics problems, to show that the full (dense) covariance matrix is not required, only the blocks corresponding to the non-zero blocks of the (sparse) inverse covariance matrix.  We further showed how to apply cubature methods (e.g., sigmapoints) within our framework resulting in a batch inference scheme that does not require analytical derivatives, yet is applicable to large-scale problems.  The methods offer performance improvements (over \ac{MAP}) that increase as the problem becomes more nonlinear and/or the posterior less concentrated.

There are several avenues for further exploration beyond this work.  First, sample-efficient cubature methods could bring the cost of our scheme down further.  While we showed that we only need to apply cubature at the factor/marginal level, this can still be expensive for marginals involving several state variables.  We used a brute-force approach requiring $M^{N_k}$ samples for a marginal of dimension $N_k$, but there may be other alternatives that could be applied to bring the cost down.  Parallel evaluation of the factor expectations could also be worth investigating.

We used the method of \citet{takahashi73} to compute the blocks of $\mbs{\Sigma}$ corresponding to the non-zero blocks of $\mbs{\Sigma}^{-1}$, but this basic method is not always optimally efficient, requiring additional (unnecessary) blocks of $\mbs{\Sigma}$ to be computed for some \ac{GVI} problems.  It should be possible to combine this with additional modern sparsity methods such as variable reordering and Givens rotations \citep{golub96,kaess08} to improve the efficiency of this step.  

Our \ac{SLAM} experiments showed that we could carry out \ac{GVI} in a tractable manner.  However, we have not yet shown that our approach is robust to outliers.  It would certainly be worth attempting to wrap each factor expression in a robust cost function to enable a variational extension of M-estimation \citep[\S5, p. 163]{barfoot17}\citep{mactavish_crv15}.  This is typically implemented as iteratively reweighted least squares \citep{holland77}, but \ac{ESGVI} might handle robust cost functions with no modification since we compute expectations at the factor level.

There are many other practical applications of \ac{ESGVI} worth exploring beyond the simple cases presented here.  We are particularly interested in how to apply our inference approach to joint estimation-control problems and have begun an investigation along this line.

We restricted our variational estimate to be a single multivariate Gaussian, but the ideas here will likely extend to mixtures of Gaussians and possibly other approximations of the posterior as well.  We have not explored this possibility yet, but the variational approach seems to offer a logical avenue along which to do so.

Finally, we showed the possibility of extending \ac{ESGVI} to include parameter estimation through the use of an \ac{EM} setup, which typically employs the same loss functional, $V(q | \mbs{\theta})$.  In particular, we would like to represent our factor models as \acp{DNN} whose weights are unknown.  We believe that \ac{ESGVI} offers a good option for the expectation step, as we may be able to use our derivative-free version to avoid the need to take the derivative of a \ac{DNN} with respect to the state being estimated, instead just carrying out hardware-accelerated feed-forward evaluations.


\begin{acks}
We are grateful to Filip Tronarp (Aalto University) as well as Paul Furgale, Colin McManus, and Chi Hay Tong (while they were at the University of Toronto) for some early discussions on this work.  This work was supported by the Natural Sciences and Engineering Research Council of Canada.
\end{acks}

\bibliographystyle{SageH}
\bibliography{refs}

\appendix

\section{Definiteness of $\mbox{\normalfont tr}(\mbf{A}\mbf{B}\mbf{A}\mbf{B})$}

\label{sec:AppDefiniteness}

We used the fact that
\begin{equation}
\mbox{tr}\left( \mbs{\Sigma}^{(i)} \, \delta\mbs{\Sigma}^{-1} \, \mbs{\Sigma}^{(i)} \, \delta\mbs{\Sigma}^{-1} \right) \ge 0,
\end{equation}
with equality if and only if $\delta\mbs{\Sigma}^{-1} = \mbf{0}$ in our local convergence guarantee in~\eqref{eq:convguarantee}.  To show this, it is sufficient to show for $\mbf{A}$ real symmetric positive definite and $\mbf{B}$ real symmetric that
\begin{equation}
\label{eq:definiteness}
\mbox{tr}(\mbf{A}\mbf{B}\mbf{A}\mbf{B}) \ge 0,
\end{equation}
with equality if and only if $\mbf{B} = \mbf{0}$.  

We can write
\begin{multline}
\mbox{tr}(\mbf{A}\mbf{B}\mbf{A}\mbf{B}) = \mbox{vec}(\mbf{A}\mbf{B}\mbf{A})^T \mbox{vec}(\mbf{B}) \\ = \mbox{vec}(\mbf{B})^T \underbrace{\left( \mbf{A} \otimes \mbf{A} \right)}_{>0} \mbox{vec}(\mbf{B}) \ge 0,
\end{multline}
using basic properties of $\mbox{vec}(\cdot)$ and the Kronecker product, $\otimes$.  The matrix in the middle is symmetric positive definite owing to our assumptions on $\mbf{A}$.  Therefore the quadratic form is positive semi-definite with equality if and only if $\mbox{vec}(\mbf{B}) = \mbf{0}$ if and only if $\mbf{B} = \mbf{0}$.

\section{Fisher Information Matrix for a Multivariate Gaussian}

\label{sec:AppFisher}

\change{
This section provides a brief derivation for the \acf{FIM} associated with our Gaussian, $q = \mathcal{N}(\mbs{\mu}, \mbs{\Sigma})$.  \citet{magnus19} or \citet{barfoot_arxiv20} provide more detail.  If we stack up our variational parameters into a vector as
\begin{equation}
\mbs{\alpha} = \bbm \mbs{\mu} \\ \mbox{vec}\left(\mbs{\Sigma}^{-1} \right) \ebm,
\end{equation}
then we seek to show that the \ac{FIM} is
\begin{equation}
\mbs{\mathcal{I}}_{\mbs{\alpha}} = \bbm \mbs{\mathcal{I}}_{\mbs{\mu}} & \mbf{0} \\ \mbf{0} & \mbs{\mathcal{I}}_{\mbs{\Sigma}^{-1}} \ebm = \bbm \mbs{\Sigma}^{-1} & \mbf{0} \\ \mbf{0} & \frac{1}{2} \left(\mbs{\Sigma} \otimes \mbs{\Sigma}\right) \ebm. 
\end{equation}
For a Gaussian, we can use the following \ac{FIM} definition \citep{fisher22}:
\begin{equation}
\label{eq:fimdefinition}
\mbs{\mathcal{I}}_{\mbs{\alpha}} = -\mathbb{E}_q \left[ \frac{\partial^2 \ln q}{\partial \mbs{\alpha}^T \partial \mbs{\alpha}} \right].
\end{equation}
The negative log-likelihood of a Gaussian is
\begin{multline}
-\ln q = \frac{1}{2} (\mbf{x} - \mbs{\mu})^T \mbs{\Sigma}^{-1} (\mbf{x} - \mbs{\mu}) \\ - \frac{1}{2} \ln | \mbs{\Sigma}^{-1} | + \mbox{constant}.
\end{multline}
The first differential is
\begin{multline}
- d \ln q = -d\mbs{\mu}^T \mbs{\Sigma}^{-1} (\mbf{x} - \mbs{\mu})  - \frac{1}{2} \tr\left( \mbs{\Sigma} \, d\mbs{\Sigma}^{-1} \right) \\ + \frac{1}{2}  (\mbf{x} - \mbs{\mu})^T d\mbs{\Sigma}^{-1} (\mbf{x} - \mbs{\mu}) .
\end{multline}
The second differential is
\begin{multline}
- d^2 \ln q = d\mbs{\mu}^T \mbs{\Sigma}^{-1} \,d\mbs{\mu} - 2 d\mbs{\mu}^T \, d\mbs{\Sigma}^{-1} (\mbf{x} - \mbs{\mu}) \\ \qquad\qquad + \frac{1}{2} \tr\left( \left( (\mbf{x} - \mbs{\mu})(\mbf{x} - \mbs{\mu})^T - \mbf{\Sigma} \right) d^2 \mbs{\Sigma}^{-1}  \right) \\ + \frac{1}{2} \tr\left( \mbf{\Sigma} \, d\mbf{\Sigma}^{-1} \, \mbf{\Sigma} \, d\mbf{\Sigma}^{-1} \right). 
\end{multline}
The expected value of the second differential over $q$ is
\begin{eqnarray}
& & \hspace*{-0.5in} -\mathbb{E}_q\left[ d^2 \ln q \right] \nonumber \\ & = & d\mbs{\mu}^T \mbs{\Sigma}^{-1} \,d\mbs{\mu}  + \frac{1}{2} \tr\left( \mbf{\Sigma} \, d\mbf{\Sigma}^{-1} \, \mbf{\Sigma} \, d\mbf{\Sigma}^{-1} \right) \nonumber \\ & = & d\mbs{\mu}^T \mbs{\Sigma}^{-1} \,d\mbs{\mu}  \\ & & \quad + \; \frac{1}{2} \vec\left(d\mbf{\Sigma}^{-1}\right)^T \left( \mbf{\Sigma} \otimes \mbf{\Sigma} \right) \vec\left(d\mbf{\Sigma}^{-1}\right). \nonumber 
\end{eqnarray}
In matrix form this is
\begin{equation}
-\mathbb{E}_q\left[ d^2 \ln q \right] =  d\mbs{\alpha}^T \bbm \mbs{\Sigma}^{-1} & \mbf{0} \\ \mbf{0} & \frac{1}{2} \left(\mbs{\Sigma} \otimes \mbs{\Sigma}\right) \ebm \,d\mbs{\alpha},
\end{equation}
where 
\begin{equation}
d\mbs{\alpha} = \bbm d\mbs{\mu} \\ \mbox{vec}\left(d\mbs{\Sigma}^{-1} \right) \ebm.
\end{equation}
Turning the differentials into partial derivatives we have
\begin{equation}
\mbs{\mathcal{I}}_{\mbs{\alpha}} = -\mathbb{E}_q \left[ \frac{\partial^2 \ln q}{\partial \mbs{\alpha}^T \partial \mbs{\alpha}} \right] = \bbm \mbs{\Sigma}^{-1} & \mbf{0} \\ \mbf{0} & \frac{1}{2} \left(\mbs{\Sigma} \otimes \mbs{\Sigma}\right) \ebm.
\end{equation}
The inverse \ac{FIM} is given by
\begin{equation}
\mbs{\mathcal{I}}_{\mbs{\alpha}}^{-1} = \bbm \mbs{\mathcal{I}}_{\mbs{\mu}}^{-1} & \mbf{0} \\ \mbf{0} & \mbs{\mathcal{I}}^{-1}_{\mbs{\Sigma}^{-1}} \ebm = \bbm \mbs{\Sigma} & \mbf{0} \\ \mbf{0} & 2 \left(\mbs{\Sigma}^{-1} \otimes \mbs{\Sigma}^{-1}\right) \ebm,
\end{equation}
using properties of linear and Kronecker algebra.
}

\section{Direct Derivation of Derivative-Free \ac{ESGVI}}

\label{sec:AppDirectESGVI}

\change{
In our main \ac{ESGVI} derivation, we chose to (i) define a scheme to minimize $V(q)$ with respect to $q$ and then (ii) exploit $\phi(\mbf{x}) = \sum_{k=1}^K \phi_k(\mbf{x}_k)$ to make the scheme efficient. Here we show that we can carry out (i) and (ii) in the opposite order to streamline the derivation of the derivative-free version of \ac{ESGVI}, avoiding the need for Stein's lemma.   

Figure~\ref{fig:commdiagram} shows two paths to get to the derivative-free version of \ac{ESGVI} starting from our loss functional, $V$.
The main body of the paper took the counter-clockwise path that goes down, across to the right, and then up.  We will now demonstrate the clockwise path that goes right and then down.

\newcommand{\tlap}[1]{\vbox to 0pt{\vss\hbox{#1}}}
\newcommand{\blap}[1]{\vbox to 0pt{\hbox{#1}\vss}}
\begin{figure}[t]
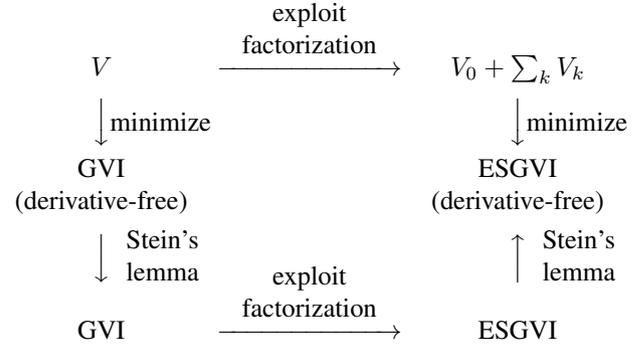

\change{
\begin{equation*}
\begin{CD}
 V @>\tlap{\begin{tabular}{c}\mbox{exploit} \\ \mbox{factorization}\end{tabular}}>> V_0 + \sum_k V_k  \\
 @VV\mbox{minimize} V        @VV \mbox{minimize} V \\
\begin{tabular}{c}\mbox{GVI} \\ \mbox{(derivative-free)}\end{tabular}    @.  \begin{tabular}{c}\mbox{ESGVI} \\ \mbox{(derivative-free)}\end{tabular}\\
@VV\begin{tabular}{c}\mbox{Stein's} \\ \mbox{lemma}\end{tabular}V        @AA \begin{tabular}{c}\mbox{Stein's} \\ \mbox{lemma}\end{tabular} A \\
\mbox{\ac{GVI}}    @>\tlap{\begin{tabular}{c}\mbox{exploit} \\ \mbox{factorization}\end{tabular}}>>  \mbox{\ac{ESGVI}} \\
\end{CD}\qquad
\end{equation*}
}
\vspace*{-0.15in}
\caption{Commutative diagram showing two paths to get to the derivative-free version of \ac{ESGVI} starting from the loss functional, $V$.}
\label{fig:commdiagram}
\vspace*{-0.1in}
\end{figure}

Inserting $\phi(\mbf{x}) = \sum_{k=1}^K \phi_k(\mbf{x}_k)$, the loss functional becomes
\begin{equation}
V(q) = \sum_{k=1}^K \underbrace{\mathbb{E}_{q_k}[ \phi_k(\mbf{x}_k)]}_{V_k(q_k)} + \underbrace{\frac{1}{2} \ln \left( |\mbs{\Sigma}^{-1}| \right)}_{V_0},
\end{equation}
where the expectations are now over the marginal, $q_k(\mbf{x}_k) = \mathcal{N}(\mbs{\mu}_k, \mbs{\Sigma}_{kk})$ with
\begin{equation}
\mbs{\mu}_k = \mbf{P}_k \mbs{\mu}, \quad \mbs{\Sigma}_{kk} = \mbf{P}_k \mbs{\Sigma} \mbf{P}_k^T,
\end{equation}
and where $\mbf{P}_k$ is a projection matrix.  We then take the first derivative with respect to $\mbs{\mu}$:
\begin{eqnarray}
\frac{\partial V(q)}{\partial \mbs{\mu}^T} & = & \frac{\partial}{\partial \mbs{\mu}^T} \left(  \sum_{k=1}^K \mathbb{E}_{q_k}[ \phi_k(\mbf{x}_k)] + \frac{1}{2} \ln \left( |\mbs{\Sigma}^{-1}| \right) \right) \nonumber \\ & = & \sum_{k=1}^K \frac{\partial}{\partial \mbs{\mu}^T} \mathbb{E}_{q_k}[ \phi_k(\mbf{x}_k)] \nonumber \\ & = & \sum_{k=1}^K \mbf{P}_k^T \frac{\partial}{\partial \mbs{\mu}_k^T} \mathbb{E}_{q_k}[ \phi_k(\mbf{x}_k)] \nonumber \\ & = & \sum_{k=1}^K \mbf{P}_k^T \mbs{\Sigma}_{kk}^{-1} \underbrace{\mathbb{E}_{q_k}\left[ (\mbf{x}_k - \mbs{\mu}_k) \phi_k(\mbf{x}_k)\right]}_{\mbox{cubature}}.
\end{eqnarray}
The last step comes from~\eqref{eq:deriv1a} applied at the marginal level.  Similarly, we have for the second derivative with respect to $\mbs{\mu}$ that
\begin{eqnarray}
& & \hspace{-0.3in}\frac{\partial^2 V(q)}{\partial \mbs{\mu}^T \partial \mbs{\mu}} \nonumber \\ & = & \frac{\partial^2}{\partial \mbs{\mu}^T \partial \mbs{\mu}} \left(  \sum_{k=1}^K \mathbb{E}_{q_k}[ \phi_k(\mbf{x}_k)] + \frac{1}{2} \ln \left( |\mbs{\Sigma}^{-1}| \right) \right) \nonumber \\ & = &  \sum_{k=1}^K \frac{\partial^2}{\partial \mbs{\mu}^T \partial \mbs{\mu}} \mathbb{E}_{q_k}[ \phi_k(\mbf{x}_k)]  \nonumber \\ & = & \sum_{k=1}^K \mbf{P}_k^T \left(\frac{\partial^2}{\partial \mbs{\mu}_k^T \partial \mbs{\mu}_k} \mathbb{E}_{q_k}[ \phi_k(\mbf{x}_k)] \right) \mbf{P}_k  \nonumber \\ & = & \sum_{k=1}^K \mbf{P}_k^T \biggl( \mbs{\Sigma}_{kk}^{-1} \underbrace{\mathbb{E}_{q_k}\left[ (\mbf{x}_k - \mbs{\mu}_k)  (\mbf{x}_k - \mbs{\mu}_k)^T \phi_k(\mbf{x}_k)\right]}_{\mbox{cubature}}  \nonumber \\ & & \qquad\qquad \times \; \mbs{\Sigma}_{kk}^{-1} - \mbs{\Sigma}_{kk}^{-1}  \underbrace{\mathbb{E}_{q_k}\left[ \phi_k(\mbf{x}_k)\right]}_{\mbox{cubature}}\biggr) \mbf{P}_k,
\end{eqnarray}
where we made use of~\eqref{eq:deriv1b} at the marginal level in the last step.

The update scheme is then
\begin{subequations}
\begin{eqnarray}
\left(\mbs{\Sigma}^{-1}\right)^{(i+1)} & = &  \left.\frac{\partial^2 V(q)}{\partial \mbs{\mu}^T \partial \mbs{\mu}}\right|_{q^{(i)}},  \\
\left(\mbs{\Sigma}^{-1}\right)^{(i+1)} \, \delta\mbs{\mu} & = & - \left.\frac{\partial V(q)}{\partial \mbs{\mu}^T}\right|_{q^{(i)}},   \\
\mbs{\mu}^{(i+1)} & = & \mbs{\mu}^{(i)} + \delta\mbs{\mu}, 
\end{eqnarray}
\end{subequations}
which is identical to the derivative-free \ac{ESGVI} approach from the main body of the paper.  The sparsity of the left-hand side comes from the use of the projection matrices.  We did not require the use of Stein's lemma \citep{stein81} using this derivation, which also means that we have made no assumptions about the differentiability of $\phi_k(\mbf{x}_k)$ with respect to $\mbf{x}_k$.   A similar streamlined derivation can be worked out for the Gauss-Newton variant.  
}

\section{Derivation of Derivative Identities~\eqref{eq:derivident}}

\label{sec:AppDerivIdent}

\change{
To show~\eqref{eq:derivident1}, we can use~\eqref{eq:deriv1a} to write
\begin{equation}
\frac{\partial}{\partial \mbs{\mu}^T} \mathbb{E}_q[ f(\mbf{x})] =  \mbs{\Sigma}^{-1} \mathbb{E}_q[ (\mbf{x} - \mbs{\mu}) f(\mbf{x})],
\end{equation}
which can also be found in~\citet{opper09}.  Applying Stein's lemma from~\eqref{eq:stein1} we immediately have
\begin{equation}
\frac{\partial}{\partial \mbs{\mu}^T} \mathbb{E}_q[ f(\mbf{x})] = \mathbb{E}_q\left[ \frac{\partial f(\mbf{x})}{\partial \mbf{x}^T}\right],
\end{equation}
the desired result.

To show~\eqref{eq:derivident2}, we can use~\eqref{eq:deriv1b} to write
\begin{multline}
\label{eq:appderiv1b}
\frac{\partial^2}{\partial \mbs{\mu}^T \partial \mbs{\mu}} \mathbb{E}_q[ f(\mbf{x})]  = \mbs{\Sigma}^{-1} \mathbb{E}_q[ (\mbf{x} - \mbs{\mu}) (\mbf{x} - \mbs{\mu})^T f(\mbf{x})] \mbs{\Sigma}^{-1}  \\ - \; \mbs{\Sigma}^{-1} \,\mathbb{E}_q[f(\mbf{x})],
\end{multline}
again due to~\citet{opper09}.  Applying Stein's lemma from~\eqref{eq:stein2} we have
\begin{equation}
\frac{\partial^2}{\partial \mbs{\mu}^T \partial \mbs{\mu}}  \mathbb{E}_q[ f(\mbf{x})] = \mathbb{E}_q\left[ \frac{\partial^2 f(\mbf{x})}{\partial \mbf{x}^T \partial \mbf{x}}\right],
\end{equation}
the desired result.  Similarly to~\eqref{eq:deriv2}, we have
\begin{multline}
\frac{\partial}{\partial \mbs{\Sigma}^{-1}} \mathbb{E}_q[ f(\mbf{x})] = -\frac{1}{2}\mathbb{E}_q[ (\mbf{x} - \mbs{\mu}) (\mbf{x} - \mbs{\mu})^T f(\mbf{x})]  \\ + \; \frac{1}{2}  \mbs{\Sigma} \,\mathbb{E}_q[f(\mbf{x})],
\end{multline}
which is again confirmed by~\citet{opper09}.  Comparing the right-hand side of this to the right-hand side of~\eqref{eq:appderiv1b}, we have
\begin{multline}
-2 \mbs{\Sigma}^{-1} \left( \frac{\partial}{\partial \mbs{\Sigma}^{-1}} \mathbb{E}_q[ f(\mbf{x})] \right) \mbs{\Sigma}^{-1} = \frac{\partial^2}{\partial \mbs{\mu}^T \partial \mbs{\mu}} \mathbb{E}_q[ f(\mbf{x})]  \\ = \mathbb{E}_q\left[ \frac{\partial^2 f(\mbf{x})}{\partial \mbf{x}^T \partial \mbf{x}}\right],
\end{multline}
which shows the second part of~\eqref{eq:derivident2}.
}

\end{document}